\documentclass[journal,twoside]{IEEEtran}

\usepackage{hyperref}
\usepackage{amsmath}
\usepackage{graphicx}
\usepackage{epstopdf}
\usepackage{cite}
\usepackage{multirow}
\usepackage{amssymb}
\usepackage{url}

\usepackage{algorithm}  
\usepackage{algpseudocode}  
\usepackage{amsmath}

\usepackage{setspace}

\usepackage{subfigure}
\usepackage{color}

\usepackage{upgreek}

\usepackage{bbm}

\ifCLASSINFOpdf

\else
 
\fi

\begin{document}

\title{Hyperspectral Image Classification With Contrastive Graph Convolutional Network}

\author{~Wentao~Yu,~\IEEEmembership{Student Member,~IEEE},~Sheng Wan,~Guangyu Li,~Jian~Yang,~\IEEEmembership{Member,~IEEE}~and Chen~Gong,~\IEEEmembership{Member,~IEEE}

\thanks{The authors are with the PCA Laboratory, the Key Laboratory of Intelligent Perception and Systems for High-Dimensional Information of Ministry of Education, the Jiangsu Key Laboratory of Image and Video Understanding for Social Security, and the School of Computer Science and Engineering, Nanjing University of Science and Technology, Nanjing 210094, China (e-mail:~wentao.yu@njust.edu.cn; wansheng315@hotmail.com;~liguangyu0627@hotmail.com;~csjyang@njust.edu.cn; chen.gong@njust.edu.cn).}
}

\markboth{IEEE TRANSACTIONS ON GEOSCIENCE AND REMOTE SENSING, ~Vol.~XX, No.~X, XX~202X}{YU \MakeLowercase{\textit{et al.}}: HSI CLASSIFICATION WITH ConGCN}

\maketitle

\begin{abstract}
Recently, \textbf{G}raph \textbf{C}onvolutional \textbf{N}etwork (GCN) has been widely used in \textbf{H}yperspectral \textbf{I}mage (HSI) classification due to its satisfactory performance. However, the number of labeled pixels is very limited in HSI, and thus the available supervision information is usually insufficient, which will inevitably degrade the representation ability of most existing GCN-based methods. To enhance the feature representation ability, in this paper, a GCN model with contrastive learning is proposed to explore the supervision signals contained in both spectral information and spatial relations, which is termed \textbf{Con}trastive \textbf{G}raph \textbf{C}onvolutional \textbf{N}etwork (ConGCN), for HSI classification. First, in order to mine sufficient supervision signals from spectral information, a semi-supervised contrastive loss function is utilized to maximize the agreement between different views of the same node or the nodes from the same land cover category. Second, to extract the precious yet implicit spatial relations in HSI, a graph generative loss function is leveraged to explore supplementary supervision signals contained in the graph topology. In addition, an adaptive graph augmentation technique is designed to flexibly incorporate the spectral-spatial priors of HSI, which helps facilitate the subsequent contrastive representation learning. The extensive experimental results on four typical benchmark datasets firmly demonstrate the effectiveness of the proposed ConGCN in both qualitative and quantitative aspects.

\end{abstract}

\begin{IEEEkeywords}
Contrastive learning, \textbf{G}raph \textbf{C}onvolutional \textbf{N}etwork (GCN), graph augumentation, \textbf{H}yperspectral \textbf{I}mage (HSI) classification.
\end{IEEEkeywords}
\IEEEpeerreviewmaketitle

\section{Introduction}
\label{Introduction}
\IEEEPARstart{H}{YPERSPECTRAL} Image (HSI) classification plays an increasingly significant role in environmental monitoring, precision agriculture, mineral identification, and land cover classification. Different from conventional remote sensing images, HSI is composed of diverse contiguous spectral bands, providing detailed spectral information in addition to spatial relations. This property is beneficial to attribute each pixel of HSI into a certain category.

In the past few decades, various methods have been proposed for HSI classification. The early-staged methods are mainly based on conventional pattern recognition methods. In spite of the achievements obtained by these methods, they suffer from hand-crafted configurations. To avoid intricate feature engineering, deep learning-based methods can obtain high-level representations via gradually aggregating the low-level features, and have attracted increasing attention in recent years. As one of the most representative methods, Convolutional Neural Networks (CNN) have achieved state-of-the-art performance due to their powerful generalization ability. However, the receptive field of CNN is restricted by a regular square area, and thus CNN fail to adaptively capture the geometric variations of different land cover categories in HSI. To deal with this deficiency, recently, a number of Graph Convolutional Network (GCN) based HSI classification methods \cite{wan2019multiscale, wan2020hyperspectral, wan2021dual, 9091940, bai2021hyperspectral} have been proposed and achieved promising performance. For example, in \cite{wan2019multiscale}, a multiscale dynamic GCN (MDGCN) was presented to obtain the dynamic graph and explore spectral-spatial correlations at multiple scales. Based on MDGCN, to tap into the advantage of contextual information, context-aware dynamic GCN \cite{wan2020hyperspectral} was proposed to further mine node features for HSI classification. To overcome the drawback of the manually constructed graph, Wan \textit{et al.} develop a new dual interactive GCN to adaptively learn a discriminative region-induced graph \cite{wan2021dual}. In \cite{9091940}, a semi-supervised nonlocal graph convolutional network was developed to exploit labeled and unlabeled data, simultaneously. Moreover, Bai \textit{et al.} \cite{bai2021hyperspectral} developed a deep attention GCN to mine HSI features via focusing on the spectral information which has a large impact on classification.

However, the representation ability of most existing GCN-based methods is still limited due to insufficient labeled pixels \cite{9372392}. To accommodate this issue, we aim to sufficiently extract the supervision signals carried by hyperspectral data themselves for network training. Specifically, we propose to use contrastive learning~\cite{chen2020simple, he2020momentum} to encode the similarities among the spectral signatures of image regions. Contrastive learning is one of the representative self-supervised learning approaches and it has emerged as a powerful technique for graph representation learning recently \cite{wan2020contrastive, wan2021contrastive}. Most graph contrastive learning methods first perform stochastic augmentation on the input graph to obtain two graph views and then maximize the agreement of representations in the two views \cite{zhu2021graph}. Different from traditional contrastive methods that only utilize unlabeled data for model training, our proposed method additionally incorporates class information to improve the discriminative power of the generated representations. To be specific, we devise a semi-supervised contrastive loss to make full use of the pairwise similarities among examples based on their spectral features. In addition to the supervision signals contained in spectral information, we develop a graph generative loss to explore supplementary supervision signals from the spatial relations among image regions. As a consequence, the originally scarce supervision information can be further enriched by exploring the knowledge from both spectral information and spatial relations of HSI, and thereby leading to enhanced data representations. Last but not least, we devise an adaptive graph augmentation technique at both graph topology and node attribute levels, which is able to incorporate spectral-spatial priors to boost the performance of contrastive learning. To the best of our knowledge, our proposed method is the first work combining GCN with contrastive learning for HSI classification, so we term our method as ``\textbf{Con}trastive \textbf{G}raph \textbf{C}onvolutional \textbf{N}etwork’’ (ConGCN).

To our best knowledge, there have been two works \cite{9664575, hu2021contrastive} employing contrastive learning for HSI classification. However, they simply use the traditional paradigm of contrastive learning by utilizing unlabeled examples for pre-training and fine-tuning the model with few labeled examples. In order to improve the discriminative power of the generated representations, our proposed method additionally incorporates the available class information by using a semi-supervised contrastive loss function. Furthermore, our proposed method incorporates spatial relations to the contrastive objective. As a result, the proposed ConGCN can produce more effective feature representations than ResNet-50 in \cite{9664575} and the transformer model in \cite{hu2021contrastive}. The advantage of our ConGCN has also been empirically demonstrated in Section~\ref{Experiments}.

To summary, the contributions of our work are as follows:
\begin{itemize}
\item To the best of our knowledge, it is the first time to adopt a contrastive GCN model for HSI classification, which helps to extract rich supervision information for enhancing representation ability and classification performance of network.

\item We devise an adaptive graph augmentation technique via incorporating spectral-spatial priors, which helps to boost the performance of contrastive learning.

\item By performing localized and hierarchical graph convolution simultaneously, both local and global contextual information of HSI can be leveraged for expressive representation learning.

\item A semi-supervised contrastive loss and a graph generative loss are designed to exploit the supervision signals contained in spectral domain and spatial relations,
respectively.
\end{itemize}

\section{Related Works}
This section reviews some typical prior works related to this paper, which include HSI classification methods, GCN models, and contrastive learning approaches.\vspace{-1em}

\subsection{HSI Classification Methods}
During the past few decades, abundant methods have been put forward for HSI classification. The early-staged methods were mainly anchored on conventional machine learning methods, such as kernel-based methods \cite{camps2005kernel}, $k$-nearest-neighbor classifier \cite{5555996}, Support Vector Machine (SVM) \cite{peng2015region}, and Markov random field \cite{zhang2017multifeature}. Unfortunately, they usually relied on empirically designed hand-crafted features, so their performances are often far from perfect.

To tackle this problem, a number of deep methods \cite{chen2016deep} based on CNN have been further employed to promote HSI classification. Despite the fact that the CNN-based methods display encouraging performances in some cases, they still have several defects. First of all, the receptive field of CNN is a regular square area, therefore CNN-based methods are unable to adaptively capture or perceive the geometric variations of different land cover categories in HSI. Besides, the weights of each convolution kernel are identical in different spectral bands. Consequently, the details of land cover boundaries are probably lost after feature abstraction, so the pixels around boundaries are likely to be misclassified due to the inflexible convolution kernel. As a sequel, more and more GCN-based HSI classification methods are proposed to address the above problems, which will be introduced below.\vspace{-1em}

\subsection{GCN and Its Application to HSI Classification}
Graph Neural Network (GNN) \cite{gori2005new} mapped the graph or its vertices to an Euclidean space via a transfer function. Owning to this, GNN is capable of processing graph-structured non-Euclidean data, therefore showing better flexibility and adaptability than CNN. After Bruna \textit{et al.} \cite{bruna2013spectral} introduced graph Laplacian matrix for graph convolution, many GCN models have been proposed and obtained promising performance. Among these methods, Kipf and Welling \cite{kipf2016semi} innovatively applied GCN to semi-supervised learning, which approximated spectral graph convolutions in a localized region so that the proposed method was able to learn hidden representation via encoding both graph structure and node features. However, it belongs to transductive methods, which fails to classify unseen or newly added nodes. To tackle with this issue, Hamilton \textit{et al.}~\cite{hamilton2017inductive} developed an inductive framework termed GraphSAGE via generalizing the simple graph convolution to trainable aggregation functions.

For HSI classification, GCN is able to capture and preserve the boundaries of different land cover categories flexibly due to its good ability in processing non-Euclidean data. As a result, GCN-based HSI classification methods have emerged and attracted a lot of attention. To the best of our knowledge, Qin \textit{et al.} \cite{qin2018spectral} were the first to introduce GCN into HSI classification, which leveraged the features of both adjacency nodes in graph and the neighbor pixels in the HSI. To avoid the imprecise initial graph, Wan \textit{et al.} \cite{wan2019multiscale} proposed to refine the graph gradually during the convolution process, so graph convolution was operated on a dynamic graph rather than a predefined fixed graph. To sufficiently explore the contextual information, Wan \textit{et al.} \cite{wan2020hyperspectral} proposed the context-aware dynamic GCN to capture relations among the regions originally far away in the original spatial positions.

However, the generalizability of above-mentioned methods is usually limited due to their transductive setting. To solve this problem, Ding \textit{et al.} \cite{ding2021multiscale} proposed a multiscale graph network combining GraphSAGE with context-aware learning to understand the global and local information in a graph. However, it only utilized spectral features to construct the graph. To incorporate the relationship between adjacent nodes at the stage of graph construction, Guo \textit{et al.} \cite{guo2021dual} put forward a dual graph U-net via integrating spatial graph and spectral graph simultaneously.

Nonetheless, the representation ability of most existing GCN-based methods is still limited due to the inadequate labeled examples. To enhance the representation ability, in our work, we explore the supervision signals contained in spectral information by employing contrastive learning, as well as spatial relations by deploying GCN. Hence, the proposed method can obtain better classification results than existing methods.

\subsection{Contrastive Learning}
Contrastive learning is one of self-supervised representation learning methods, which aims to obtain discriminative feature representation by leveraging the similarities and dissimilarities between examples \cite{liu2021understand}. It has gained increasing attention due to the promising results in various fields such as computer vision \cite{chen2020simple, he2020momentum} and natural language processing~\cite{kong2019mutual}.

SimCLR \cite{chen2020simple} is a well-known contrastive learning method and is anchored on a Siamese network that learns latent representations via maximizing agreement between differently augmented views of the same example. Similar to SimCLR, MoCo \cite{he2020momentum} utilized InfoNCE loss \cite{oord2018representation} and stored negative examples by using a memory bank rather than a large mini-batch size.

The marriage of contrastive learning and GCN has also been observed in recent years. For example, Petar \textit{et al.} \cite{velickovic2019deep} proposed to learn graph representations in an unsupervised manner via classifying local-global pairs and negative-sampled counterparts. Besides, to address the challenge of data heterogeneity in graphs, You \textit{et al.} \cite{you2020graph} developed graph contrastive learning with four types of graph data augmentations, each of which imposed certain prior on graph data and encoded the extent and pattern. Apart from this work, Peng \textit{et al.} \cite{peng2020graph} introduced a novel concept of graphical mutual information for graph representation learning, which generalized the idea of conventional mutual information computations from vector space to the graph domain.

Nevertheless, these methods are not applicable to HSI classification studied in this paper, as they fail to consider the spatial and spectral cues which are naturally and uniquely inherited by HSIs. Besides, they also ignore the class information carried by a handful of labeled data for HSI classification. Consequently, in this paper, we devise a new contrastive GCN which is able to acquire discriminative representations for accurate HSI classification.

\section{Pipeline of the Proposed Method}
\begin{figure*}[htpb]
	\centering
	\includegraphics[width=18cm]{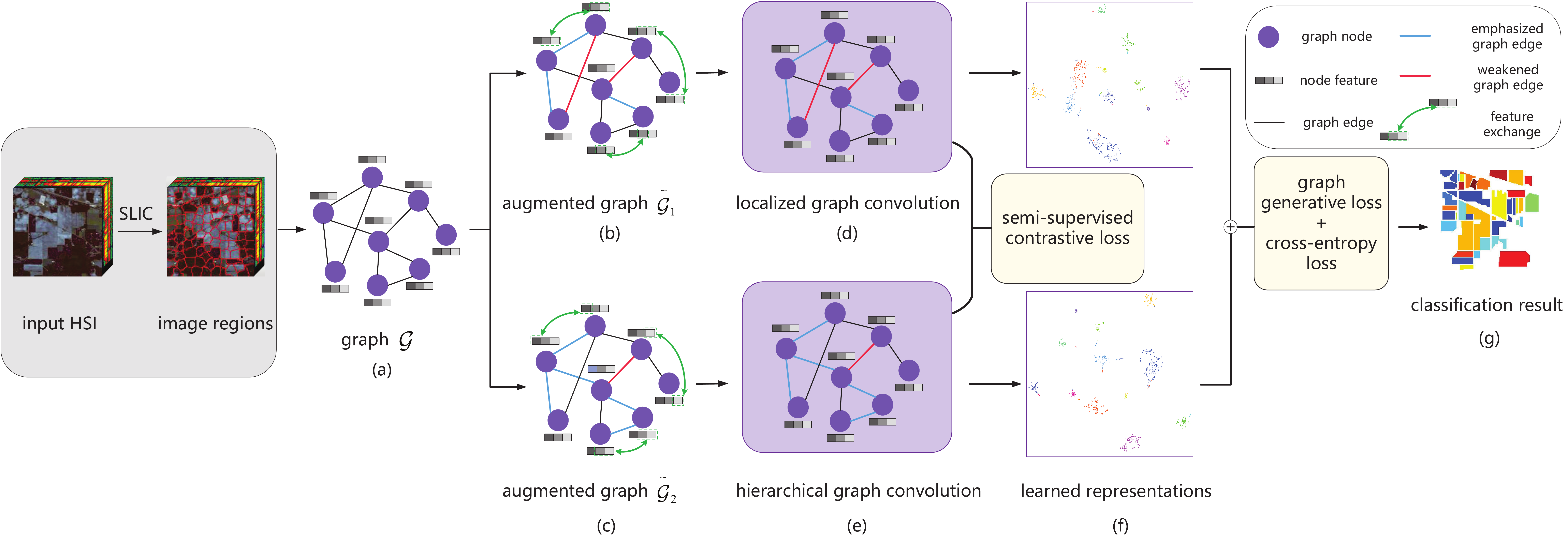}
	\caption{Framework of our proposed ConGCN. In this figure, purple circles and black lines represent graph nodes and edges, respectively. In (a), a graph $\mathcal{G}$ is constructed over the image regions produced by SLIC algorithm. The notations $\tilde{\mathcal{G}}_{1}$ and $\tilde{\mathcal{G}}_{2}$ in (b) and (c) denote two different graph views generated by the proposed adaptive graph augmentation technique. In (b) and (c), the blue lines and red lines represent emphasized edges and weakened edges, respectively. Besides, the green arrows denote the exchange of node features. (d) and (e) represent localized graph convolution and hierarchical graph convolution, respectively. The semi-supervised contrastive loss is computed across the two graph views. (f) denotes the node representations learned from the two views. In (g), the classification result is acquired via integrating the learned representations. Here, cross-entropy loss is used to penalize the difference between the output and the labels of the initially labeled seed superpixels. In addition, the graph generative loss is employed to exploit the topological information of the graph.}
	\label{fig1}
\end{figure*}

This section describes the pipeline of our proposed ConGCN method (see Fig.~\ref{fig1}). When an input HSI is given, it is firstly segmented into a set of compact image regions by using the Simple Linear Iterative Clustering (SLIC) algorithm~\cite{achanta2012slic}. Next, a graph (\textit{i.e.}, $\mathcal{G}$) is constructed by treating each of the image regions as a graph node. After that, the proposed adaptive graph augmentation (Section~\ref{IV-B}) is conducted on this graph to obtain two augmented graphs (\textit{i.e.}, $\tilde{\mathcal{G}}_{1}$ and $\tilde{\mathcal{G}}_{2}$). Finally, localized and hierarchical graph convolution (Section~\ref{IV-C}) are performed based on the two augmented graphs to obtain node representations, where the semi-supervised contrastive loss (Section~\ref{IV-D}) and the graph generative loss (Section~\ref{IV-E}) are utilized for network training (Section~\ref{IV-F}).

As mentioned above, classical SLIC \cite{achanta2012slic} algorithm is adopted to segment the entire HSI into a few compact superpixels. This is because there are numerous image pixels in an HSI, and thus constructing a pixel-level graph could be extremely time-consuming. By using SLIC, each superpixel clustering a set of homogeneous pixels with strong spectral-spatial similarity is treated as a graph node. Here, the node features correspond to the average spectral signatures of the pixels involved in the corresponding superpixel. Hence, the number of graph nodes can be significantly reduced to improve the computational efficiency. Another benefit of using the superpixel segmentation is that the generated image regions can preserve the local structural information of HSI, as adjacent pixels with high spatial consistency usually have a large probability to belong to the same land cover category.

In our proposed GCN-based method, an undirected graph made up of $n = l + u$ nodes is defined as $\mathcal{G}=\langle\mathcal{V}, \mathcal{E}\rangle$, where $\mathcal{V}=\left\{\mathbf{x}_{1}, \dots, \mathbf{x}_{l}, \mathbf{x}_{l+1}, \dots, \mathbf{x}_{n}\right\}$ is the node set containing all examples (\textit{i.e.}, superpixels), $\mathcal{E}$ is the edge set modeling the connectivity among the nodes. In $\mathcal{V}$, the first $l$ nodes are labeled, while the remaining $u$ nodes are unlabeled. Note that the label of each superpixel is determined by the most frequent label within this superpixel. The adjacency matrix of $\mathcal{G}$ denoted as $\mathbf{A}$ can be calculated as
\begin{equation}\label{eq1}
{{\mathbf{A}}_{ij}}=\left\{ \begin{array}{*{35}{l}}
  {{e}^{-\gamma {{\left\| {{\mathbf{x}}_{i}}-{{\mathbf{x}}_{j}} \right\|}_{2}^{2}}}}, & \text{ if }{{\mathbf{x}}_{i}}\in \mathcal{N}\left( {{\mathbf{x}}_{j}} \right)\text{ or }{{\mathbf{x}}_{j}}\in \mathcal{N}\left( {{\mathbf{x}}_{i}} \right)  \\
  0, & \text{ otherwise}  \\
\end{array} ,\right.
\end{equation}
where $\gamma$ is a temperature parameter and is set to 0.2 according to~\cite{wan2019multiscale, wan2020hyperspectral, wan2021dual}, $||\mathbf{x}_{i} - \mathbf{x}_{j}||_{2}$ calculates the Euclidean distance between the graph nodes $\mathbf{x}_{i}$ and $\mathbf{x}_{j}$, and $\mathcal{N}(\mathbf{x}_{i})$ is the neighbor set of $\mathbf{x}_{i}$. Other important symbols used throughout this paper are listed in~\autoref{table_symbols}.
\begin{table}[t]
\footnotesize
\centering
\caption{Symbols used throughout this paper.}
\label{table_symbols}
\begin{tabular}{cc}
\hline
\hline
Symbols                                 & Description                                                        \\ \hline
$\mathbf{X}$                          & feature matrix of graph $\mathcal{G}$                          \\
$\mathbf{X}_\text{labeled}$             & feature matrix of labeled examples                               \\
$\mathbf{Y}$                          & label matrix of graph $\mathcal{G}$                          \\
$\mathbf{Z}$                           & network output                              \\
$\mathbf{x}_{i}$                           & the $i$-th node in graph $\mathcal{G}$                              \\
$d$                                    & feature dimension of $\mathbf{x}_{i}$                        \\
$D$                                    & generalized Mahalanobis distance                              \\
$I$                                    & mutual information                              \\ \hline
\hline
\end{tabular}
\end{table}

\section{Adaptive Graph Augmentation}
\label{IV-B}
The core of contrastive learning is to maximize the agreement between differently augmented views of the same example, where data augmentation turns out to be an important prerequisite in contrastive learning. Nevertheless, how to obtain an augmentation technique that is beneficial to graph representation learning remains a challenge due to the non-Euclidean properties of graph-structured data. To address this issue, we propose an adaptive graph augmentation technique, which is able to incorporate spectral-spatial priors in HSI to boost the performance of contrastive learning. Inspired by \cite{zhu2021graph}, the principle of our proposed graph augmentation method is to preserve important structures and attributes while perturbing possibly unimportant edges and features. The reason is that it will guide the model to ignore the noise introduced by unimportant edges and features, thus helping to learn important patterns underneath the input graph. To evaluate the importance of edges and features, we leverage generalized Mahalanobis distance \cite{wan2020hyperspectral} and mutual information \cite{ross2014mutual}, respectively, which will be later explained.

The proposed augmentation technique for each graph view is made up of spatial-level graph augmentation (Section~\ref{Spatial-Level}) and spectral-level graph augmentation (Section~\ref{Spectral-Level}). In spatial-level graph augmentation, we adaptively emphasize the important edges while weakening unimportant edges, to preserve the intrinsic structure of the graph topology. In spectral-level graph augmentation, we adaptively exchange the features of adjacent nodes, to explore diverse contexts across different graph views for contrastive learning and help the network obtain the improved representations.\vspace{-0.73em}

\subsection{Spatial-Level Graph Augmentation}
\label{Spatial-Level}
In our proposed method, graph edges reflect the spatial relations among image regions and also constitute the topological structure of data. As a result, the edges are critical in ConGCN to yield good representations. However, most existing graph contrastive methods neglect the intrinsic topological properties of graphs when performing graph augmentation, since they uniformly drop the graph edges, which could result in suboptimal performance. For example, removing some influential edges will deteriorate the graph topology and thus result in inaccurate representations. To tackle this problem, we aim to adaptively emphasize or weaken edges according to their influences in the graph. 

\begin{figure}[t]
  \centering
  \includegraphics[width=5.5cm]{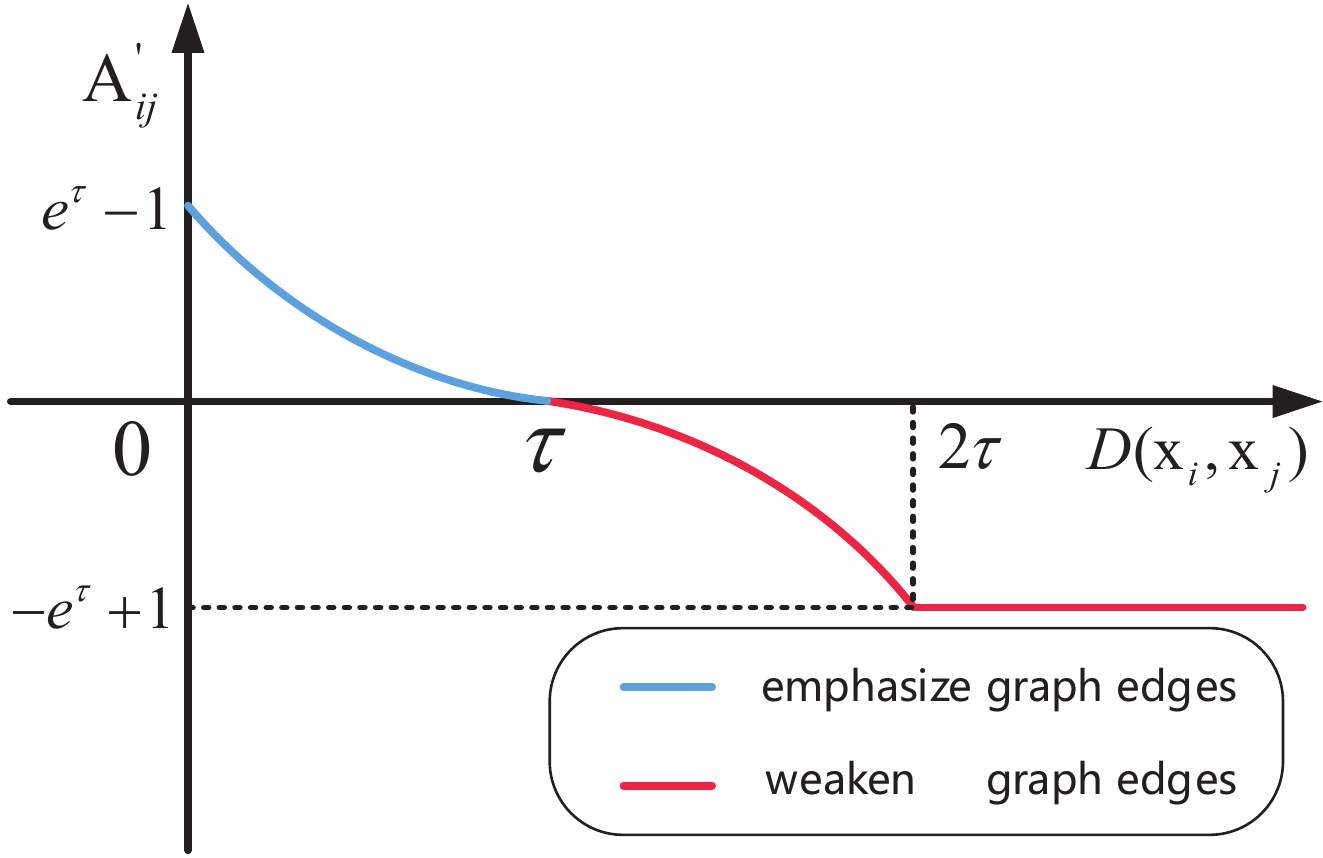}
  \caption{Visual explanation of Eq.~\eqref{eq3}. The horizontal axis corresponds to $D({{ \mathbf{x}}_{i}},{{ \mathbf{x}}_{j}})$ and the vertical axis corresponds to $\mathbf{A}_{ij}^{\prime}$. When $\mathbf{A}_{ij}^{\prime} > 0$ (drawn in blue), the original graph edges will be emphasized as the value of $\mathbf{A}_{ij}^{\prime}$ is positive; when $\mathbf{A}_{ij}^{\prime} < 0$ (drawn in red), the original graph edges will be weakened as the value of $\mathbf{A}_{ij}^{\prime}$ is negative.}
  \label{fig_eq}
  \end{figure}

First, we construct a generalized Mahalanobis distance $D$ to measure the distance between each pair of nodes, which is formulated as
\begin{equation}\label{eq2}
D\left(\mathbf{x}_{i}, \mathbf{x}_{j} \right)=\sqrt{{{\left( \mathbf{x}_{i}-\mathbf{x}_{j} \right)}^{\top }}{\mathbf{{W}}_\text{D}}\mathbf{W}_\text{D}^{\top}\left( \mathbf{x}_{i}-\mathbf{x}_{j} \right)},
\end{equation}
where $\mathbf{W}_\text{D}$ denotes a trainable weight matrix. This distance could be utilized to estimate the edge influence between each pair of nodes. Specifically, a small distance $D(\mathbf{x}_{i}, \mathbf{x}_{j})$ often corresponds to a minor edge influence of the edge between $\mathbf{x}_{i}$ and $\mathbf{x}_{j}$. Anchored on Eq.~\eqref{eq2}, to adaptively emphasize or weaken the edges, we modify the original adjacency matrix $\mathbf{A}$ by adding an auxiliary matrix $\mathbf{A}^{\prime}$, which can be calculated as
\begin{equation}\label{eq3}
\mathbf{A}_{ij}^{\prime}=\left\{\begin{array}{lc}-\min ({{e}^{D({{ \mathbf{x}}_{i}},{{ \mathbf{x}}_{j}})-\tau}},e^{\tau} )+1,&\text{if}~D({{ \mathbf{x}}_{i}},{{ \mathbf{x}}_{j}})> \tau\\ {{e}^{-D({{ \mathbf{x}}_{i}},{{ \mathbf{x}}_{j}})+\tau}-1},&\text{if}~D({{ \mathbf{x}}_{i}},{{ \mathbf{x}}_{j}}) \leq \tau\end{array},\right.
\end{equation}
where $\mathbf{x}_{i} \in \mathcal{N}( {{\mathbf{x}}_{j}})$ or $\mathbf{x}_{j} \in \mathcal{N}( {{\mathbf{x}}_{i}})$, $\tau > 0$ is a learnable parameter, and the truncation parameter $e^{\tau}$ is used to avoid excessively large values. Visual explanation of Eq.~\eqref{eq3} is shown in Fig.~\ref{fig_eq}, where $\tau$ is a threshold. As can be observed in Fig.~\ref{fig_eq}, when $D({{ \mathbf{x}}_{i}},{{ \mathbf{x}}_{j}}) \leq \tau$, the edge between $\mathbf{x}_{i}$ and $\mathbf{x}_{j}$ may have a large influence on the graph and thus could be emphasized. In contrast, when $D({{ \mathbf{x}}_{i}},{{ \mathbf{x}}_{j}}) > \tau$, the corresponding edge could be weakened, considering its minor influence on the graph. In addition, the function curve is truncated to $-e^{\tau}+1$ when $D({{ \mathbf{x}}_{i}},{{ \mathbf{x}}_{j}}) > 2\tau$, to avoid excessively small values. To construct a randomly augmented graph view, we then generate a random masking matrix $\tilde{\mathbf{R}} \in\{0,1\}^{n \times n}$. Here, the element $\tilde{\mathbf{R}}_{ij}=\tilde{\mathbf{R}}_{ji}$ is sampled from a Bernoulli distribution $B(1, p_\text{sample})$ if $i \neq j$, and the value of each diagonal element $\tilde{\mathbf{R}}_{ii}$ is fixed to one. The hyperparameter $p_\text{sample}$ denotes the probability of performing edge emphasizing or weakening. Afterwards, the adjacency matrix $\tilde{\mathbf{A}}$ of the augmented graph can be computed as
\begin{equation}\label{eq3_1}
\tilde{\mathbf{A}}=\mathbf{A}  + \mathbf{A}^{\prime}\circ \tilde{\mathbf{R}},
\end{equation}
where $\circ$ denotes Hadamard product. By using Eq.~\eqref{eq3_1}, the spatial-level graph augmentation can be performed on the randomly sampled edges. Next, for $\mathbf{x}_{i}$ and $\mathbf{x}_{j}$ satisfying $\mathbf{x}_{i} \in \mathcal{N}( {{\mathbf{x}}_{j}})$ or $\mathbf{x}_{j} \in \mathcal{N}( {{\mathbf{x}}_{i}})$, the corresponding element $\tilde{\mathbf{A}}_{ij}$ can be normalized to $[0,1]$ by
\begin{equation}\label{eq3_2}
\tilde{\mathbf{A}}_{ij} \leftarrow\frac{\tilde{\mathbf{A}}_{ij} - \tilde{\mathbf{A}}_\text{min}}{\tilde{\mathbf{A}}_\text{max} - \tilde{\mathbf{A}}_\text{min}},
\end{equation}
where $\tilde{\mathbf{A}}_\text{min}$ and $\tilde{\mathbf{A}}_\text{max}$ are the minimum and the maximum values of $\tilde{\mathbf{A}}_{ij}$, respectively. With Eq.~\eqref{eq3_2}, we can make all the values of $\tilde{\mathbf{A}}$ no less than 0. Therefore, by employing the spatial-level augmentation, the graph edges can be adaptively emphasized or weakened according to their corresponding importance.\vspace{-1em}

\subsection{Spectral-Level Graph Augmentation}
\label{Spectral-Level}
Different from the spatial-level graph augmentation that focuses on the spatial relations of image regions in HSI, our spectral-level augmentation aims to perturb the spectral signatures (\textit{i.e.}, node features) for graph augmentation. To achieve this target, \cite{velickovic2019deep} uniformly shuffles node features for graph augmentation, which actually treats all feature dimensions equally and ignores their distinct contributions. Inspired by chromosomal crossover \cite{creighton1931correlation} in biology, we come up with a new augmentation method which exchanges partial dimensions of node features according to the spectral information of HSI.

\begin{figure} \centering   
  \subfigure[]
  {
   \label{fig_spectral_exchange:1}     
  \includegraphics[width=2.5cm]{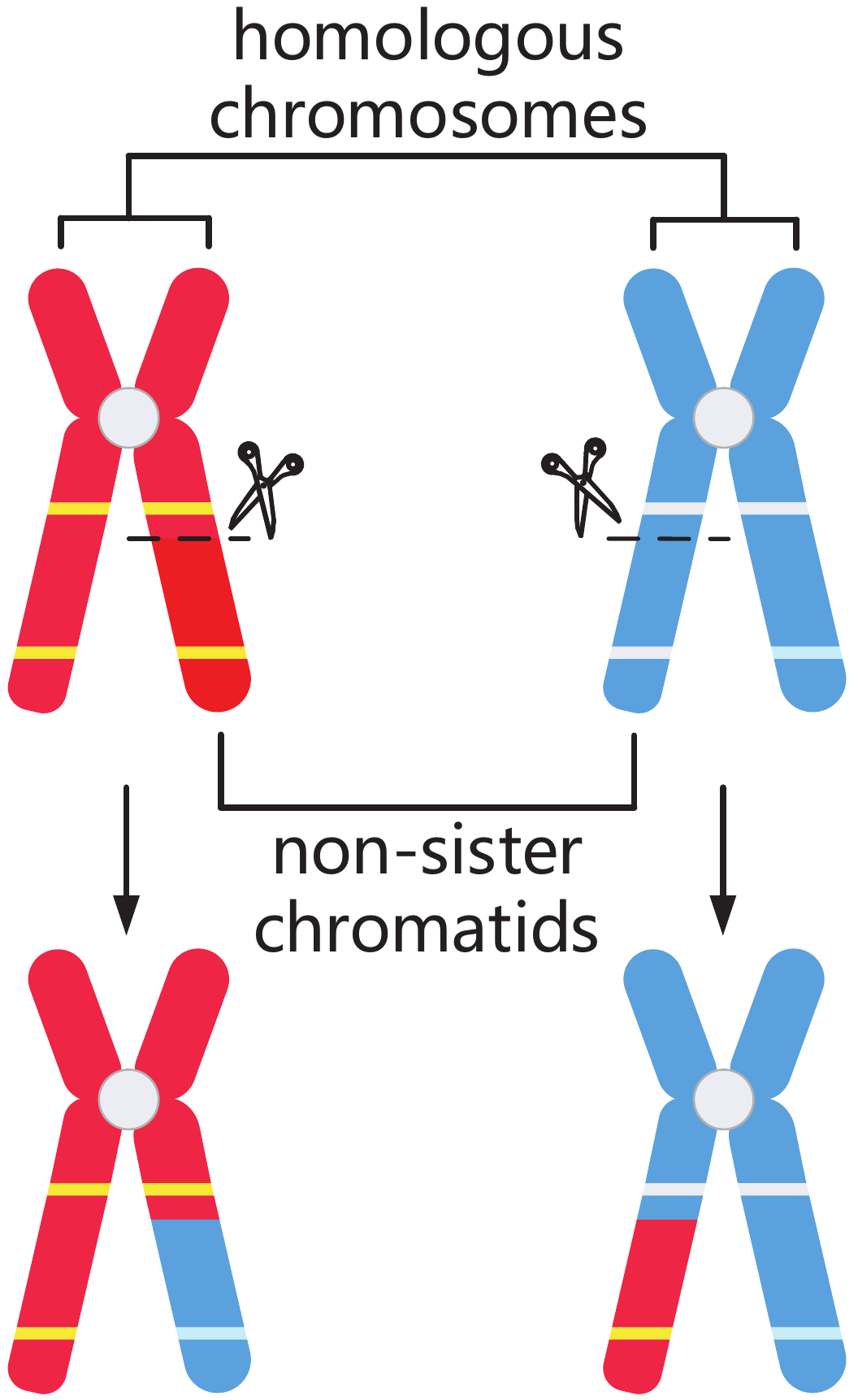}  
  }
  \subfigure[]
  { 
  \label{fig_spectral_exchange:2}
  \includegraphics[width=2.05cm]{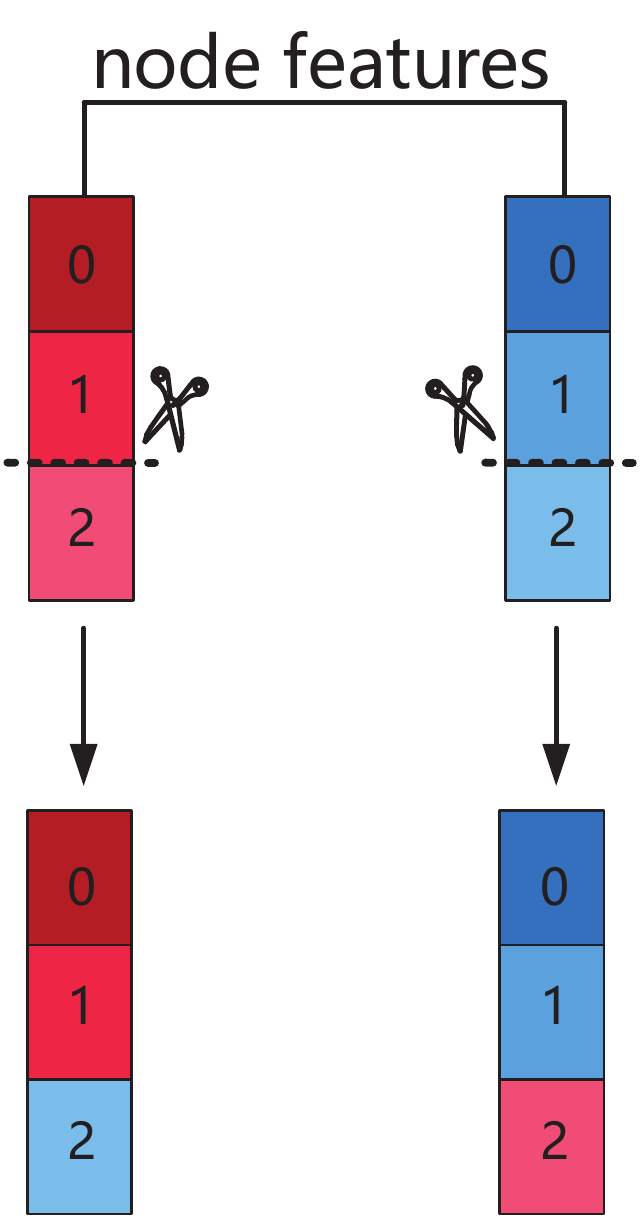}    
  }
  \caption{The analogy of chromosomal crossover and feature exchange. (a) is the illustration of the exchange of genetic material during sexual reproduction between two homologous chromosomes' non-sister chromatids. (b) is the illustration of spectral-level graph augmentation. In (b), the colored small rectangles represent node features, where ``0", ``1", and ``2" in the figure represent the IDs of different dimensions. Here, the features are exchanged at the dimension with ID ``2".\vspace{-1em}}
  \label{fig_spectral_exchange}
  \end{figure}

Chromosomal crossover represents the exchange of genetic material at the stage of sexual reproduction between two homologous chromosomes' non-sister chromatids, which leads to recombinant chromosomes and significantly increases genetic diversity. Schematic diagram of chromosomal crossover is shown in Fig.~\ref{fig_spectral_exchange:1}. For spectral-level graph augmentation, we treat two adjacent nodes as homologous chromosomes' non-sister chromatids and regard their node features as genetic material. Then partial exchange of node features is performed across the graph edges, which is shown in Fig.~\ref{fig_spectral_exchange:2}. The aim of feature exchange is to provide diverse perturbations for contrastive learning and guide the network to obtain improved representation. Notably, the feature exchange is limited in the adjacent nodes (\textit{i.e.}, the nodes connected by the edges of $\mathcal{G}$) to avoid corrupting the essential structure of graphs.

Motivated by \cite{zhu2021graph}, features in influential dimensions are expected to be exchanged with small probability, as they often carry valuable spectral information that is beneficial to accurate classification. In the proposed spectral augmentation, we firstly generate a random vector $\mathbf{m}$ with the $h$-th element $\mathbf{m}_{h}$ ($h=1,2,\dots,d$) drawn from a Bernoulli distribution independently, namely $\mathbf{m}_{h} \sim B(1,1 - p_{h})$, where the parameter ${p}_{h}$ reflects the importance of the $h$-th feature dimension. Afterwards, feature exchange between each pair of nodes will be performed at the $h$-th dimension if $\mathbf{m}_h=1$. Here, the parameter ${p}_{h}$ is obtained by mutual information, since the mutual information $I(\mathbf{X}_{\text{labeled}[:,h]}, \mathbf{Y})$ is able to characterize the relationship between the $h$-th feature dimension of $\mathbf{X}_\text{labeled}$ (\textit{i.e.}, $\mathbf{X}_{\text{labeled}[:,h]}$) and the label matrix $\mathbf{Y}$. The notation $\mathbf{X}_\text{labeled}$ denotes the features of $l$ labeled examples. The notation $\mathbf{Y} \in \mathbb{R}^{l \times c}$ represents the label matrix, where $c$ is the number of land cover categories. The element $\mathbf{Y}_{ij} = 1$ if the $i$-th labeled example belongs to the $j$-th land cover category, and $\mathbf{Y}_{ij} = 0$, otherwise. However, it could be difficult to directly calculate $I(\mathbf{X}_{\text{labeled}[:,h]}, \mathbf{Y})$, as the probability distribution of $\mathbf{X}_{\text{labeled}[:,h]}$ is unavailable. Fortunately, according to~\cite{ross2014mutual}, $I(\mathbf{X}_{\text{labeled}[:,h]}, \mathbf{Y})$ can be estimated by partitioning them into bins of finite size and approximated via the finite sum. As a result, we partition $\mathbf{X}_{\text{labeled}[:,h]}$ and $\mathbf{Y}$ into $N$ bins, where $n_{x}(i)$ and $n_{y}(j)$ represent the number of examples falling into the $i$-th bin of $\mathbf{X}_{\text{labeled}[:,h]}$ and the $j$-th bin of $\mathbf{Y}$, respectively. The notation $n(i,j)$ denotes the number of examples in their intersection. Consequently, $I(\mathbf{X}_{\text{labeled}[:,h]}, \mathbf{Y})$ can be estimated as
\begin{equation}\label{eq4}
I(\mathbf{X}_{\text{labeled}[:,h]}, \mathbf{Y})\approx \sum\nolimits_{i, j} p(i, j) \log \frac{p(i, j)}{p_{x}(i) p_{y}(j)},
\end{equation}
where $p_{x}(i)\approx\frac{n_{x}(i)}{N}$, $p_{y}(j)\approx\frac{n_{y}(j)}{N}$, and $p(i,j)\approx\frac{n(i,j)}{N}$. According to the definition of mutual information, the $h$-th feature dimension is important in determining the land cover category if the value of $I(\mathbf{X}_{\text{labeled}[:,h]}, \mathbf{Y})$ is large. As a result, we can let ${p}_{h} = I(\mathbf{X}_{\text{labeled}[:,h]}, \mathbf{Y})$ ($h=1,2,\dots,d$) after normalizing $I(\mathbf{X}_{\text{labeled}[:,h]}, \mathbf{Y})$ to $[0,1]$.

It is notable that for each graph node, the spectral-level augmentation can be performed no more than once to avoid excessive damage to graph information. Meanwhile, the spectral-level augmentation will be performed on the node pairs in descending order by the value of $\mathbf{A}_{ij}$, with the expectation that feature exchange across similar nodes can probably preserve the original topological information.

\section{Localized and Hierarchical Graph Convolution}
\label{IV-C}
In HSI, the image regions that are far away in the original 2D space may belong to the same land cover category. However, most existing GCN models fail to exploit the long range dependencies among image regions, as they mainly aim at encoding the pairwise importance among local image regions, which could lead to degraded performance. To deal with this issue, we propose to use localized and hierarchical graph convolution, simultaneously, so as to incorporate both local and global contextual information.\vspace{-0.7em}

\subsection{Localized Graph Convolution}
There are various outstanding GCN architectures available to obtain representations from the local view, such as the well-known GCN \cite{kipf2016semi} and graph attention network \cite{velivckovic2017graph}. For simplicity, we select the classical GCN to perform localized graph convolution. The convolution process can be denoted as
\begin{equation}\label{eq7}
\mathbf{Z}^\text{local}=\hat{\mathbf{A}} \sigma\left(\hat{\mathbf{A}} \mathbf{X}_{1} \mathbf{W}^{(0)}\right) \mathbf{W}^{(1)},
\end{equation}
where $\hat{\mathbf{A}}=\tilde{\mathbf{D}}^{-\frac{1}{2}} \bar{\mathbf{A}} \tilde{\mathbf{D}}^{-\frac{1}{2}}$ is leveraged to normalize the adjacency matrix, $\bar{\mathbf{A}}=\mathbf{A}_{1}+\mathbf{I}$ is used to add the self-connection to adjacency matrix, $\mathbf{A}_{1}$ is the adjacency matrix of $\tilde{\mathcal{G}}_{1}$, $\mathbf{I}$ is the identity matrix, and $\tilde{\mathbf{D}}_{i i}=\sum\nolimits_{j} \bar{\mathbf{A}}_{i j}$. In Eq.~\eqref{eq7}, $\mathbf{W}^{(0)}$ and $\mathbf{W}^{(1)}$ indicate the learnable weight matrices, $\sigma(\cdot)$ denotes the activation function (\textit{i.e.}, ReLU \cite{agarap2018deep} in our ConGCN), $\mathbf{X}_{1} \in \mathbb{R}^{n \times d}$ represents the feature matrix of $\tilde{\mathcal{G}}_{1}$, and $\mathbf{Z}^\text{local}$ represents the representations learned from the local view.\vspace{-1em}

\subsection{Hierarchical Graph Convolution}
To incorporate the global contextual information of HSI, we adopt the Hierarchical Graph Convolutional Network (HGCN)~\cite{hu2019hierarchical} to perform hierarchical graph convolution. Generally speaking, HGCN aggregates the nodes with similar structures to a set of hyper-nodes constantly. Therefore, it is able to generate coarsened graphs via successive convolution and enlarge the receptive field. In addition, the symmetric graph refining layers are utilized to reconstruct the original graph topology for node-level representation. By this means, the global contextual information can be gradually encoded via hierarchical graph convolution. Hence, the representations $\mathbf{Z}^\text{global}$ generated from the global view can well complement to $\mathbf{Z}^\text{local}$ and help improve the representation ability of the proposed method.

\section{Loss function}
In this section, we introduce the loss functions employed by our method.\vspace{-1em}
\subsection{Semi-Supervised Contrastive Loss Function}
\label{IV-D}
In the proposed ConGCN, we intend to leverage the abundant spectral information to help guide the model training. Fortunately, contrastive learning can naturally exploit the spectral signatures of HSI to generate representations. However, traditional unsupervised contrastive loss \cite{wang2021understanding} can only utilize unlabeled data for model training, which might ignore the precious label information of HSI. To cope with this issue, we devise a semi-supervised contrastive loss to improve the discriminative power of generated representation. Specifically, it can be divided into two parts, \textit{i.e.}, the unsupervised and supervised contrastive losses, respectively.

Here, the unsupervised loss in the local graph view $\mathcal{L}_\text{uc}^\text{local}(\mathbf{x}_{i})$ can be calculated as Eq.~\eqref{eq10}. Analogously, the unsupervised loss in the global graph view $\mathcal{L}_\text{uc}^\text{global}(\mathbf{x}_{i})$ can be calculated by Eq.~\eqref{eq11} as follows:
\begin{equation}\label{eq10}
\mathcal{L}_\text{uc}^\text{local}({{\mathbf{x}}_{i}})=-\frac{1}{2n}\log \frac{{{e}^{\left\langle \mathbf{z}_{i}^\text{local},\mathbf{z}_{i}^\text{global} \right\rangle }}}{\sum\nolimits_{j=1}^{n}{{{e}^{\left\langle \mathbf{z}_{i}^\text{local},\mathbf{z}_{j}^\text{global} \right\rangle }}}},
\end{equation}
\begin{equation}\label{eq11}
\mathcal{L}_\text{uc}^\text{global}({{\mathbf{x}}_{i}})=-\frac{1}{2n}\log \frac{{{e}^{\left\langle \mathbf{z}_{i}^\text{global},\mathbf{z}_{i}^\text{local} \right\rangle }}}{\sum\nolimits_{j=1}^{n}{{{e}^{\left\langle \mathbf{z}_{i}^\text{global},\mathbf{z}_{j}^\text{local} \right\rangle }}}},
\end{equation}
where $\mathbf{z}_{i}^\text{local}=\mathbf{Z}_{i,:}^\text{local}$ and $\mathbf{z}_{i}^\text{global}=\mathbf{Z}_{i,:}^\text{global}$ indicate the representations of $\mathbf{x}_{i}$ learned from local and global views, respectively, and $\langle \cdot \rangle$ expresses the inner product. The notation $\mathbf{Z}_{i,:}^\text{local}$ and $\mathbf{Z}_{i,:}^\text{global}$ denote the $i$-th row of $\mathbf{Z}^\text{local}$ and $\mathbf{Z}^\text{global}$, respectively.

To exploit the scarce yet valuable class information for model training, the supervised contrastive loss function can be defined as
\begin{equation}\label{eq13}
\mathcal{L}_\text{sc}^\text{local}\left( {{\mathbf{x}}_{i}} \right)=-\frac{1}{2l}\log \frac{\sum\nolimits_{k=1}^{l}{\text{ }}{\mathbbm{1}_{\left[ {{y}_{i}}={{y}_{k}} \right]}}{{e}^{\left\langle \mathbf{z}_{i}^\text{local},\mathbf{z}_{k}^\text{global} \right\rangle }}}{\sum\nolimits_{j=1}^{l}{{{e}^{\left\langle \mathbf{z}_{i}^\text{local},\mathbf{z}_{j}^\text{global} \right\rangle }}}},
\end{equation}
\begin{equation}\label{eq14}
\mathcal{L}_\text{sc}^\text{global}\left( {{\mathbf{x}}_{i}} \right)=-\frac{1}{2l}\log \frac{\sum\nolimits_{k=1}^{l}{\text{ }}{\mathbbm{1}_{\left[ {{y}_{i}}={{y}_{k}} \right]}}{{e}^{\left\langle \mathbf{z}_{i}^\text{global},\mathbf{z}_{k}^\text{local} \right\rangle }}}{\sum\nolimits_{j=1}^{l}{{{e}^{\left\langle \mathbf{z}_{i}^\text{global},\mathbf{z}_{j}^\text{local} \right\rangle }}}},
\end{equation}
where $\mathcal{L}_\text{sc}^\text{local}(\mathbf{x}_{i})$ and $\mathcal{L}_\text{sc}^\text{global}(\mathbf{x}_{i})$ represent the supervised pairwise contrastive losses of $\mathbf{x}_{i}$ in local and global views, respectively, $\mathbbm{1}_{[\cdot]}$ is an indicator function which equals to 1 if the argument inside the bracket holds and 0, otherwise. In Eq.~\eqref{eq13} and Eq.~\eqref{eq14}, $y_{i}$ and $y_{k}$ are the labels of $\mathbf{x}_{i}$ and $\mathbf{x}_{k}$, respectively. Different from the unsupervised contrastive loss in Eq.~\eqref{eq10} and Eq.~\eqref{eq11}, the positive and negative pairs in Eq.~\eqref{eq13} and Eq.~\eqref{eq14} can also be constructed based on the class information. That is to say, the nodes belonging to identical/different class are regarded as positive/negative pair.

In consequence, the proposed semi-supervised contrastive loss function $\mathcal{L}_\text{ssc}$ can be represented as
\begin{equation}\label{eqnew}
\begin{aligned}
\mathcal{L}_\text{ssc}=&\sum\nolimits_{i=1}^{n}{(\mathcal{L}_\text{uc}^\text{local}(\mathbf{x}_{i}) + \mathcal{L}_\text{uc}^\text{global}(\mathbf{x}_{i}))} +\\ & \sum\nolimits_{i=1}^{l}{(\mathcal{L}_\text{sc}^\text{local}(\mathbf{x}_{i}) + \mathcal{L}_\text{sc}^\text{global}(\mathbf{x}_{i}))}.
\end{aligned}
\end{equation}
By minimizing $\mathcal{L}_\text{ssc}$, our proposed ConGCN can enhance the discriminative power of generated representations and further improve the subsequent HSI classification result.\vspace{-0.5em}

\subsection{Graph Generative Loss Function}
\label{IV-E}
In addition to the spectral information exploited by contrastive learning, we also intend to implicitly leverage spatial relations to better guide the representation learning process. Here, the graph generative loss function is designed to explore supervision signals from the spatial relations among image regions, which can lead to enhanced data representations and improved classification results.

Motivated by the generative models~\cite{ma2019flexible}, we create a binary random variable $e_{i j}$ which equals to 1 if there is an edge between $\mathbf{x}_{i}$ and $\mathbf{x}_{j}$, and 0, otherwise. We presume that $e_{i j}$ is conditionally independent, so given $\mathbf{Z}^\text{local}$ and $\mathbf{Z}^\text{global}$, the conditional probability of the input graph $\mathcal{G}$ can be expressed by maximizing the following likelihood estimation:
\begin{equation}\label{eq16}
p\left(\mathcal{G} \mid \mathbf{Z}^\text{local}, \mathbf{Z}^\text{global}\right)=\prod\nolimits_{i, j} p\left(e_{i j} \mid \mathbf{Z}^\text{local}, \mathbf{Z}^\text{global}\right).
\end{equation}
Furthermore, we reasonably assume that the probability of $e_{i j}$ only depends on the representations of $\mathbf{x}_{i}$ and $\mathbf{x}_{j}$ according to \cite{ma2019flexible}. As a result, the conditional probability of $e_{ij}$ can be computed as $p\left(e_{i j} \mid \mathbf{Z}^\text{local}, \mathbf{Z}^\text{global}\right)=p\left(e_{i j} \mid \mathbf{z}_{i}^\text{local}, \mathbf{z}_{j}^\text{global}\right)$.

In the end, the logistic function is utilized to encode the above-mentioned conditional probability, which turns out to be
\begin{equation}\label{eq18}
\begin{aligned}
p\left(\mathcal{G} \mid \mathbf{Z}^\text{local}, \mathbf{Z}^\text{global}\right)&=\prod\nolimits_{i, j} p\left(e_{i j} \mid \mathbf{z}_{i}^\text{local}, \mathbf{z}_{j}^\text{global}\right)
\\&=\prod\nolimits_{i, j} \delta\left(\left[\mathbf{z}_{i}^\text{local}, \mathbf{z}_{j}^\text{global}\right] \mathbf{w}\right),
\end{aligned}
\end{equation}
where $\delta(\cdot)$ denotes the logistic function, $\mathbf{w}$ represents a trainable parameter vector, and $[\cdot, \cdot]$ denotes the concatenation operation. Consequently, the proposed graph generative loss function can be represented as $\mathcal{L}_{\text{g}^{2}}=-p\left(\mathcal{G} \mid \mathbf{Z}^\text{local}, \mathbf{Z}^\text{global}\right)$, which is used by our proposed ConGCN to extract precious yet implicit spatial relations among image regions in HSI.\vspace{-1em}

\subsection{Model Training}
\label{IV-F}
After integrating the graph representation from global and local graph views, the final output of our proposed ConGCN can be computed as $\mathbf{O}=\lambda_\text{local} \mathbf{Z}^\text{local}+\left(1-\lambda_\text{local}\right) \mathbf{Z}^\text{global}$, where $0 < \lambda_\text{local} < 1$ denotes the weight assigned to $\mathbf{Z}^\text{local}$. In addition, the cross-entropy loss function $\mathcal{L}_\text{ce}=-\sum\nolimits_{i=1}^{l} \sum\nolimits_{j=1}^{c} \mathbf{Y}_{i j} \ln \mathbf{O}_{i j}$ is utilized to penalize the label differences between the final output $\mathbf{O}$ and the initially labeled seed superpixels.

Finally, the overall loss function of our proposed ConGCN is shown in Eq.~\eqref{eq22} via assembling the cross-entropy loss function $\mathcal{L}_\text{ce}$, semi-supervised contrastive loss function $\mathcal{L}_\text{ssc}$, and graph generative loss function $\mathcal{L}_{\text{g}^{2}}$, namely
\begin{equation}\label{eq22}
\mathcal{L}=\mathcal{L}_\text{ce} + \lambda_\text{ssc}\mathcal{L}_\text{ssc} + \lambda_{\text{g}^2}\mathcal{L}_{\text{g}^{2}},
\end{equation}
where $\lambda_\text{ssc} >0$ and $\lambda_{\text{g}^2} > 0$ are hyperparameters adjusting the impact of $\mathcal{L}_\text{ssc}$ and $\mathcal{L}_{\text{g}^{2}}$, respectively. The process of the proposed ConGCN is exhibited in Algorithm \autoref{algorithm2}.

\begin{algorithm}[t]
\setstretch{1}
  \caption{The proposed ConGCN algorithm.}  
  \label{algorithm2}
  \begin{algorithmic}[1]
    \Require
      Feature matrix $\mathbf{X}$;
      label matrix $\mathbf{Y}$;
      maximum number of iterations $\mathcal{T}$.
    \Ensure
      Predicted label for each unlabeled graph node.
    \For{$t = 1$ to $\mathcal{T}$}
        \State Generate two augmented graphs $\tilde{\mathcal{G}}_{1}$ and $\tilde{\mathcal{G}}_{2}$;
        \State Perform localized and hierarchical graph convolution to obtain graph representation $\mathbf{Z}^\text{local}$ and $\mathbf{Z}^\text{global}$, respectively;
        \State Compute semi-supervised contrastive loss function $\mathcal{L}_\text{ssc}$ via Eq.~\eqref{eqnew};
        \State Compute graph generative loss function $\mathcal{L}_{\text{g}^{2}}$;
        \State Compute cross-entropy loss $\mathcal{L}_\text{ce}$;
        \State Update network parameters via operating back propagation according to overall loss function $\mathcal{L}$ in Eq.~\eqref{eq22};
        \State $t:=t+1$;
    \EndFor
    \State Predict labels based on the trained network.
  \end{algorithmic}
\end{algorithm}

\section{Experiments}
\label{Experiments}
To demonstrate the effectiveness of our proposed ConGCN, intensive experiments are conducted on four well-known HSI datasets, namely \textit{Indian Pines}, \textit{University of Pavia}, \textit{Salinas}, and \textit{Houston University} datasets, of which the details will be introduced in appendix. Concretely, we first compare our proposed ConGCN with other state-of-the-art methods using four metrics according to~\cite{wan2019multiscale, wan2020hyperspectral, wan2021dual}, namely per-class accuracy, overall accuracy (OA), average accuracy (AA), and kappa coefficient. After that, ablative experiments are carried out to verify the effectiveness of semi-supervised contrastive loss function, graph generative loss function, and our proposed graph augmentation technique.\vspace{-0.9em}

\subsection{Experimental Settings}
In our experiments, the proposed ConGCN is implemented by TensorFlow with Adam optimizer. For each dataset, 30 labeled pixels (\textit{i.e.}, examples) of HSI are randomly selected in each land cover category for training. If there are less than 30 examples, only 15 labeled examples are chosen for the corresponding land cover category. In the training phase, 90\% of the labeled examples are utilized to train the network parameters and 10\% of the labeled examples are leveraged as validation set to fine-tune the hyperparameters (\textit{e.g.}, $\lambda_\text{local}$ and $\lambda_\text{ssc}$). All unlabeled examples are used for testing to evaluate the classification performance. The network architecture of our proposed ConGCN is kept identical for all datasets. Moreover, the learning rate and the maximum number of iterations are set to 0.01 and 4000, respectively.

To demonstrate the effectiveness of our proposed ConGCN method, other state-of-the-art HSI classification methods are also utilized for comparison. Concretely, we adopt three GCN-based methods, namely, dual-level deep Spatial Manifold Representation (SMR) network~\cite{wang2021toward}, Multilevel Superpixel Structured Graph U-net (MSSGU)~\cite{liu2021multilevel}, and Superpixel Graph Learning (SGL)~\cite{sellars2020superpixel}. Besides, one contrastive learning based method~\cite{9664575} termed ``Self-Supervised Contrastive Learning" (SSCL) is employed for comparison. In addition, we incorporate two CNN-based methods, namely, Attention-based Adaptive Spectral-Spatial Kernel (A$^{2}$S$^{2}$K)~\cite{roy2020attention} and Adaptive Spectral-Spatial Multiscale Network (ASSMN)~\cite{wang2020adaptive1}. The compared baseline methods also include one generative based method, \textit{i.e.}, Adaptive Dropblock-enhanced Generative Adversarial Network (ADGAN)~\cite{wang2020adaptive}. Besides, the proposed ConGCN is compared with two traditional HSI classification methods, namely, Multiple Feature Learning (MFL)~\cite{li2014multiple} as well as Joint collaborative representation and SVM with Decision Fusion (JSDF)~\cite{bo2015hyperspectral}. All methods are repeated ten times on the four datasets, where the mean accuracies and standard deviations are also reported.\vspace{-1em}

\subsection{Experimental Results}
\label{ExperimentalResults}
To evaluate the performance of our proposed ConGCN, ConGCN is compared with the above-mentioned baseline methods in both quantitative and qualitative aspects on the four datasets.

\subsubsection{Results on \textit{Indian Pines} Dataset}
The quantitative results obtained by different methods on the \textit{Indian Pines} dataset are summarized in \autoref{table5}. We observe that the proposed ConGCN achieves the top-level performance among all the methods in terms of OA and Kappa coefficient. The standard deviations are relatively small as well.  Note that the proposed method achieves 100\% accuracies on five land cover categories (\textit{i.e.}, ${\rm{ID}}=\text{4, 7, 8, 9, and 13}$), and can generally acquire stable and high classification accuracies on the remaining categories. Therefore, it is reasonable to infer that the proposed ConGCN is more stable and effective than other compared methods. The AA of our ConGCN is slightly lower than that of MSSGU, as MSSGU can adaptively incorporate suitable features on different level graphs for various land cover categories. Due to the instability of Generative Adversarial Networks (GAN) during training~\cite{qin2020training}, ADGAN has large standard deviations. Although SSCL employs contrastive learning, its classification performance is far from perfect, especially in the land cover categories with ${\rm{ID}}=\text{1, 4, 7, 9, 15, and 16}$. This is because SSCL only utilizes unlabeled examples to pre-train the encoder and simply fine-tunes the network with few labeled examples. Different from SSCL, our proposed ConGCN utilizes the unlabeled data and the available class information, simultaneously, via using a semi-supervised contrastive loss. As a result, our proposed method outperforms the baseline methods.

\begin{figure*} \centering    
    \subfigure[]
    {
     \label{fig9:a}     
    \includegraphics[width=2cm]{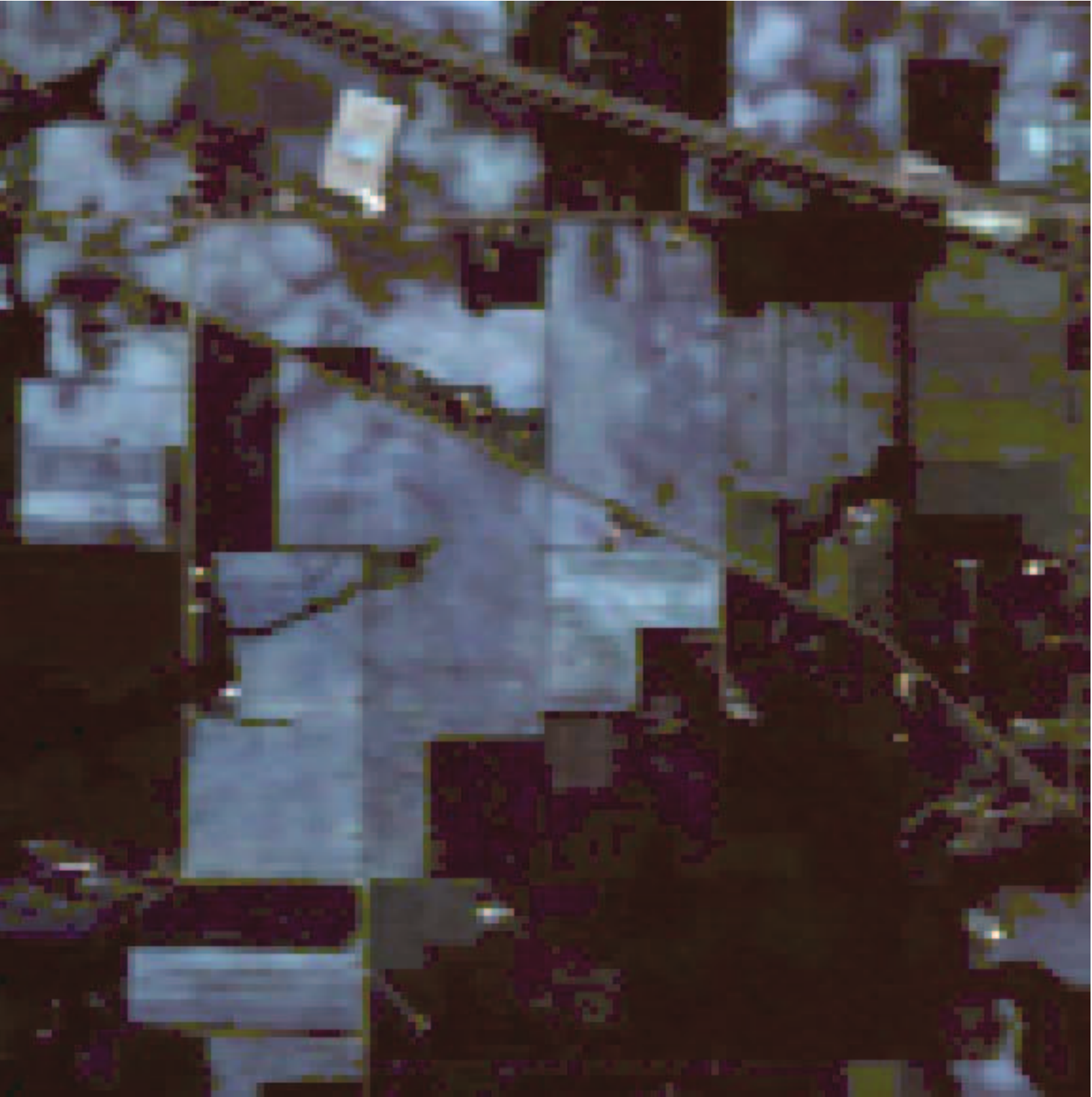}  
    }
    \subfigure[]
    { 
    \label{fig9:b}
    \includegraphics[width=2cm]{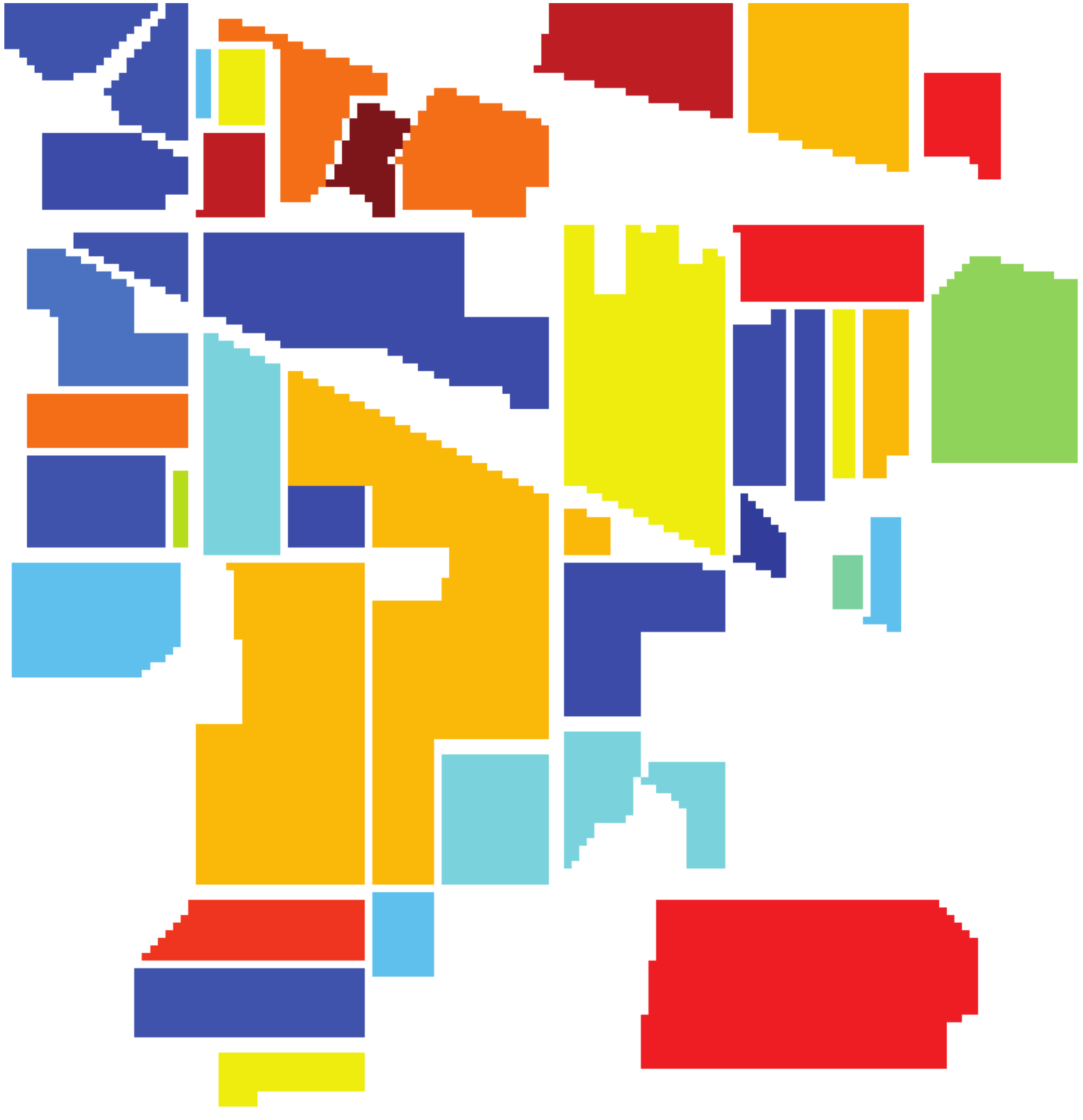}     
    }    
    \subfigure[]
    { 
    \label{fig9:c}     
    \includegraphics[width=2cm]{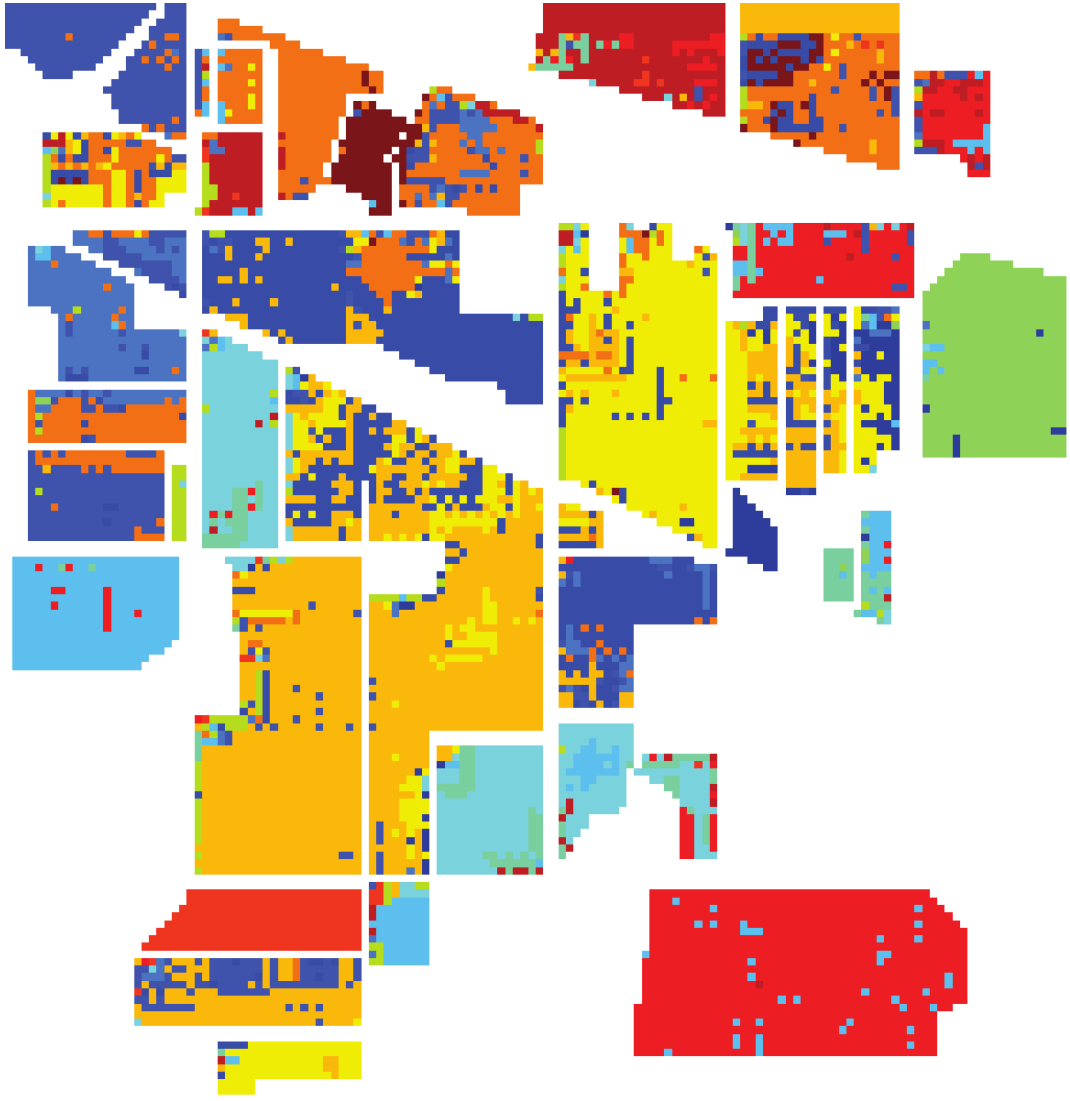}     
    }
    \subfigure[]
    { 
    \label{fig9:d}     
    \includegraphics[width=2cm]{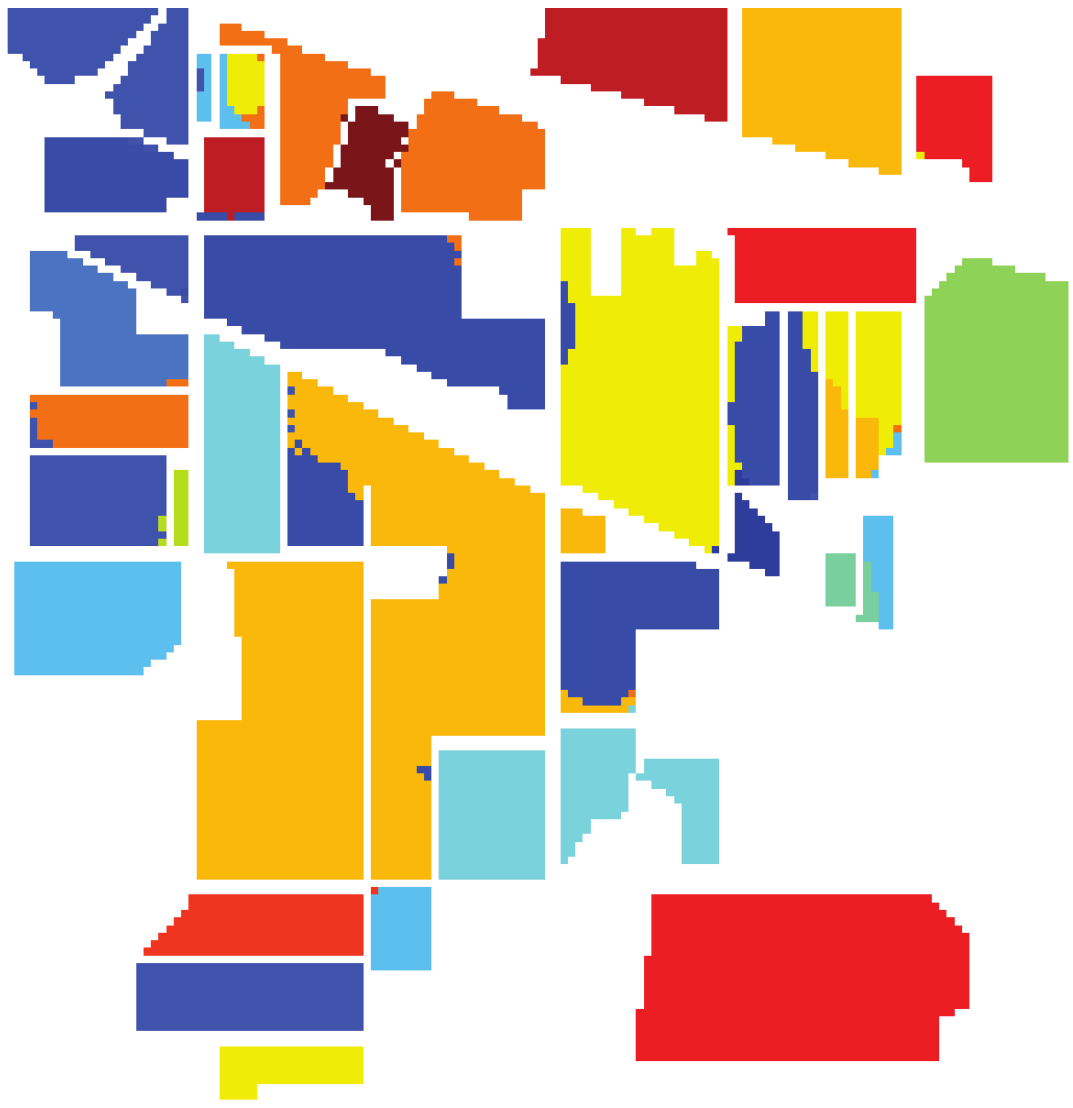}     
    }
    \subfigure[]
    { 
    \label{fig9:e}     
    \includegraphics[width=2cm]{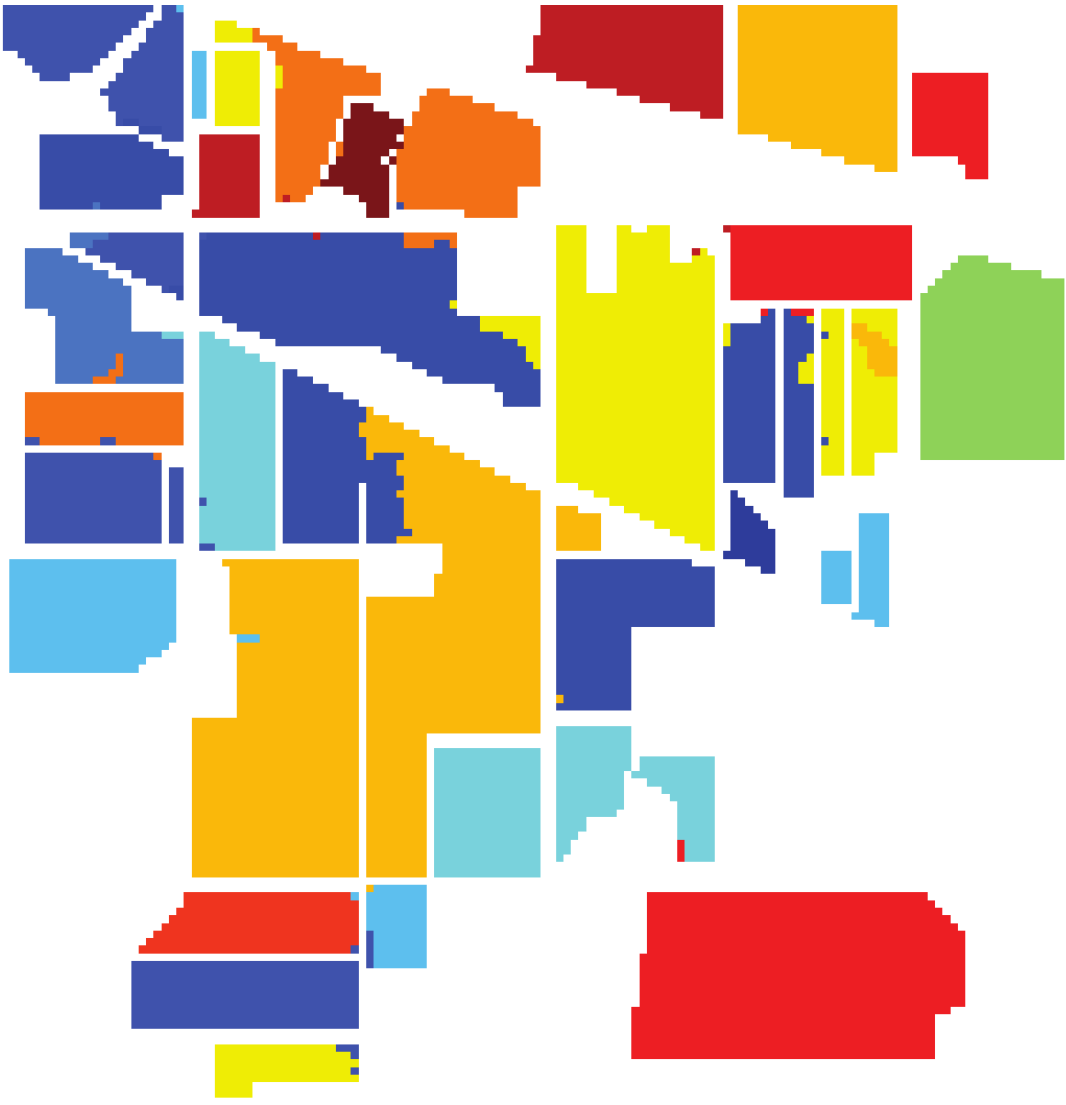}     
    }
    \subfigure[]
    { 
    \label{fig9:f}     
    \includegraphics[width=2cm]{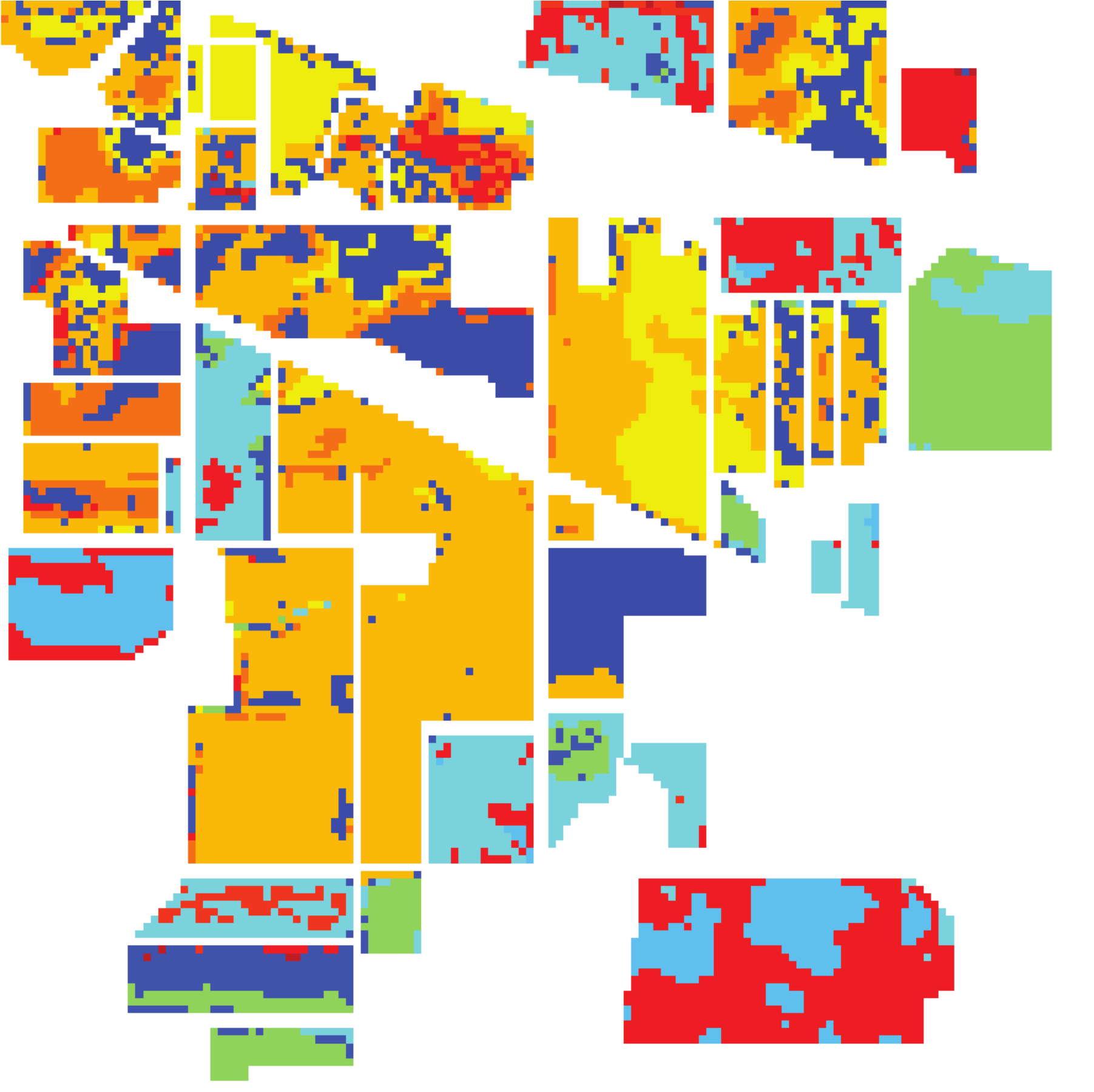}     
    }
    \subfigure[]
    { 
    \label{fig9:g}     
    \includegraphics[width=2cm]{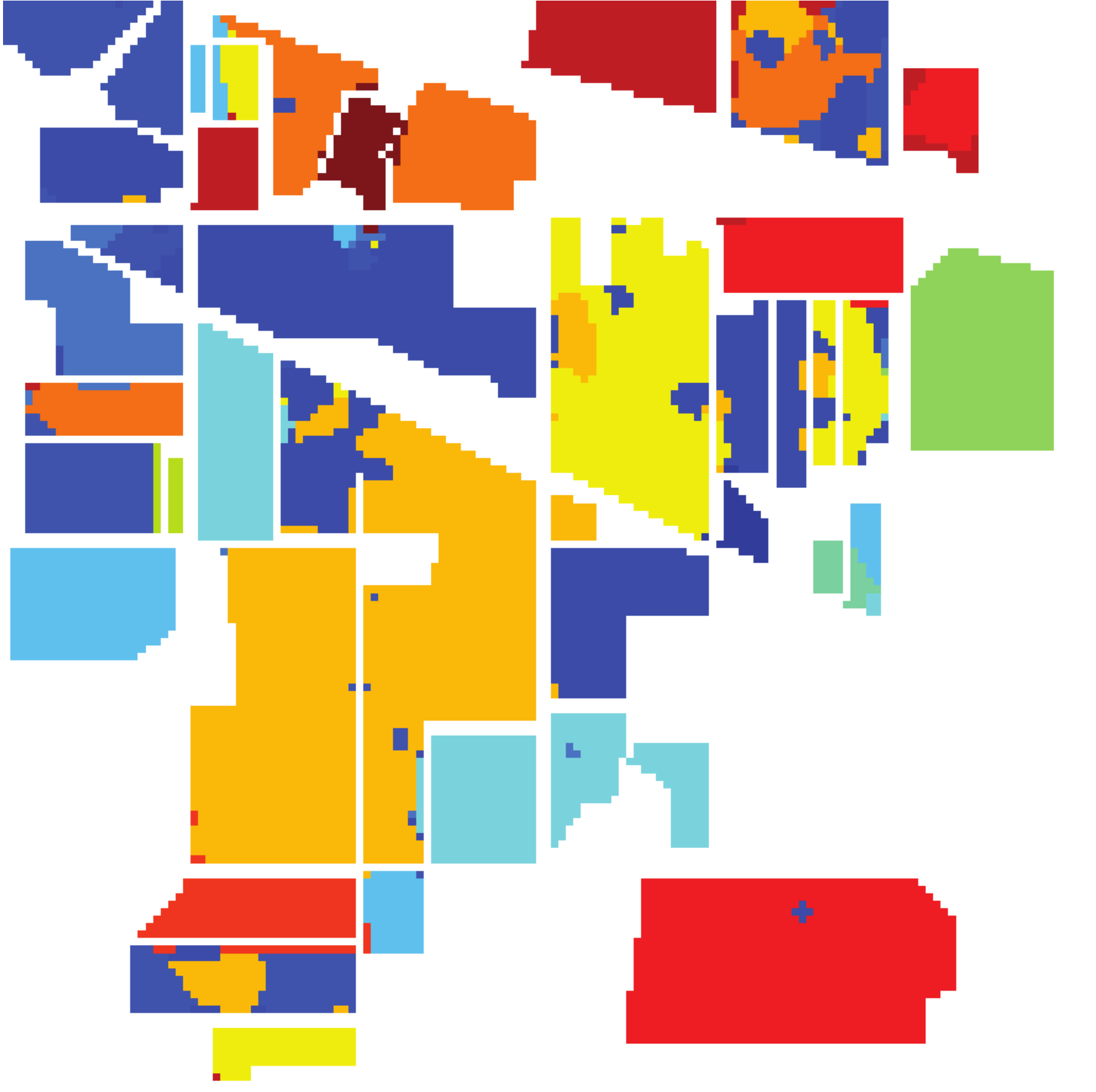}     
    }
    \subfigure[]
    { 
    \label{fig9:h}     
    \includegraphics[width=2cm]{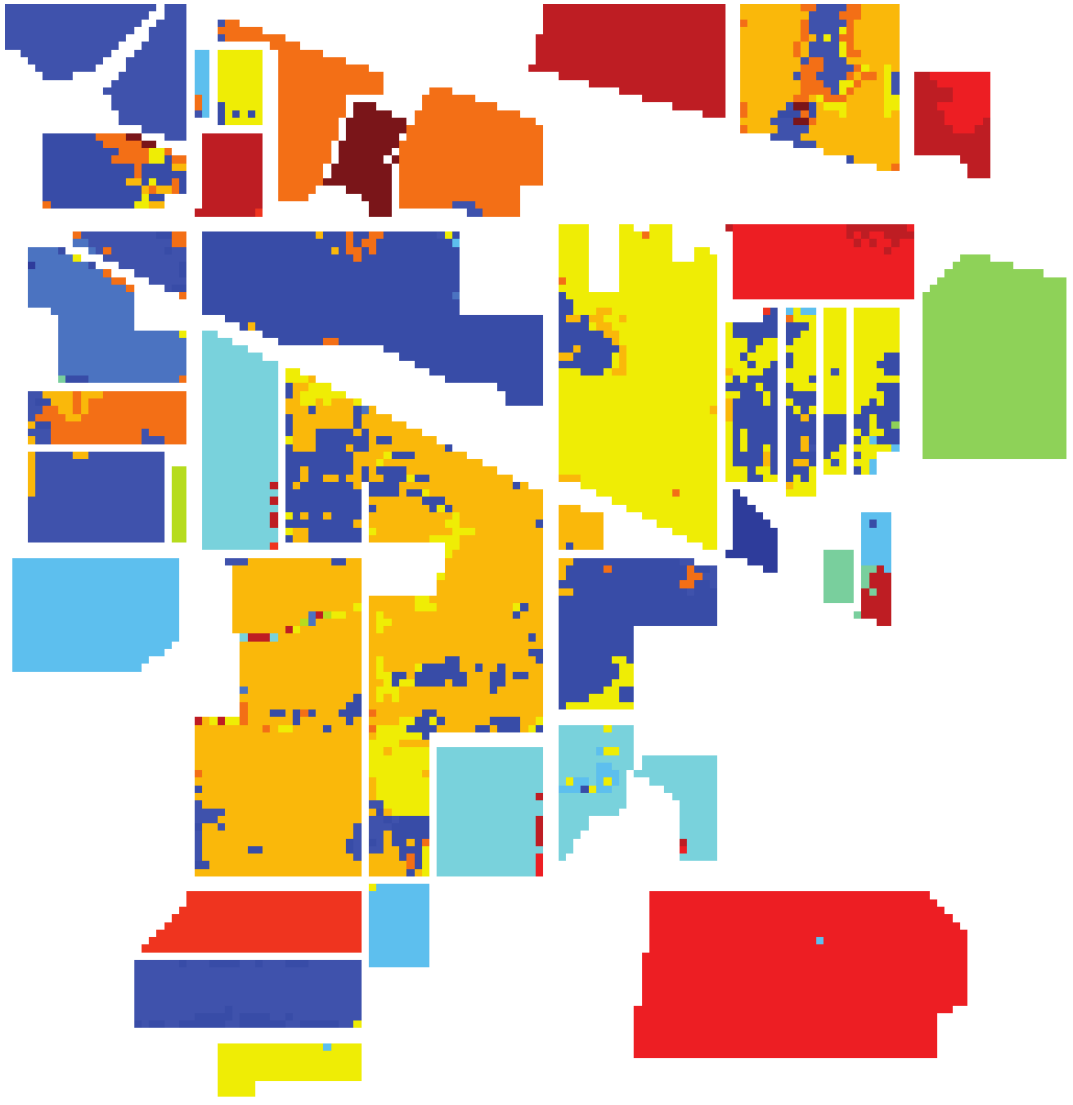}     
    }
    \subfigure[]
    { 
    \label{fig9:i}     
    \includegraphics[width=2cm]{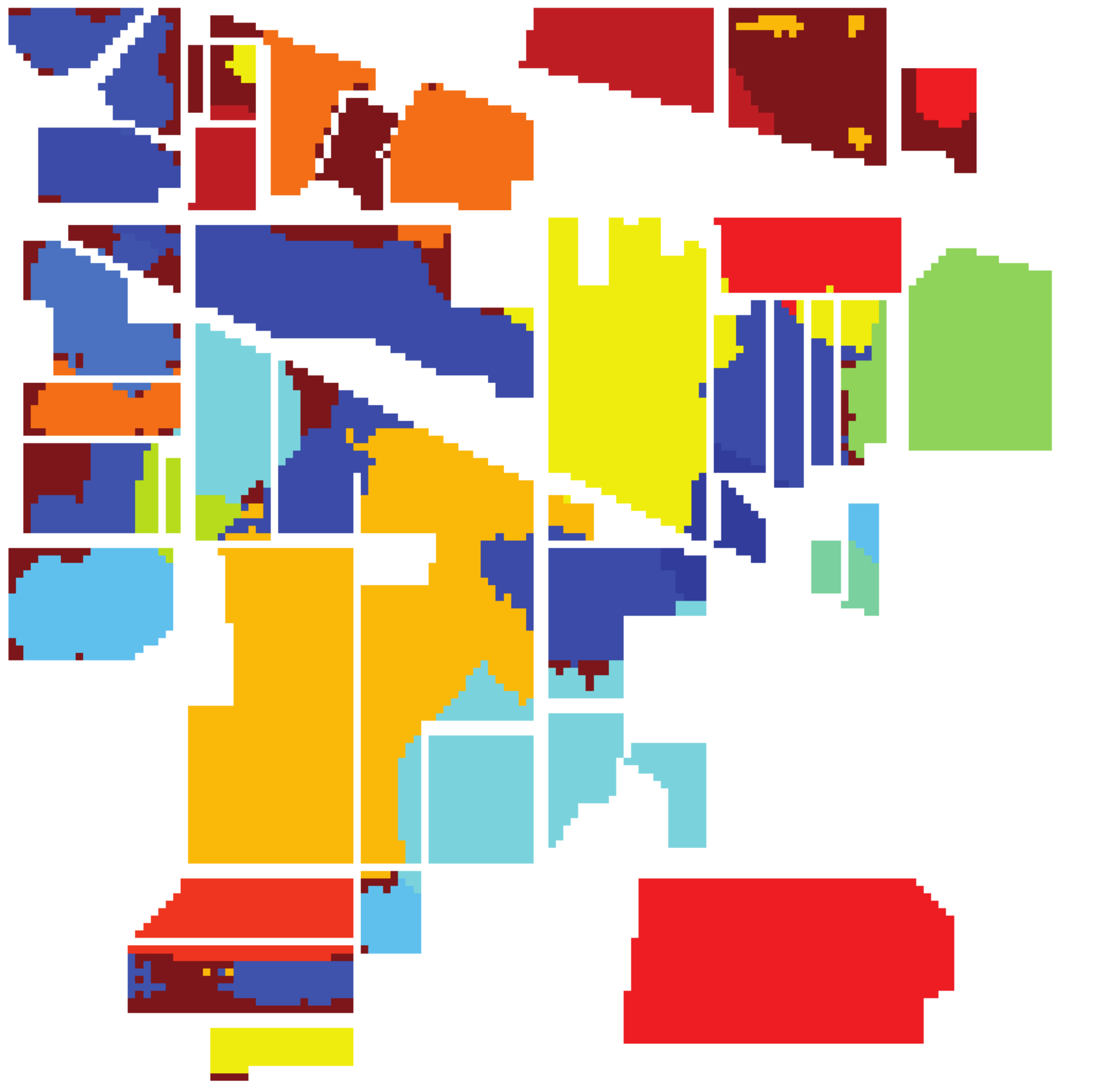}     
    }
    \subfigure[]
    { 
    \label{fig9:j}     
    \includegraphics[width=2cm]{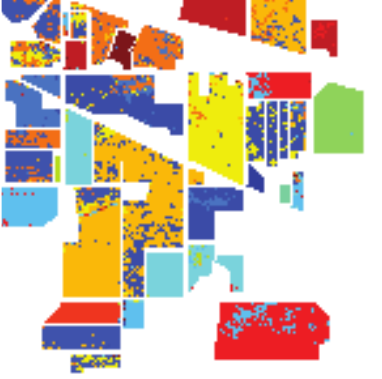}     
    }
    \subfigure[]
    { 
    \label{fig9:k}     
    \includegraphics[width=2cm]{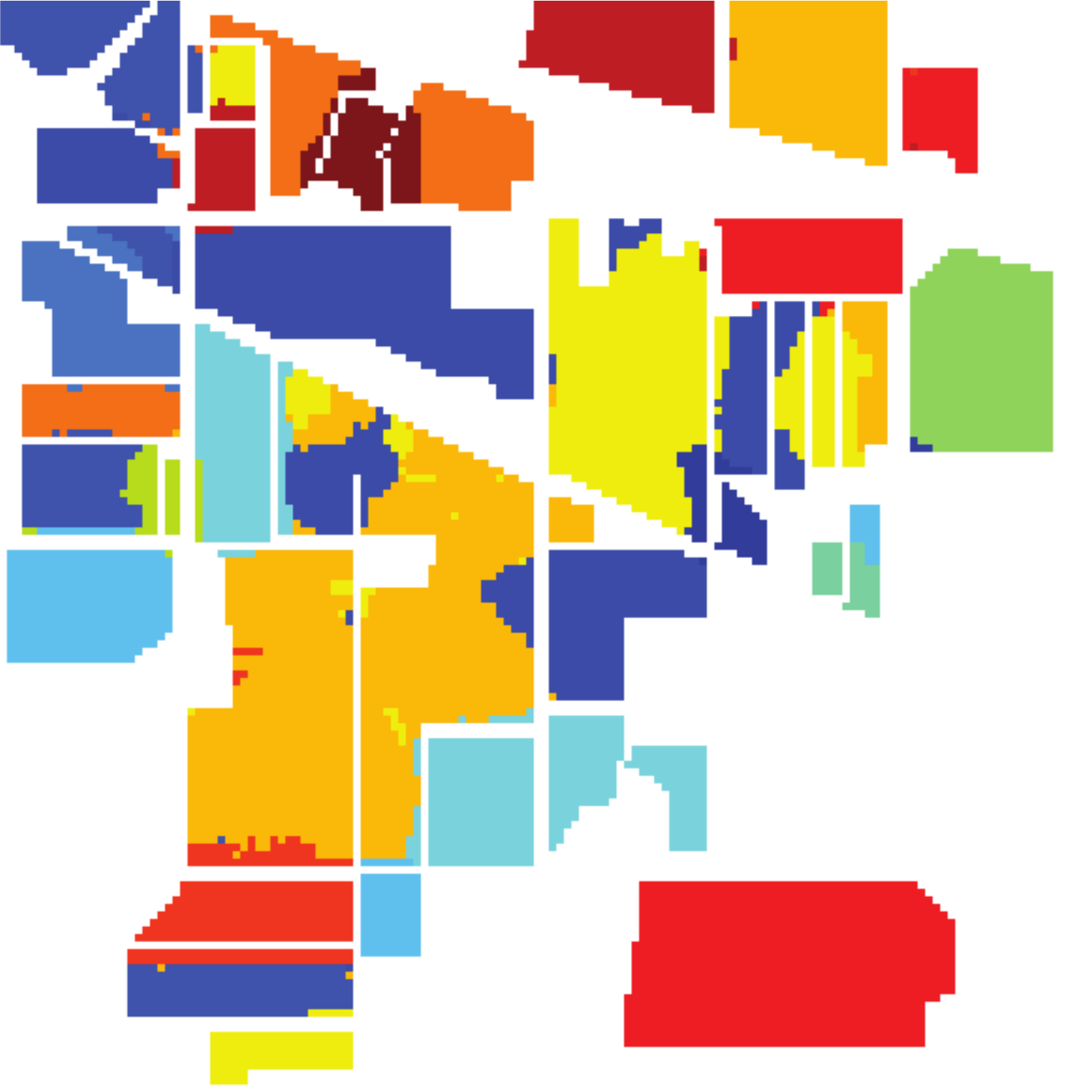}     
    }
    \subfigure[]
    { 
    \label{fig9:l}     
    \includegraphics[width=2cm]{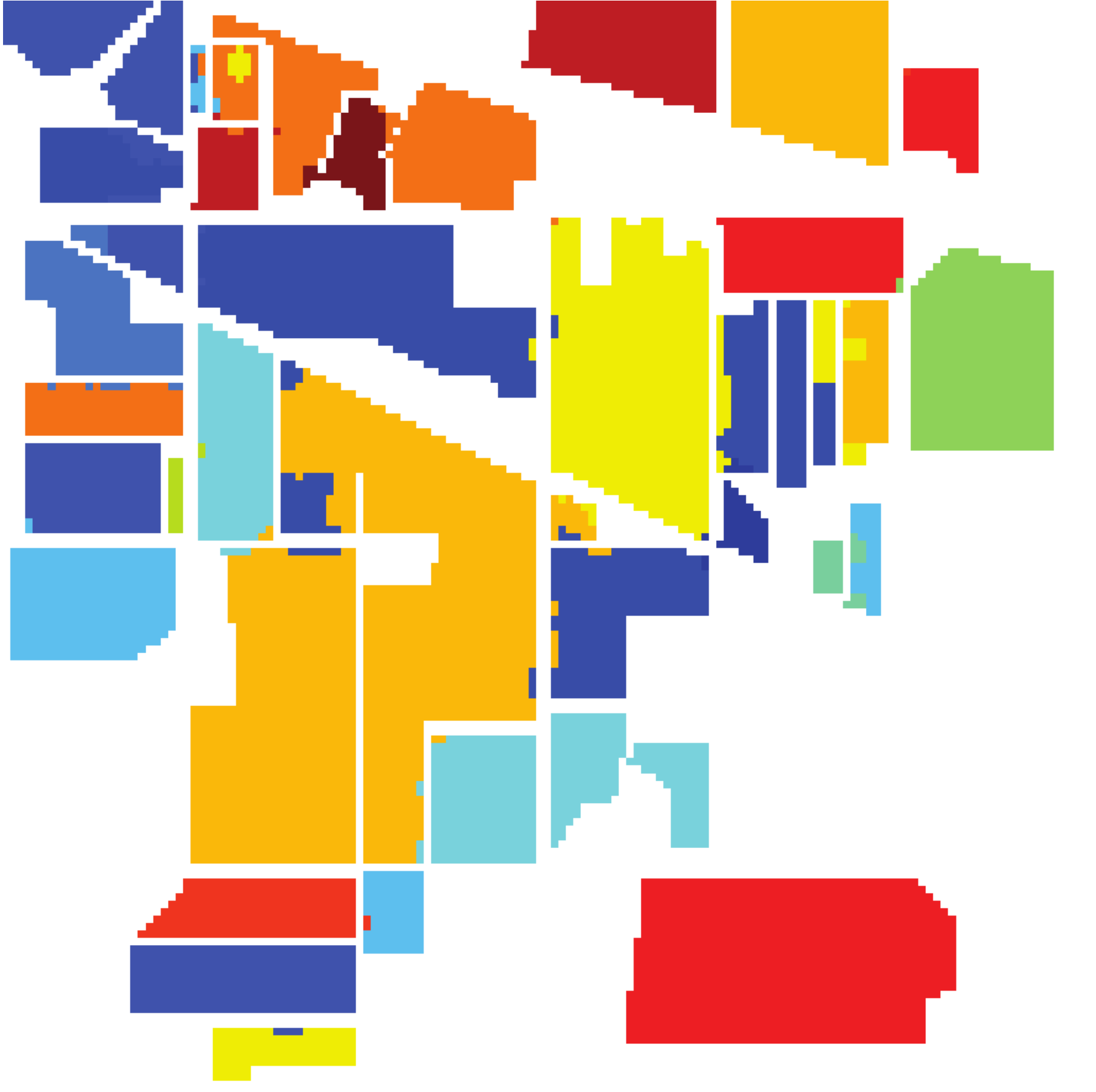}     
    }
    \caption{Classification maps obtained by different methods on \textit{Indian Pines} dataset. (a) False-color image. (b) Ground-truth map. (c) SMR. (d) MSSGU. (e) SGL. (f) SSCL. (g) A$^{2}$S$^{2}$K. (h) ASSMN. (i) ADGAN. (j) MFL. (k) JSDF. (l) ConGCN (Proposed).}     
    \label{fig9}
    \end{figure*}

Fig. \ref{fig9} shows a visual comparison of the classification maps generated by different methods on the \textit{Indian Pines} dataset, where the ground-truth map is exhibited in Fig.~\ref{fig9:b}. The classification maps obtained by SMR (Fig.~\ref{fig9:c}), ASSMN (Fig.~\ref{fig9:h}), and MFL (Fig.~\ref{fig9:j}) suffer from pepper-noise-like mistakes within multiple areas. For instance, in the classification maps obtained by SMR (Fig.~\ref{fig9:c}), ASSMN (Fig.~\ref{fig9:h}), and MFL (Fig.~\ref{fig9:j}), the middle parts of them are highly confusing. Comparatively, the classification map of the proposed ConGCN method yields a smoother visual effect and shows fewer misclassifications than other compared methods.

\begin{table*}[t]
  \scriptsize
  \centering
  \caption{Per-class accuracy, OA, AA (\%), and Kappa coefficient achieved by different methods on \textit{Indian Pines} dataset. The best and second best records in each row are \textbf{bolded} and \underline{underlined}, respectively.}
  \label{table5}
  \begin{tabular}{c|c|c|c|c|c|c|c|c|c|c}
  \hline
  \hline
  ID    & SMR \cite{wang2021toward} & MSSGU \cite{liu2021multilevel} & SGL \cite{sellars2020superpixel}  & SSCL \cite{9664575}  & A$^{2}$S$^{2}$K \cite{roy2020attention}      & ASSMN \cite{wang2020adaptive1}  & ADGAN \cite{wang2020adaptive}  & MFL \cite{li2014multiple} & JSDF \cite{bo2015hyperspectral}  & ConGCN                   \\ \hline
  1     & \underline{98.91$\pm$3.26}            & \textbf{100.00$\pm$0.00}       & \textbf{100.00$\pm$0.00}          & 00.00$\pm$0.00       & 79.31$\pm$18.46                              & \textbf{100.00$\pm$0.00}       & \textbf{100.00$\pm$0.00}                & 98.06$\pm$0.58            & \textbf{100.00$\pm$0.00}         & 98.75$\pm$3.75 \\
  2     & 63.81$\pm$13.64           & \underline{91.69$\pm$1.52}                 & 89.08$\pm$4.56                    & 43.62$\pm$4.44       & 87.85$\pm$4.01                               & 78.83$\pm$3.79                 & 68.59$\pm$10.54                & 74.72$\pm$0.66            & 90.75$\pm$3.19                   & \textbf{92.07$\pm$1.55}  \\
  3     & 69.32$\pm$15.02           & \textbf{98.35$\pm$0.60}        & 90.00$\pm$3.24                    & 32.84$\pm$9.89       & 86.00$\pm$6.73                               & 83.78$\pm$6.61                 & 58.80$\pm$11.33                & 82.14$\pm$0.70            & 77.84$\pm$3.81                   & \underline{97.50$\pm$0.64}           \\
  4     & 92.36$\pm$11.51           & 98.13$\pm$0.56                 & 97.10$\pm$4.43                    & 0.93$\pm$2.00        & 86.14$\pm$7.12                               & 94.49$\pm$1.15                 & 95.02$\pm$2.98                 & 93.60$\pm$0.55            & \underline{99.86$\pm$0.33}                   & \textbf{100.00$\pm$0.00} \\
  5     & 82.88$\pm$9.25            & \underline{95.92$\pm$0.81}                 & \textbf{97.75$\pm$1.73}           & 29.15$\pm$17.33      & 91.95$\pm$2.83                               & 90.84$\pm$1.57                 & 75.74$\pm$9.50                 & 92.54$\pm$0.43            & 87.20$\pm$2.73                   & 94.50$\pm$1.79           \\
  6     & 87.77$\pm$6.99            & \textbf{99.84$\pm$0.23}        & \underline{99.30$\pm$0.42}                    & 69.05$\pm$9.90       & 95.36$\pm$1.53                               & 90.56$\pm$1.75                 & 91.54$\pm$9.22                 & 98.40$\pm$0.27            & 98.54$\pm$0.28                   & 98.99$\pm$0.04           \\
  7     & 97.14$\pm$8.57            & \textbf{100.00$\pm$0.00}       & 0.00$\pm$0.00                     & 0.00$\pm$0.00        & 47.39$\pm$9.54                               & \textbf{100.00$\pm$0.00}       & \textbf{100.00$\pm$0.00}                & \underline{97.28$\pm$0.45}            & \textbf{100.00$\pm$0.00}         & \textbf{100.00$\pm$0.00} \\
  8     & 95.29$\pm$8.04            & \textbf{100.00$\pm$0.00}       & \textbf{100.00$\pm$0.00}          & 88.28$\pm$13.10      & \underline{99.88$\pm$0.28}                               & \textbf{100.00$\pm$0.00}       & 94.84$\pm$13.74                & 99.82$\pm$0.05            & 99.80$\pm$0.31                   & \textbf{100.00$\pm$0.00} \\
  9     & \underline{98.50$\pm$4.50}            & \textbf{100.00$\pm$0.00}       & 0.00$\pm$0.00                     & 0.00$\pm$0.00        & 30.17$\pm$5.98                               & \textbf{100.00$\pm$0.00}       & \textbf{100.00$\pm$0.00}                & \textbf{100.00$\pm$0.00}  & \textbf{100.00$\pm$0.00}         & \textbf{100.00$\pm$0.00} \\
  10    & 82.90$\pm$12.13           & \textbf{96.26$\pm$1.52}        & 90.50$\pm$6.12                    & 48.85$\pm$14.17      & 67.29$\pm$7.01                               & 89.68$\pm$3.23                 & 82.00$\pm$5.35                 & 84.59$\pm$0.53            & 89.99$\pm$4.24                   & \underline{93.57$\pm$2.86}           \\
  11    & 59.46$\pm$10.33           & 91.54$\pm$0.78                 & 94.76$\pm$3.50                    & 77.03$\pm$4.96       & \textbf{97.21$\pm$1.51}                      & 76.62$\pm$3.82                 & 65.56$\pm$9.75                 & 83.73$\pm$0.39            & 76.75$\pm$5.12                   & \underline{97.12$\pm$1.19}           \\
  12    & 79.45$\pm$16.60           & \textbf{98.47$\pm$0.51}        & 94.33$\pm$2.67                    & 37.67$\pm$13.70      & 83.37$\pm$8.30                               & 91.72$\pm$3.10                 & 83.53$\pm$7.38                 & 83.68$\pm$0.72            & 87.10$\pm$2.82                   & \underline{97.57$\pm$0.87}           \\
  13    & 94.63$\pm$6.37            & \textbf{100.00$\pm$0.00}       & 99.09$\pm$0.38                    & 47.02$\pm$39.63      & 85.05$\pm$4.42                               & \textbf{100.00$\pm$0.00}       & \textbf{100.00$\pm$0.00}                & 99.20$\pm$0.06            & \underline{99.89$\pm$0.36}                   & \textbf{100.00$\pm$0.00} \\
  14    & 93.91$\pm$4.40            & \textbf{99.98$\pm$0.04}        & \underline{99.84$\pm$0.24}                    & 78.35$\pm$8.32       & 99.43$\pm$0.35                               & 94.01$\pm$2.02                 & 91.17$\pm$6.65                 & 96.80$\pm$0.40            & 97.21$\pm$2.78                   & 99.83$\pm$0.02           \\
  15    & 62.62$\pm$15.17           & \textbf{99.77$\pm$0.36}        & \underline{99.58$\pm$0.42}                    & 12.56$\pm$11.33      & 87.97$\pm$5.28                               & 98.46$\pm$1.19                 & 92.89$\pm$0.99                 & 97.86$\pm$0.20            & \underline{99.58$\pm$0.68}                   & 99.41$\pm$0.08           \\
  16    & 96.67$\pm$6.84            & \textbf{100.00$\pm$0.00}       & \textbf{100.00$\pm$0.00}          & 1.94$\pm$3.81        & 85.43$\pm$4.50                               & \textbf{100.00$\pm$0.00}       & \underline{99.05$\pm$1.05}                 & 98.72$\pm$0.35            & \textbf{100.00$\pm$0.00}         & 87.14$\pm$2.18           \\ \hline
  OA    & 75.84$\pm$9.40            & \underline{95.87$\pm$0.11}                 & 94.35$\pm$0.93                    & 55.53$\pm$1.63       & 88.31$\pm$1.51                               & 86.38$\pm$1.78                 & 77.51$\pm$5.53                 & 87.38$\pm$0.12            & 88.34$\pm$1.39                   & \textbf{96.74$\pm$0.50}  \\
  AA    & 84.73$\pm$8.30            & \textbf{98.12$\pm$0.09}        & 84.46$\pm$0.42                    & 35.46$\pm$3.18       & 81.86$\pm$1.38                               & 93.06$\pm$1.00                 & 87.42$\pm$3.36                 & 92.57$\pm$0.10            & 94.03$\pm$0.55                   & \underline{97.28$\pm$0.29}           \\
  Kappa & 72.87$\pm$10.33           & \underline{95.28$\pm$0.13}                 & 93.53$\pm$1.06                    & 48.47$\pm$2.03       & 86.76$\pm$1.67                               & 84.51$\pm$2.02                 & 74.98$\pm$5.93                 & 85.64$\pm$0.14            & 86.80$\pm$1.55                   & \textbf{96.27$\pm$0.57}  \\ \hline
  \hline
  \end{tabular}
  \end{table*}

\subsubsection{Results on \textit{University of Pavia} Dataset}
\autoref{table6} presents the quantitative results of different methods on the \textit{University of Pavia} dataset. Similar to the results on the \textit{Indian Pines} dataset, the results in \autoref{table6} indicate that the proposed ConGCN outperforms the compared methods in terms of OA and Kappa coefficient, which again validates the strength of our proposed contrastive learning-based graph convolution. Besides, it is also notable that the performance of SMR on the \textit{University of Pavia} dataset is better than that on the \textit{Indian Pines} dataset. Since SMR can flexibly capture the variations around irregular boundaries with different convolutional kernels, the advantage of SMR becomes prominent on the datasets containing various boundaries, such as the \textit{University of Pavia} dataset. Moreover, compared with the CNN-based methods (\textit{i.e.}, A$^{2}$S$^{2}$K and ASSMN), the proposed ConGCN increases the OA by 8.15\% and 13.43\%, respectively, which suggests that the spatial relations captured by our ConGCN are more useful than the information characterized by the fixed convolutional kernels of CNN.

\begin{figure*} \centering    
    \subfigure[]
    {
     \label{fig10:a}     
    \includegraphics[width=2cm]{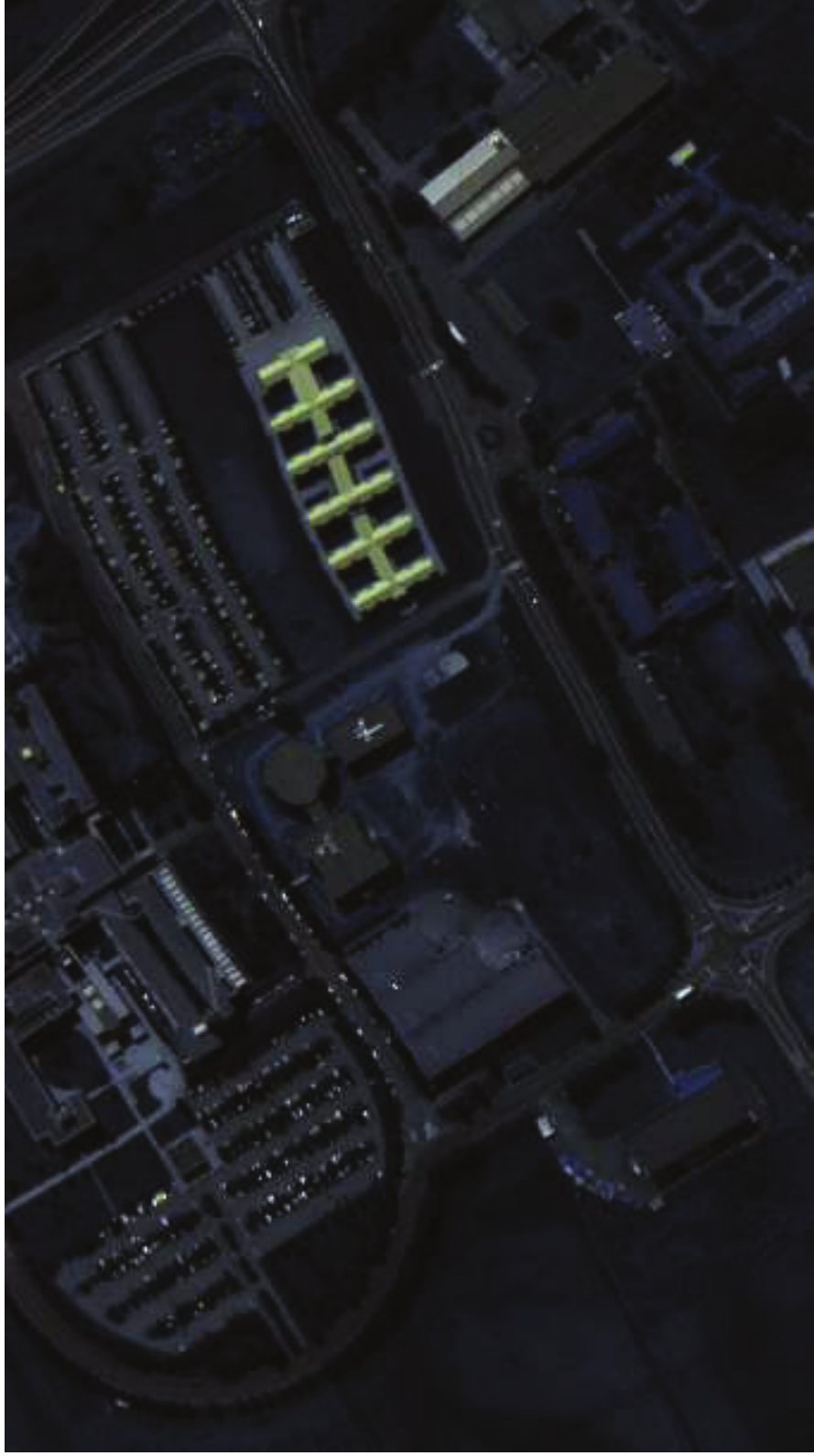}  
    }
    \subfigure[]
    { 
    \label{fig10:b}
    \includegraphics[width=2cm]{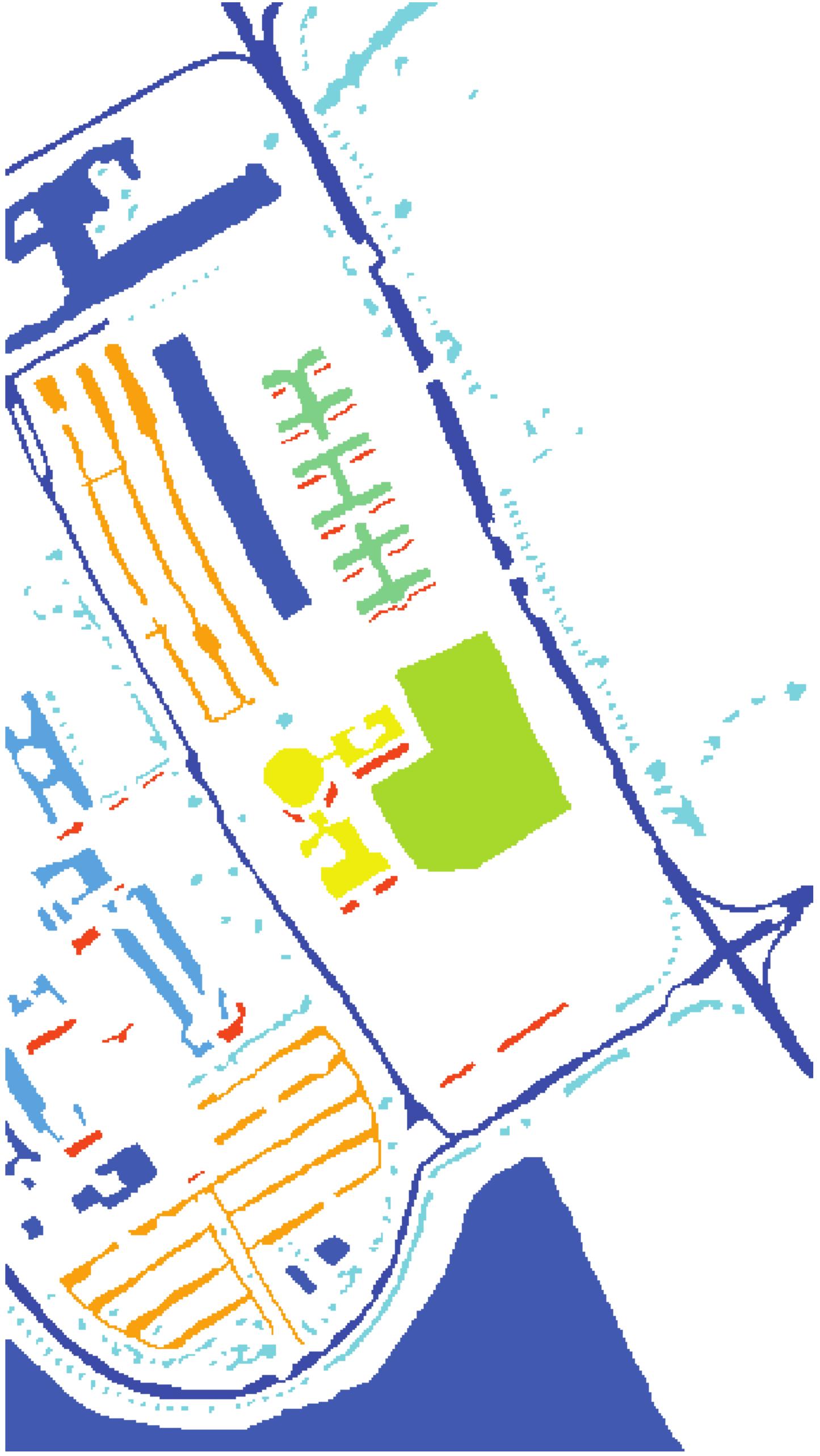}     
    }    
    \subfigure[]
    { 
    \label{fig10:c}     
    \includegraphics[width=2cm]{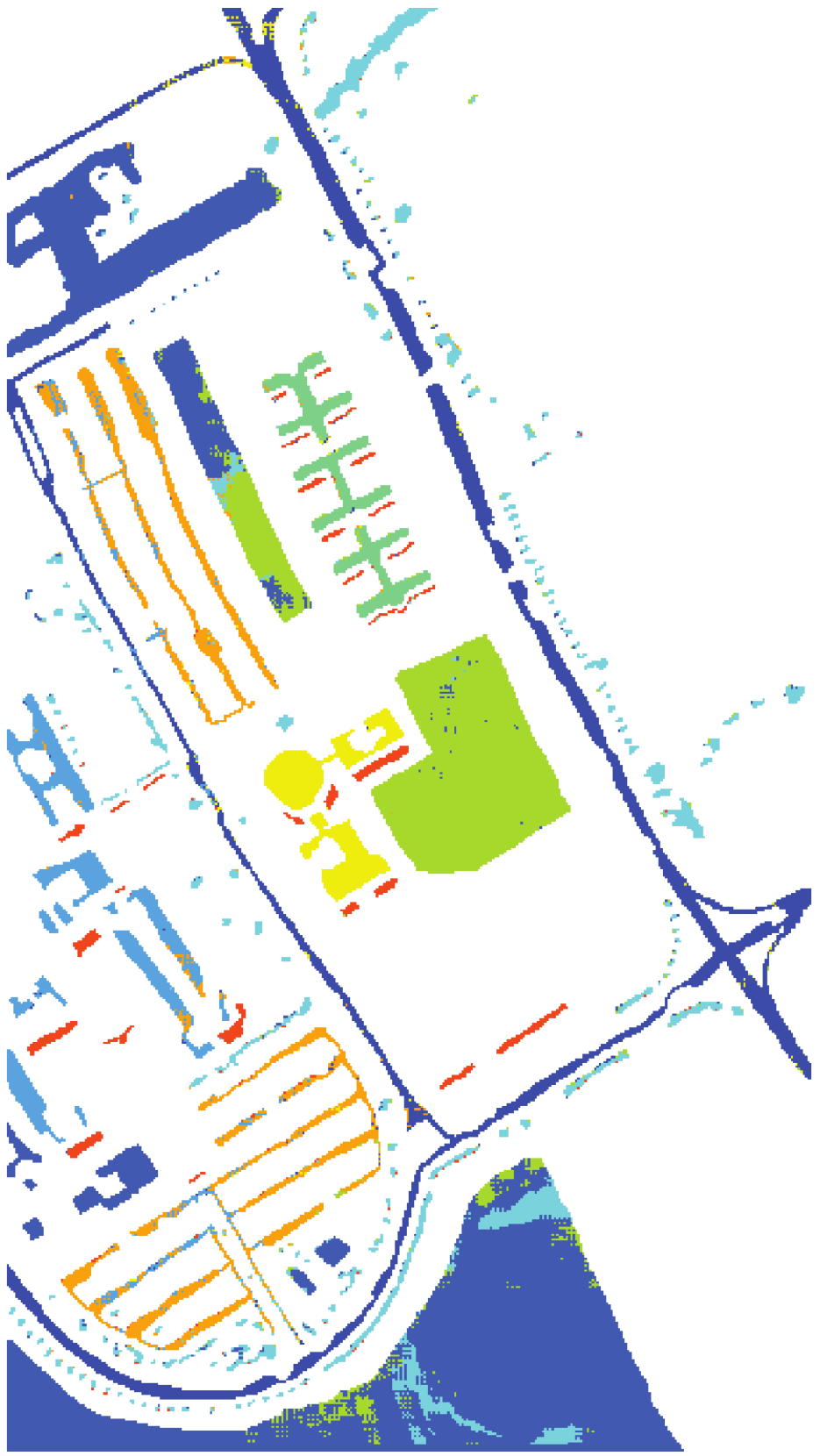}     
    }
    \subfigure[]
    { 
    \label{fig10:d}     
    \includegraphics[width=2cm]{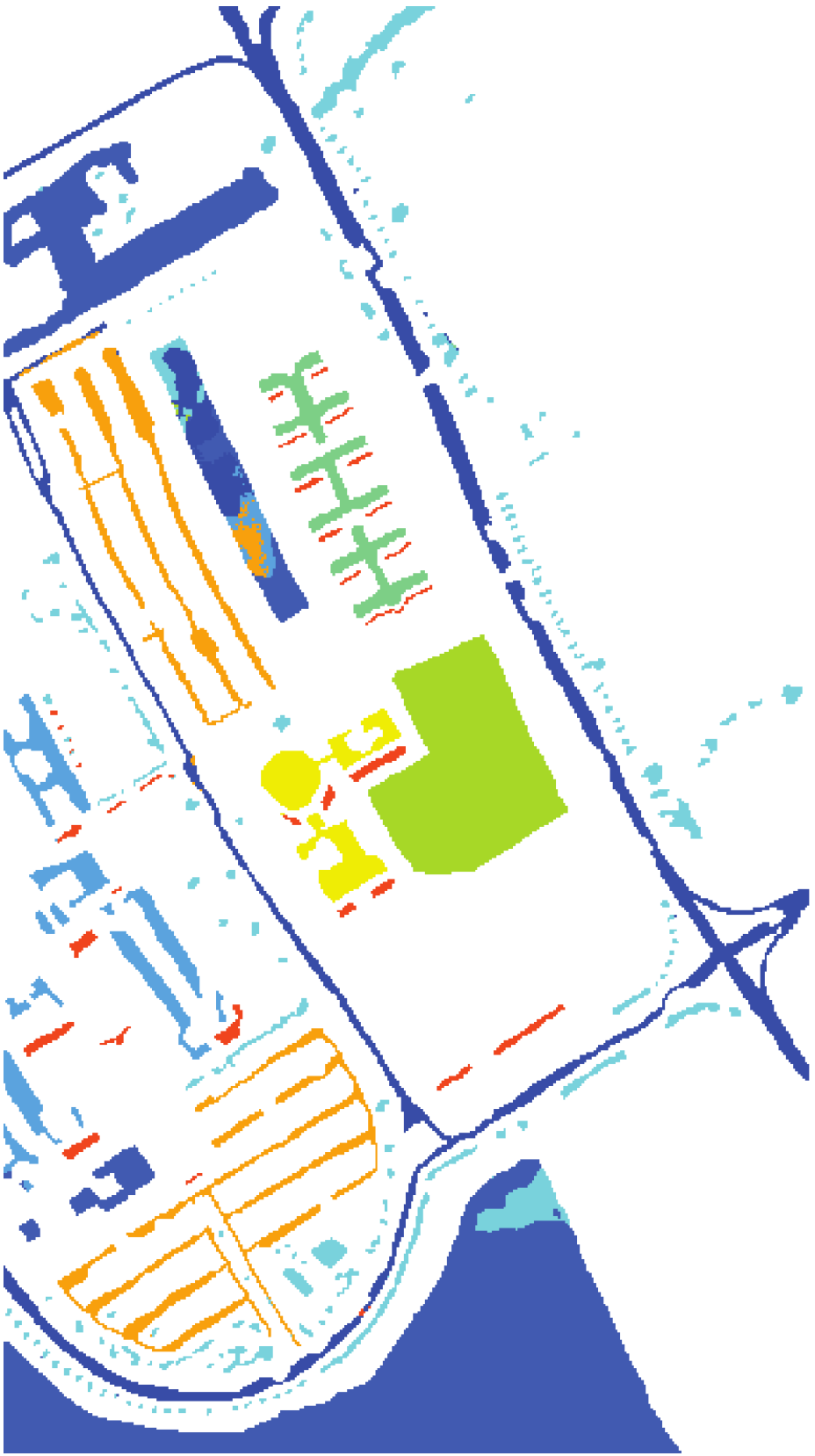}     
    }
    \subfigure[]
    { 
    \label{fig10:e}     
    \includegraphics[width=2cm]{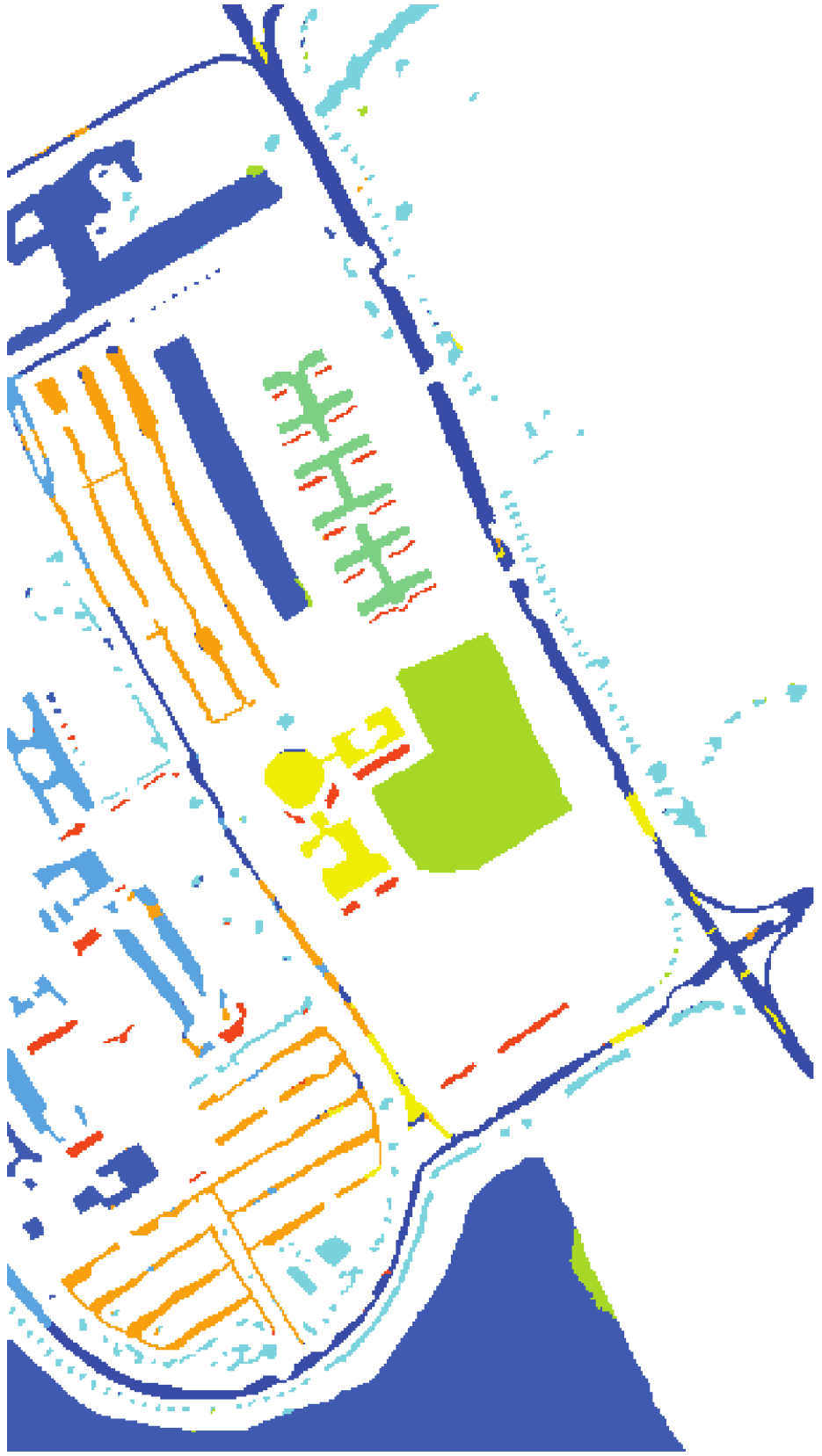}     
    }
    \subfigure[]
    { 
    \label{fig10:f}     
    \includegraphics[width=2cm]{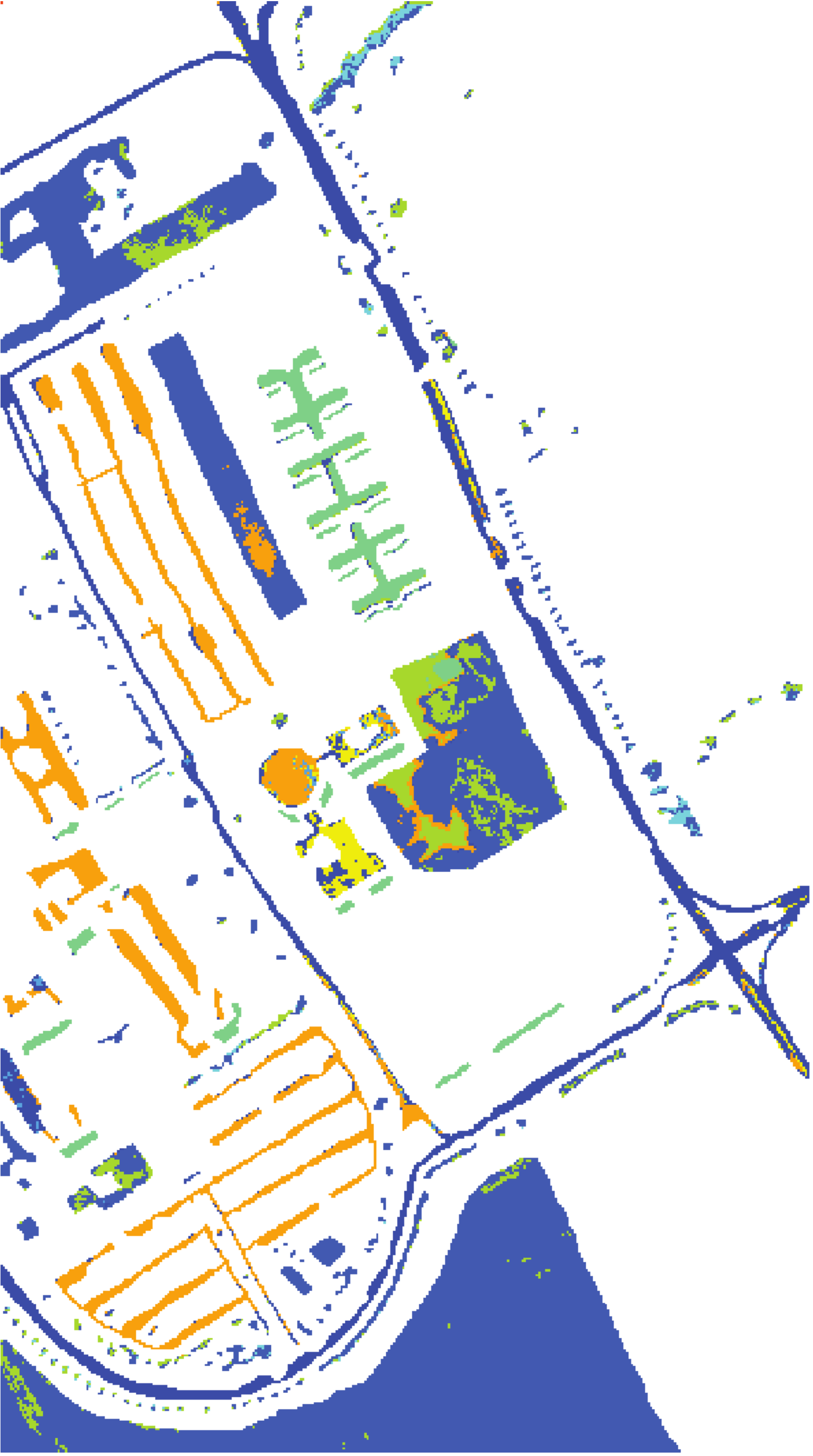}     
    }
    \subfigure[]
    { 
    \label{fig10:g}     
    \includegraphics[width=2cm]{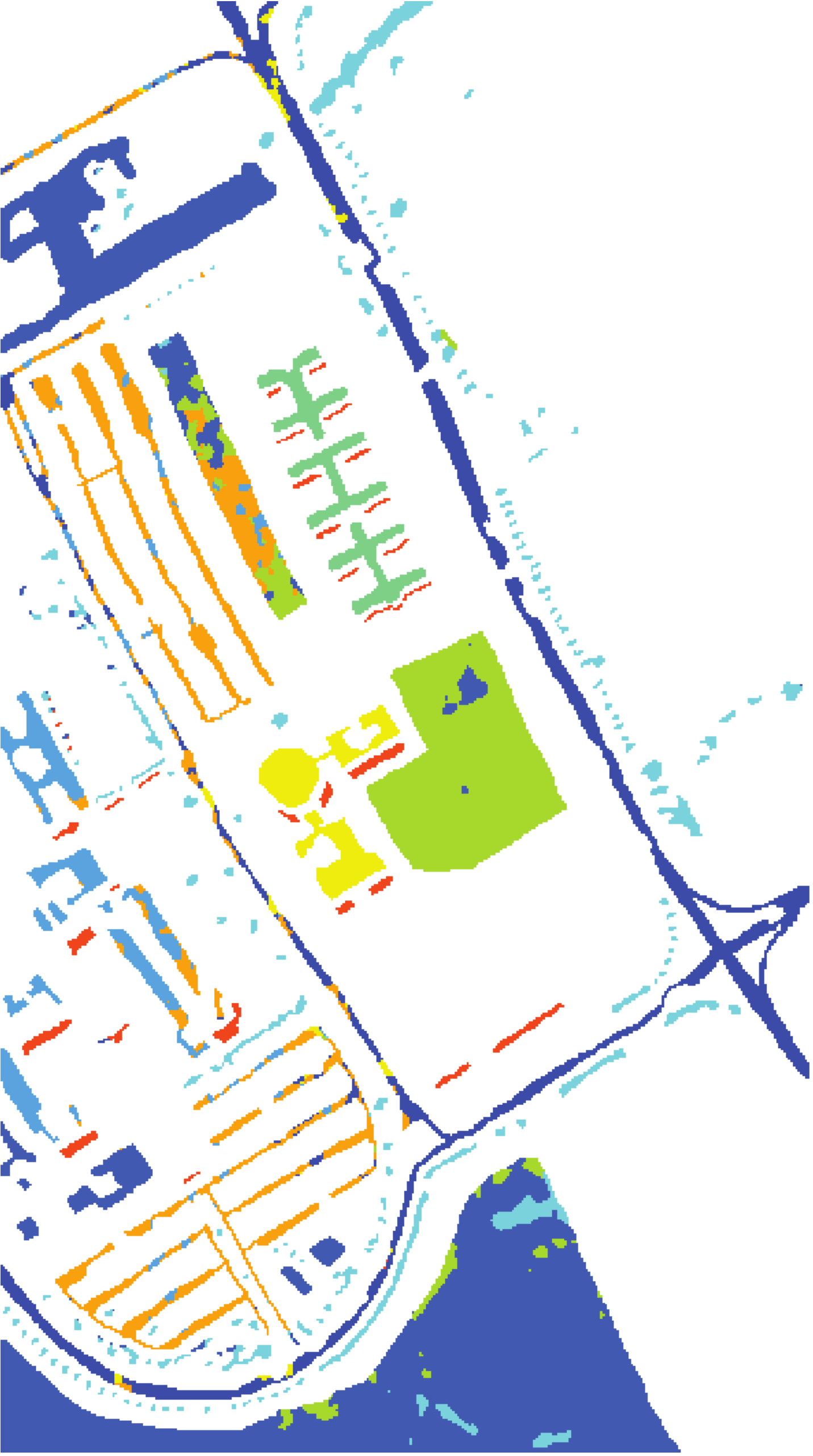}     
    }
    \subfigure[]
    { 
    \label{fig10:h}     
    \includegraphics[width=2cm]{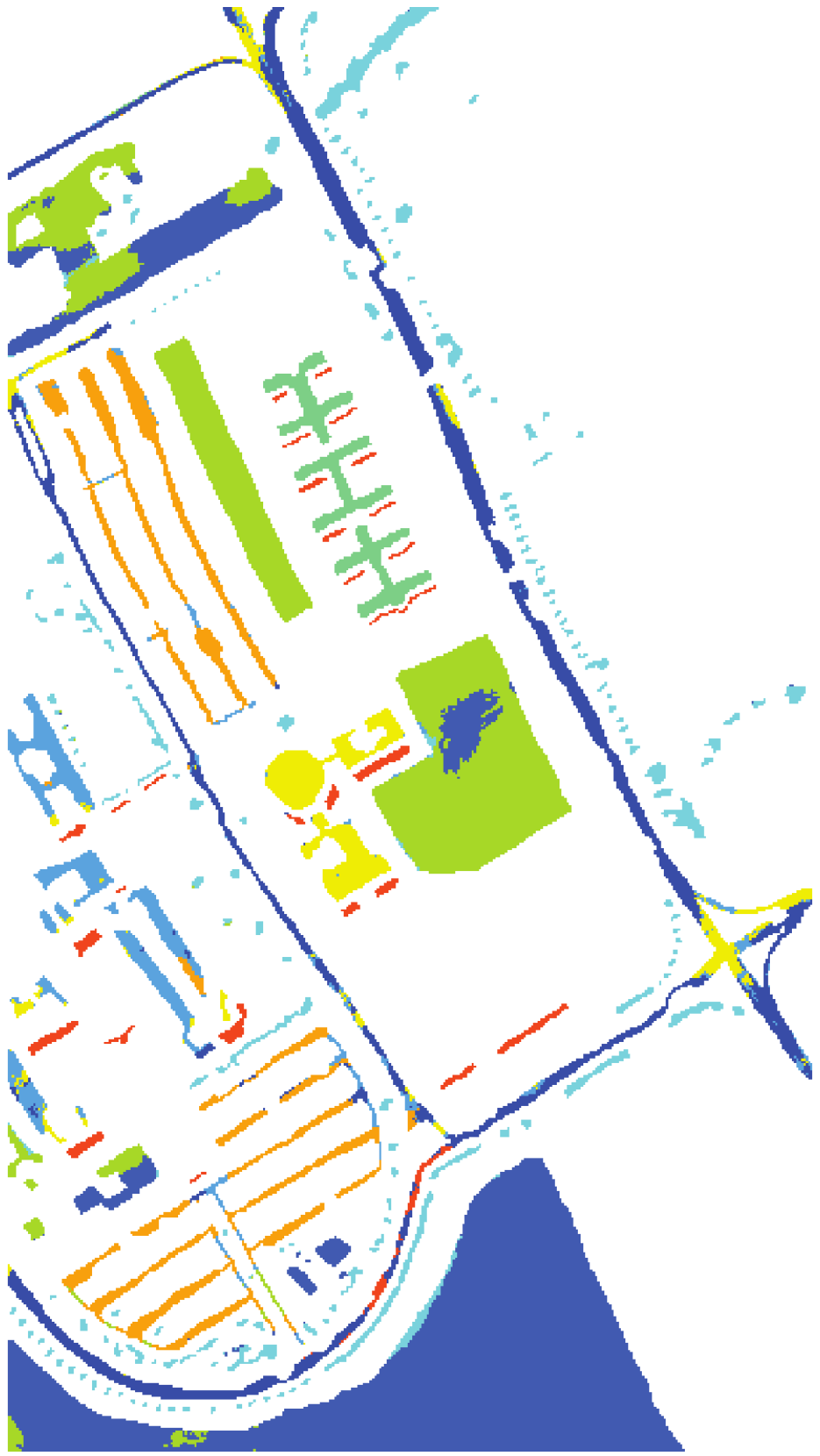}     
    }
    \subfigure[]
    { 
    \label{fig10:i}     
    \includegraphics[width=2cm]{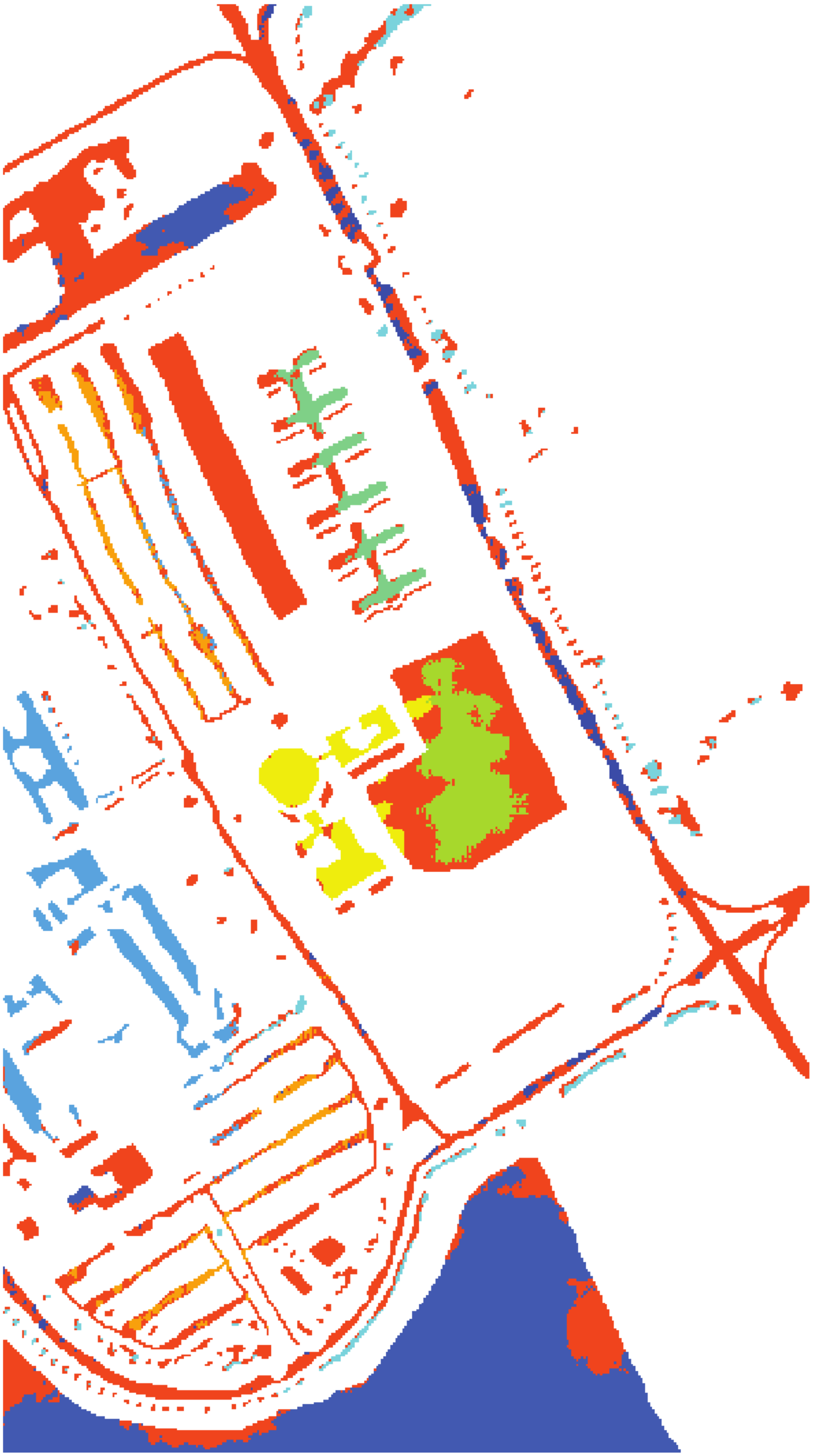}     
    }
    \subfigure[]
    { 
    \label{fig10:j}     
    \includegraphics[width=2cm]{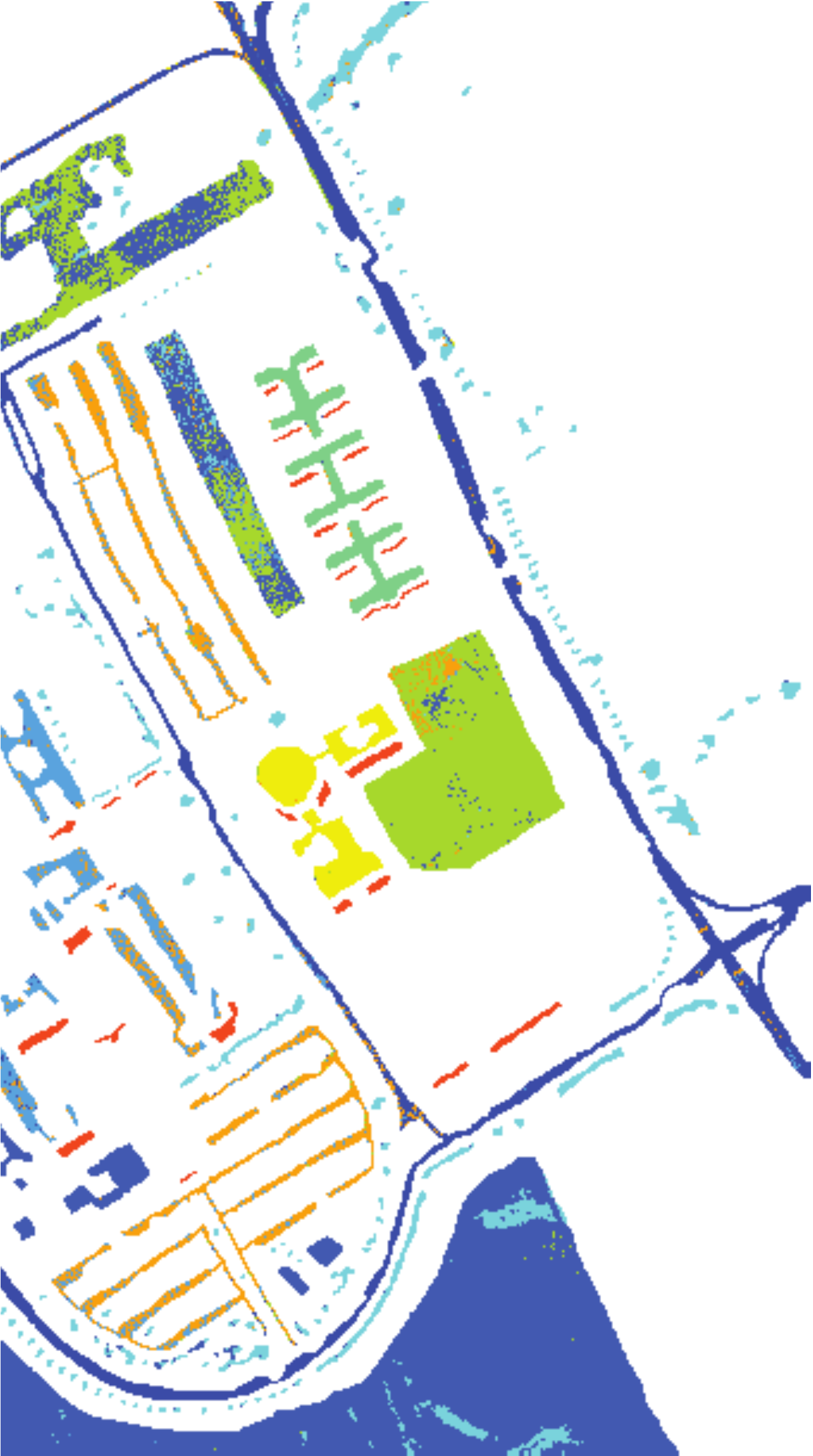}     
    }
    \subfigure[]
    { 
    \label{fig10:k}     
    \includegraphics[width=2cm]{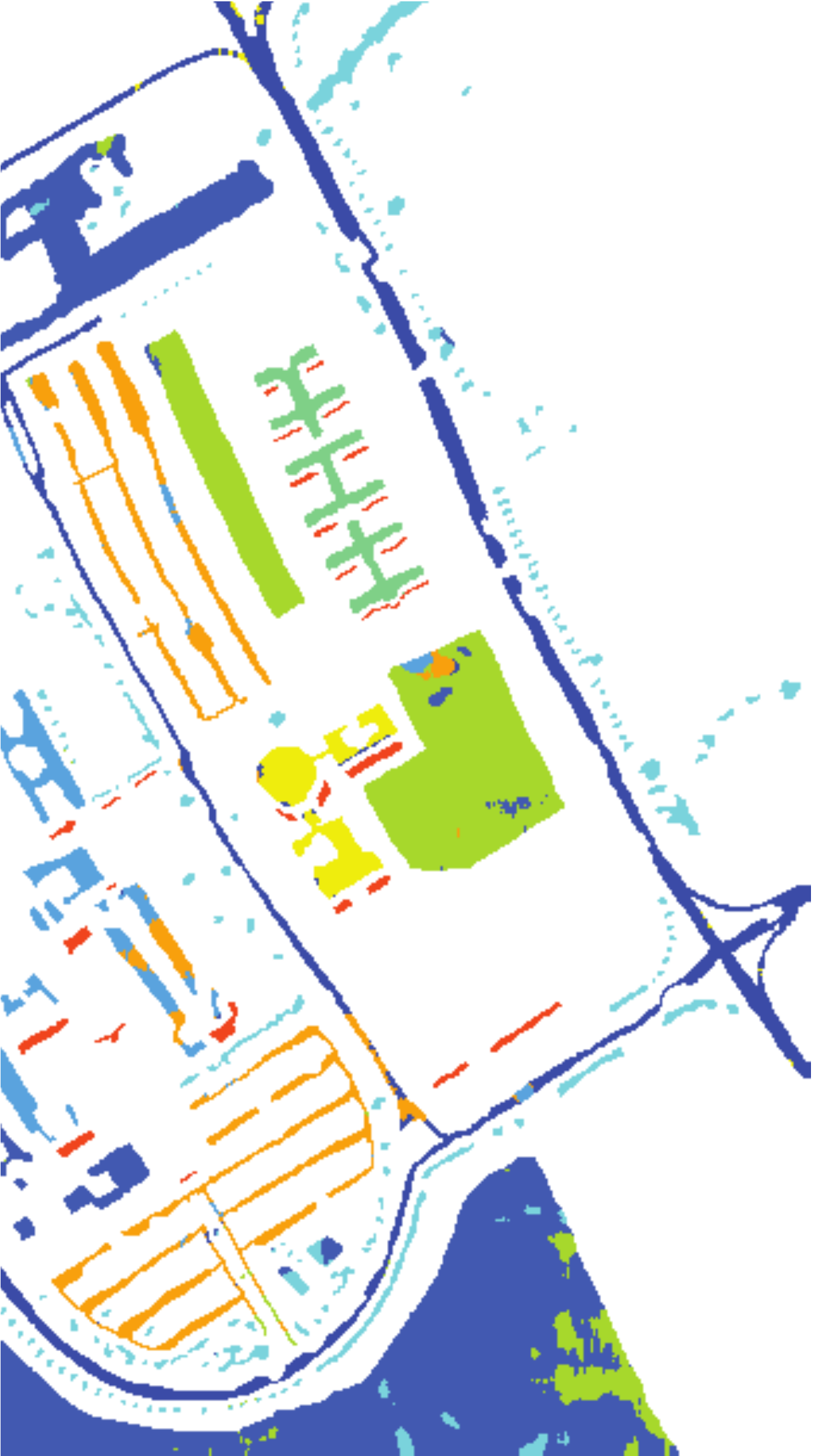}     
    }
    \subfigure[]
    { 
    \label{fig10:l}     
    \includegraphics[width=2cm]{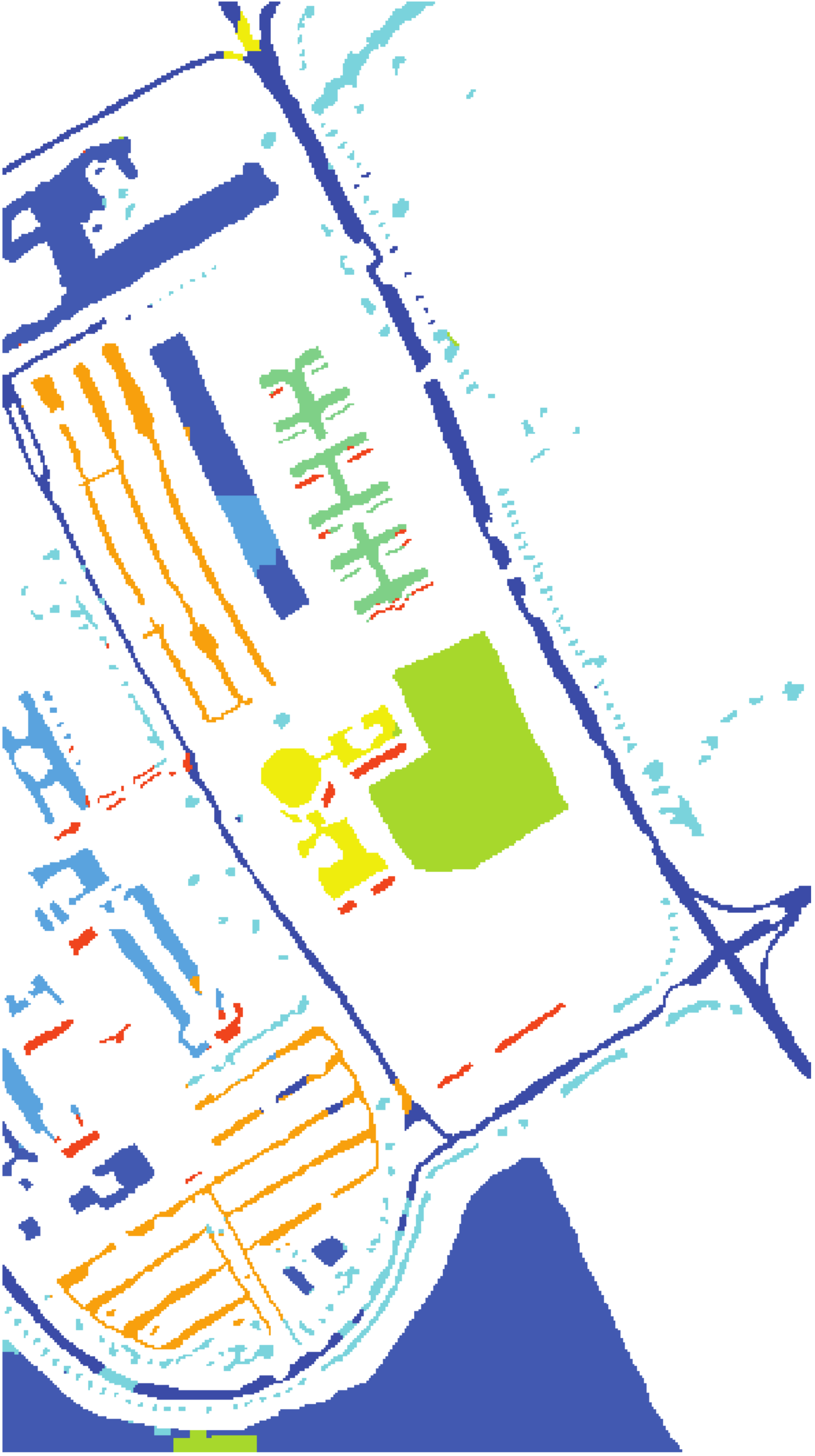}     
    }
    \caption{Classification maps obtained by different methods on \textit{University of Pavia} dataset. (a) False-color image. (b) Ground-truth map. (c) SMR. (d) MSSGU. (e) SGL. (f) SSCL. (g) A$^{2}$S$^{2}$K. (h) ASSMN. (i) ADGAN. (j) MFL. (k) JSDF. (l) ConGCN (Proposed).\vspace{-3em}}     
    \label{fig10}
    \end{figure*}

Fig.~\ref{fig10} visualizes the classification results generated by different methods on the \textit{University of Pavia} dataset. As depicted in Fig.~\ref{fig10:l}, the classification map of our proposed ConGCN is noticeably closer to the ground-truth map (Fig.~\ref{fig10:b}) than those of other methods, which is consistent with previous results in \autoref{table6}. Besides, A$^{2}$S$^{2}$K (Fig.~\ref{fig10:g}) and ASSMN (Fig.~\ref{fig10:h}), which use the fixed convolutional kernels, produce more errors than ConGCN. In the classification map of ADGAN (Fig.~\ref{fig10:i}), most of the pixels are misclassified as the same land cover category, which illustrates the reason for ADGAN's poor OA. It is also notable that large numbers of pixels in the classification map of SSCL (Fig.~\ref{fig10:f}) are misclassified. It indicates that the performance of simply using the traditional paradigm of contrastive learning is far from satisfactory, especially with few labeled examples.

\begin{table*}[t]
  \scriptsize
  \centering
  \caption{Per-class accuracy, OA, AA (\%), and Kappa coefficient achieved by different methods on \textit{University of Pavia} dataset. The best and second best records in each row are \textbf{bolded} and \underline{underlined}, respectively.}
  \label{table6}
  \begin{tabular}{c|c|c|c|c|c|c|c|c|c|c}
  \hline
  \hline
  ID    & SMR \cite{wang2021toward} & MSSGU \cite{liu2021multilevel} & SGL \cite{sellars2020superpixel} & SSCL \cite{9664575}     & A$^{2}$S$^{2}$K \cite{roy2020attention}      & ASSMN \cite{wang2020adaptive1}  & ADGAN \cite{wang2020adaptive} & MFL \cite{li2014multiple} & JSDF \cite{bo2015hyperspectral}    & ConGCN                   \\ \hline
  1     & \underline{96.96$\pm$1.89}            & \textbf{98.14$\pm$0.72}        & 88.26$\pm$3.06                   & 96.94$\pm$1.76          & 94.25$\pm$2.75                               & 78.33$\pm$3.30                  & 49.97$\pm$20.68               & 94.45$\pm$0.25            & 82.40$\pm$4.07                     & 93.11$\pm$2.43           \\
  2     & 82.00$\pm$6.11            & 82.90$\pm$2.30                 & 97.24$\pm$1.45                   & \underline{97.90$\pm$1.90} & \textbf{97.92$\pm$2.08}                               & 74.97$\pm$1.80                  & 33.82$\pm$23.16               & 90.17$\pm$0.65            & 90.76$\pm$3.74                     & 96.55$\pm$1.87           \\
  3     & 95.21$\pm$2.60            & \textbf{100.00$\pm$0.00}       & 94.19$\pm$2.64                   & 14.89$\pm$16.14         & 72.27$\pm$11.87                              & 81.14$\pm$6.84                  & 70.07$\pm$35.81               & 85.05$\pm$0.54            & 86.71$\pm$4.14                     & \underline{97.24$\pm$1.83}           \\
  4     & 94.73$\pm$3.47            & \textbf{98.44$\pm$0.45}        & 92.76$\pm$1.97                   & 42.87$\pm$19.21         & 78.73$\pm$5.59                               & \underline{97.76$\pm$1.06}                  & 54.79$\pm$24.08               & 93.31$\pm$0.28            & 92.88$\pm$2.16                     & 93.91$\pm$0.25           \\
  5     & 99.03$\pm$1.48            & \textbf{100.00$\pm$0.00}       & 99.30$\pm$0.82                   & 87.00$\pm$13.48         & 99.92$\pm$0.02                               & \underline{99.98$\pm$0.03}                  & 91.10$\pm$25.17               & 99.38$\pm$0.02            & \textbf{100.00$\pm$0.00}           & 98.80$\pm$0.15           \\
  6     & 99.79$\pm$0.36            & \textbf{100.00$\pm$0.00}       & \underline{99.80$\pm$0.51}                   & 9.43$\pm$7.15           & 81.74$\pm$7.58                               & 90.76$\pm$3.69                  & 58.62$\pm$31.24               & 93.31$\pm$0.20            & 94.30$\pm$4.55                     & \textbf{100.00$\pm$0.00} \\
  7     & 97.59$\pm$5.41            & \textbf{100.00$\pm$0.00}       & 99.31$\pm$0.23                   & 14.21$\pm$12.56         & 81.40$\pm$9.44                               & 98.43$\pm$1.26                  & 85.29$\pm$29.34               & \underline{99.39$\pm$0.04}            & 96.62$\pm$1.37                     & 99.12$\pm$0.44           \\
  8     & 88.79$\pm$14.78           & \textbf{99.48$\pm$0.97}        & 94.07$\pm$2.12                   & 81.36$\pm$14.61         & 70.44$\pm$6.86                               & 89.38$\pm$1.92                  & 48.79$\pm$31.33               & 85.30$\pm$0.53            & 94.69$\pm$3.74                     & \underline{94.76$\pm$1.93}           \\
  9     & 99.09$\pm$1.47            & \underline{99.86$\pm$0.11}                 & 99.67$\pm$0.00                   & 2.30$\pm$6.87           & 95.60$\pm$4.33                               & \textbf{99.87$\pm$0.18}         & 90.08$\pm$7.88                & 99.76$\pm$0.02            & 99.56$\pm$0.36                     & 82.81$\pm$3.20           \\ \hline
  OA    & 90.13$\pm$3.94            & 92.06$\pm$1.01                 & \underline{95.58$\pm$0.60}                   & 72.85$\pm$2.44          & 87.82$\pm$1.47                               & 82.54$\pm$0.99                  & 48.35$\pm$17.95               & 91.54$\pm$0.30            & 90.82$\pm$1.30                     & \textbf{95.97$\pm$0.90}  \\
  AA    & 94.80$\pm$3.11            & \textbf{97.65$\pm$0.26}        & \underline{96.07$\pm$0.47}                   & 49.66$\pm$3.58          & 85.81$\pm$1.82                               & 90.07$\pm$1.20                  & 64.73$\pm$16.71               & 93.35$\pm$0.10            & 93.10$\pm$0.65                     & 95.14$\pm$0.57           \\
  Kappa & 87.51$\pm$4.86            & 89.81$\pm$1.26                 & \underline{94.17$\pm$0.78}                   & 61.18$\pm$3.55          & 84.32$\pm$1.88                               & 77.87$\pm$1.20                  & 42.94$\pm$17.63               & 88.98$\pm$0.38            & 88.02$\pm$1.62                     & \textbf{94.69$\pm$1.17}  \\ \hline
  \hline
  \end{tabular}
  \end{table*}

\subsubsection{Results on \textit{Salinas} Dataset}
The experimental results of different methods on \textit{Salinas} dataset are presented in \autoref{table7}. It is apparent that the performances of nearly all methods on the \textit{Salinas} dataset are better than those on the \textit{Indian Pines} dataset and the \textit{University of Pavia} dataset. The reason might be that the boundaries in the \textit{Salinas} dataset are more regular than those in the \textit{Indian Pines} and the \textit{University of Pavia} datasets. As a result, the regions of the \textit{Salinas} dataset are more distinguishable than those of the \textit{Indian Pines} and the \textit{University of Pavia} datasets. Although the proposed ConGCN is in the second place, it is only 0.18\% lower than MSSGU in terms of OA.

\begin{figure*} \centering    
      \subfigure[]
      {
       \label{fig11:a}
      \includegraphics[width=1.547cm]{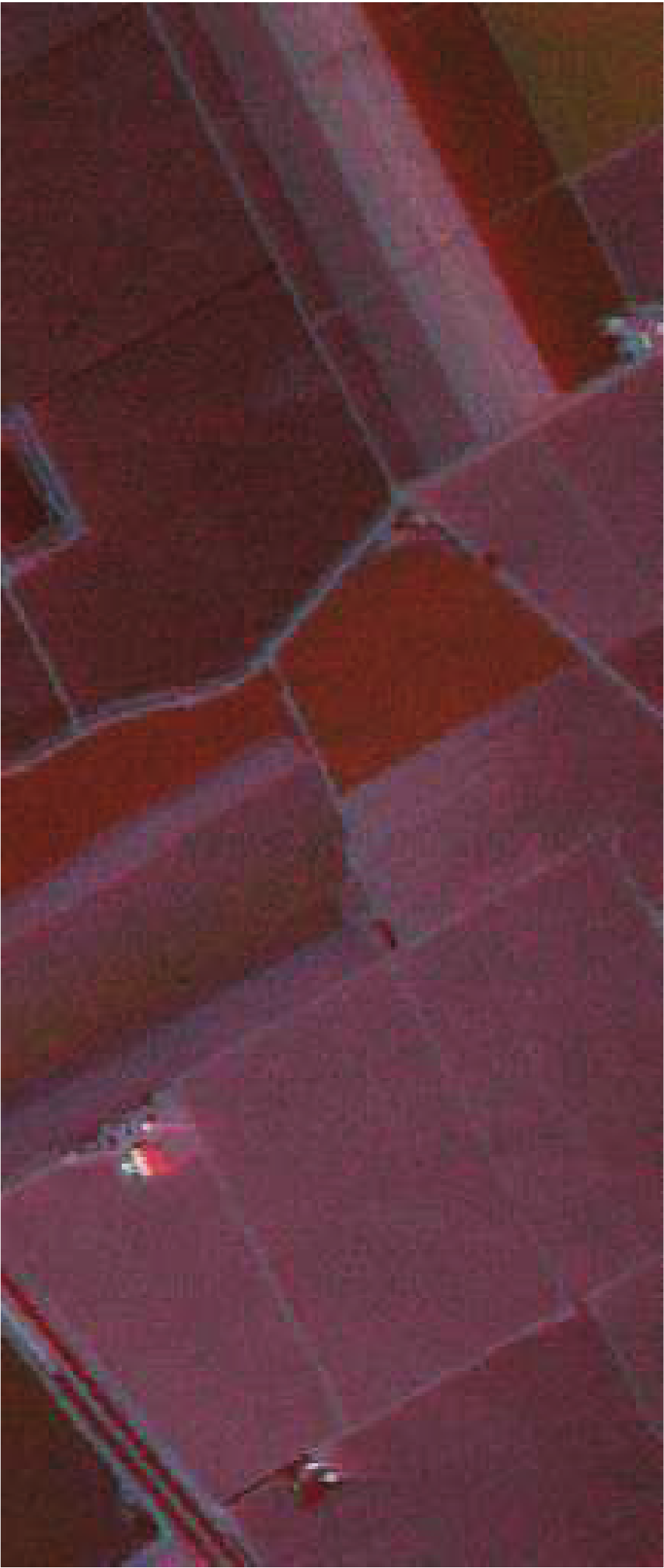}  
      }
      \subfigure[]
      { 
      \label{fig11:b}
      \includegraphics[width=1.547cm]{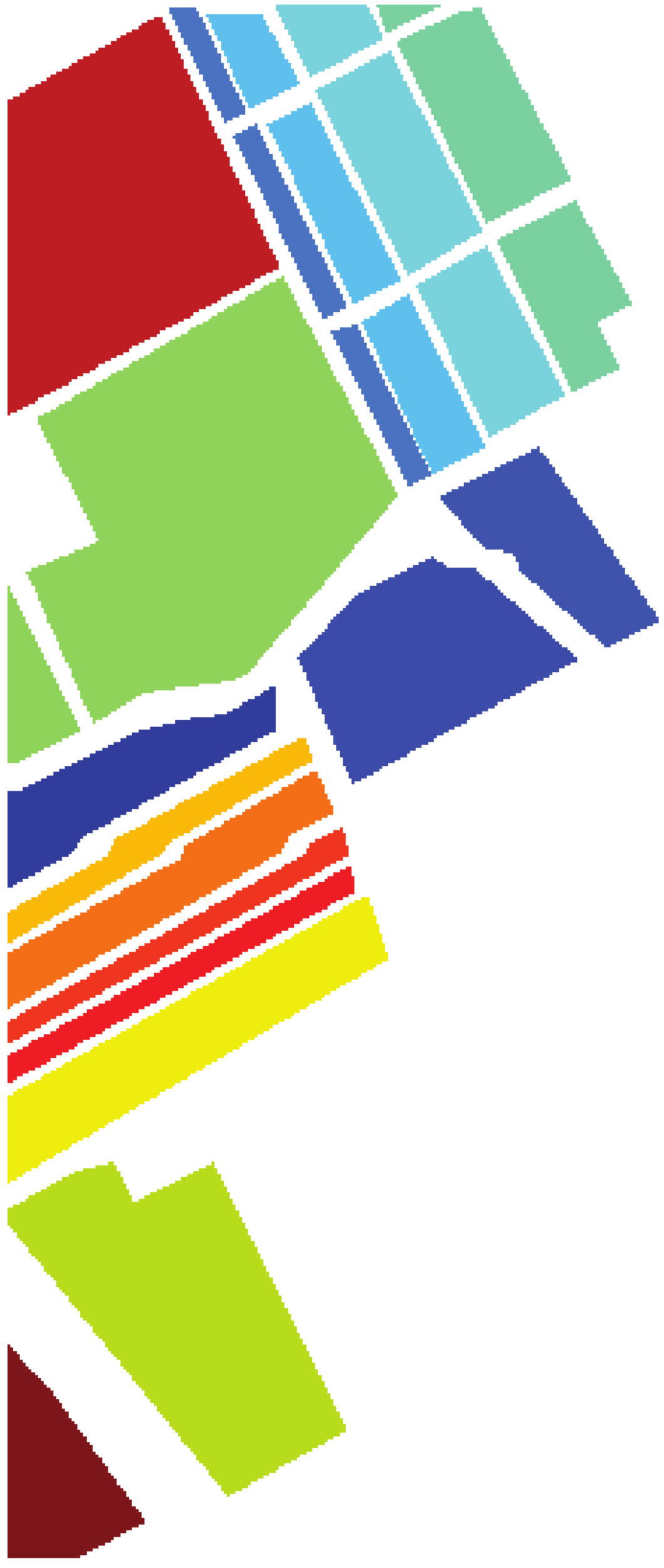}     
      }    
      \subfigure[]
      { 
      \label{fig11:c}     
      \includegraphics[width=1.547cm]{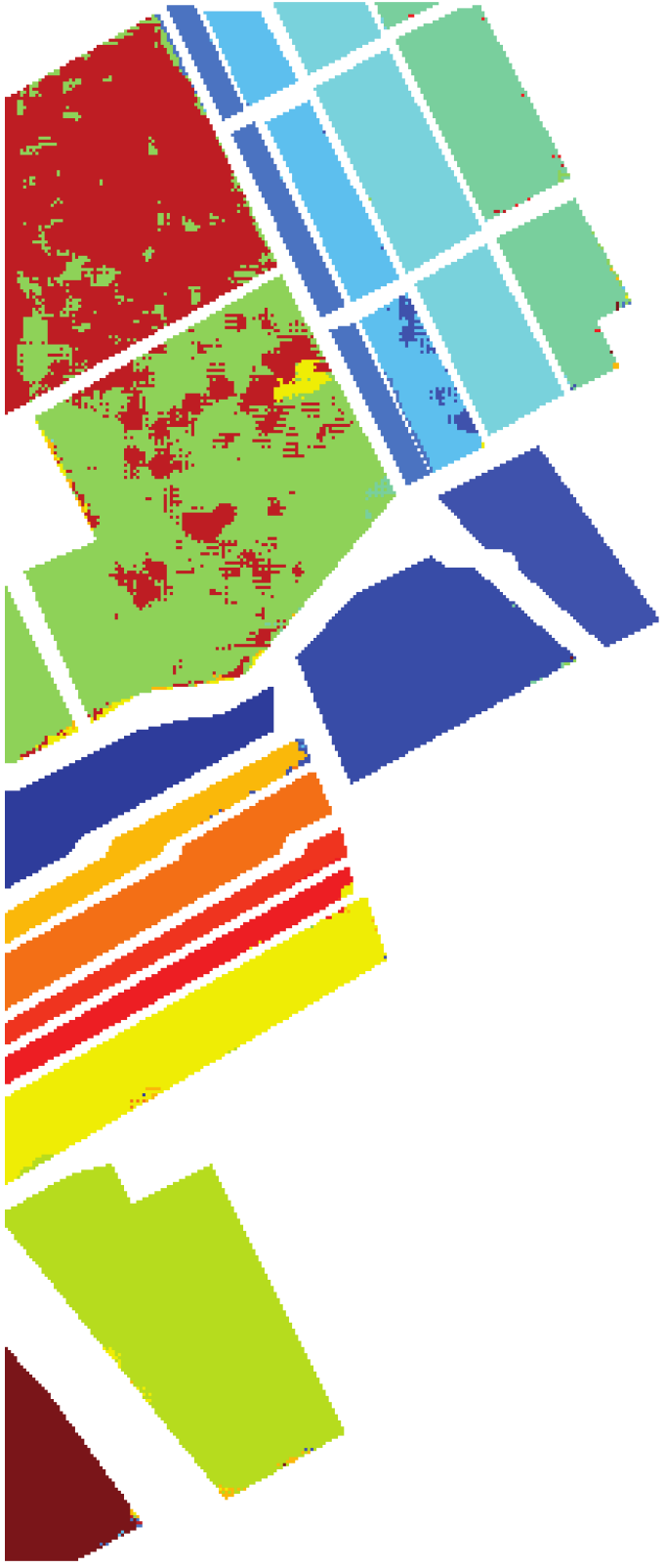}     
      }
      \subfigure[]
      { 
      \label{fig11:d}     
      \includegraphics[width=1.547cm]{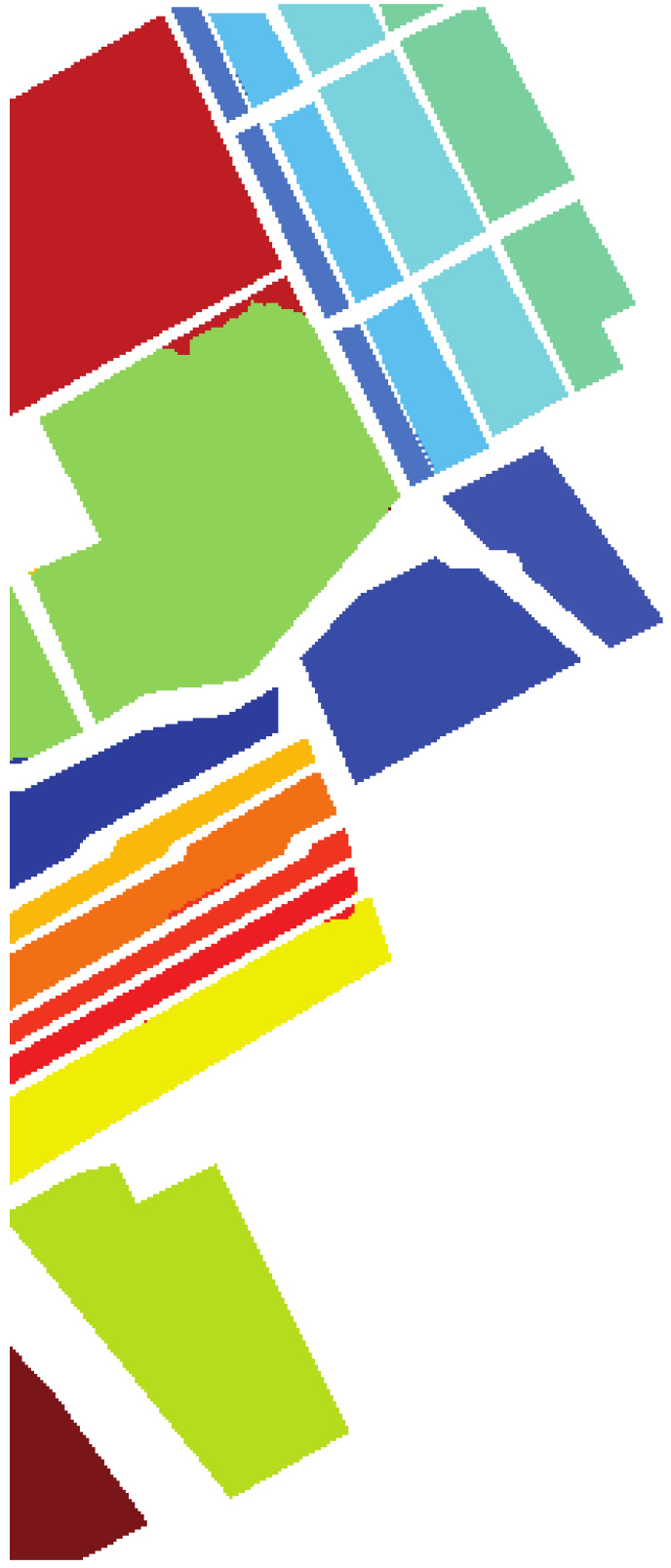}     
      }
      \subfigure[]
      { 
      \label{fig11:e}     
      \includegraphics[width=1.547cm]{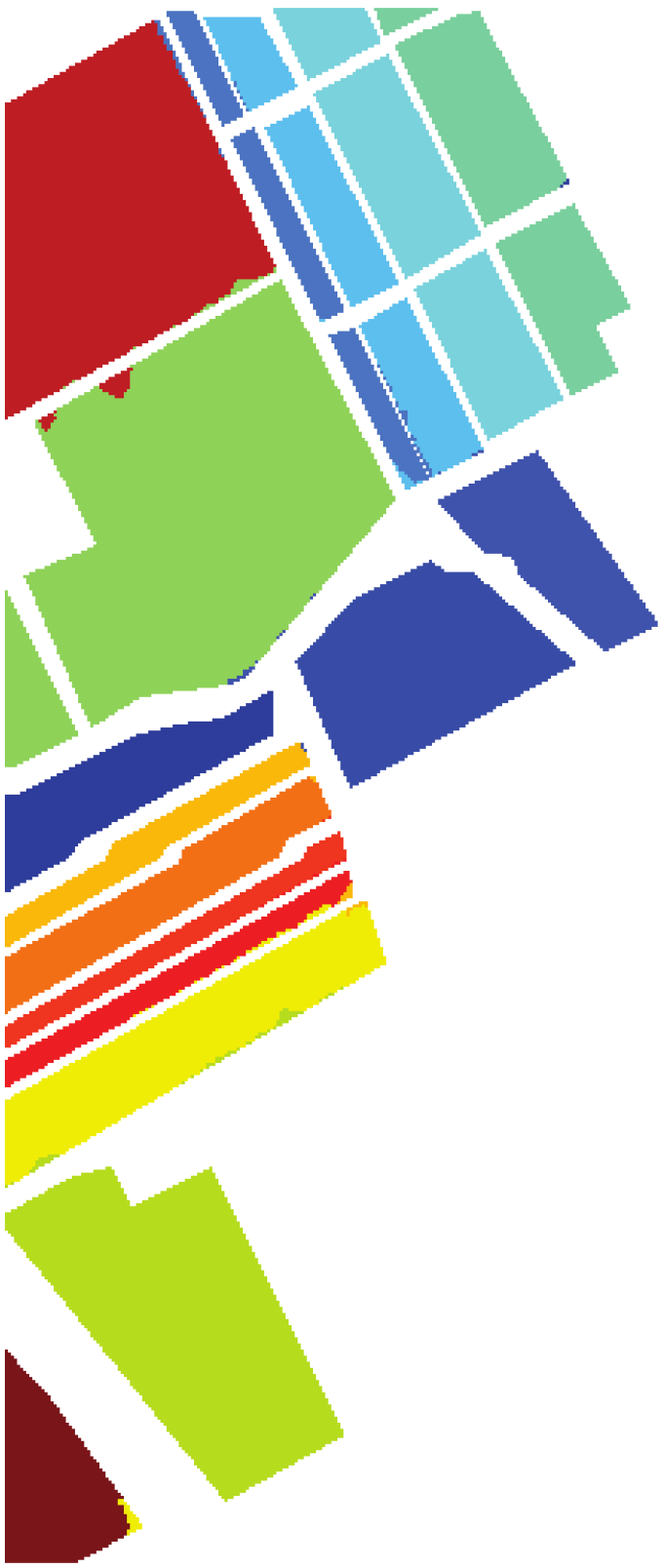}     
      }
      \subfigure[]
      { 
      \label{fig11:f}     
      \includegraphics[width=1.547cm]{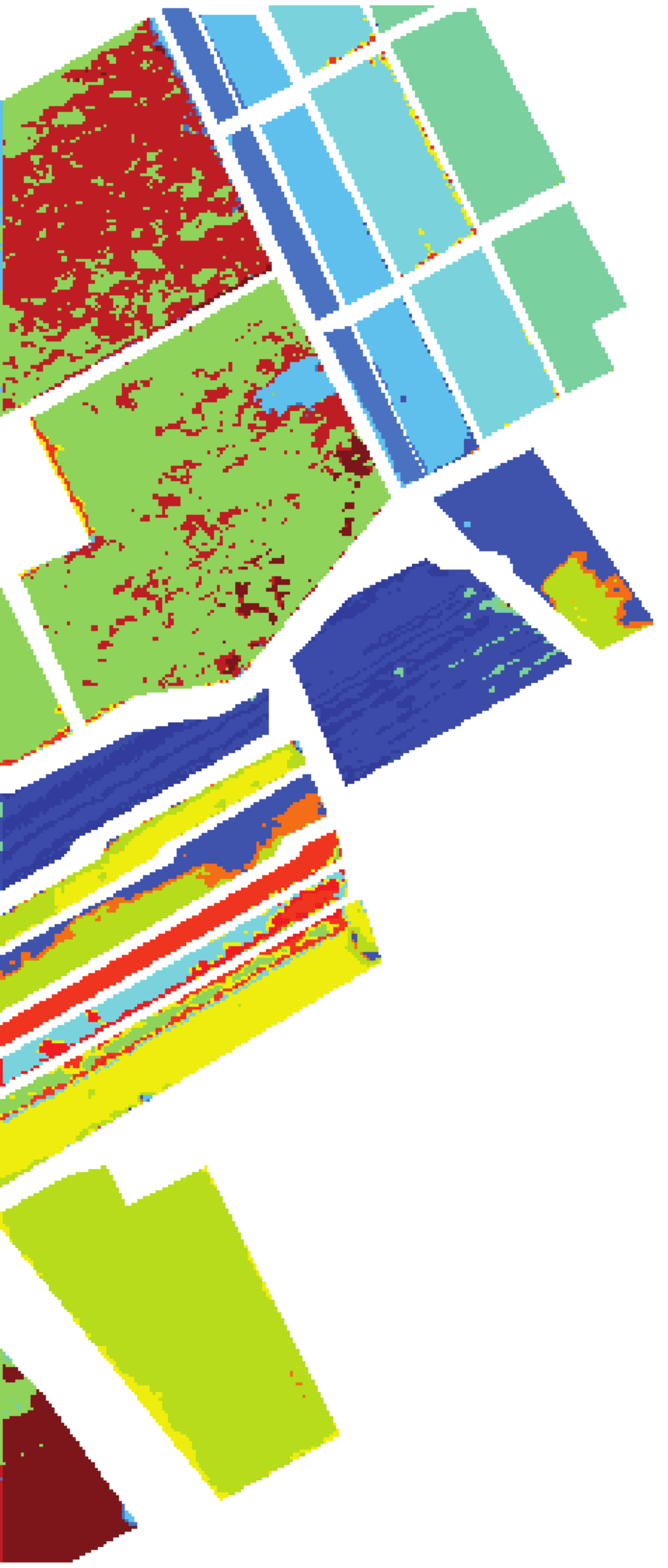}     
      }
      \subfigure[]
      { 
      \label{fig11:g}     
      \includegraphics[width=1.547cm]{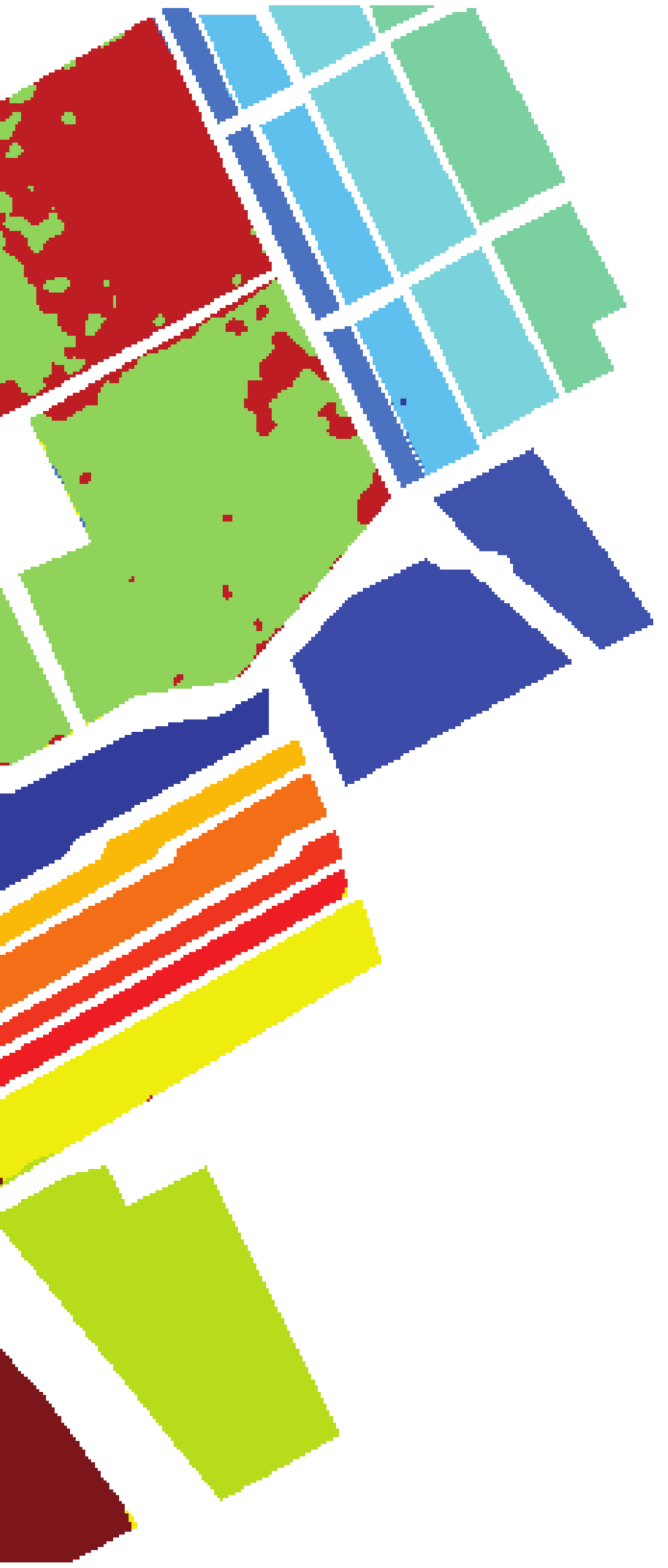}     
      }
      \\
      \subfigure[]
      { 
      \label{fig11:h}     
      \includegraphics[width=1.547cm]{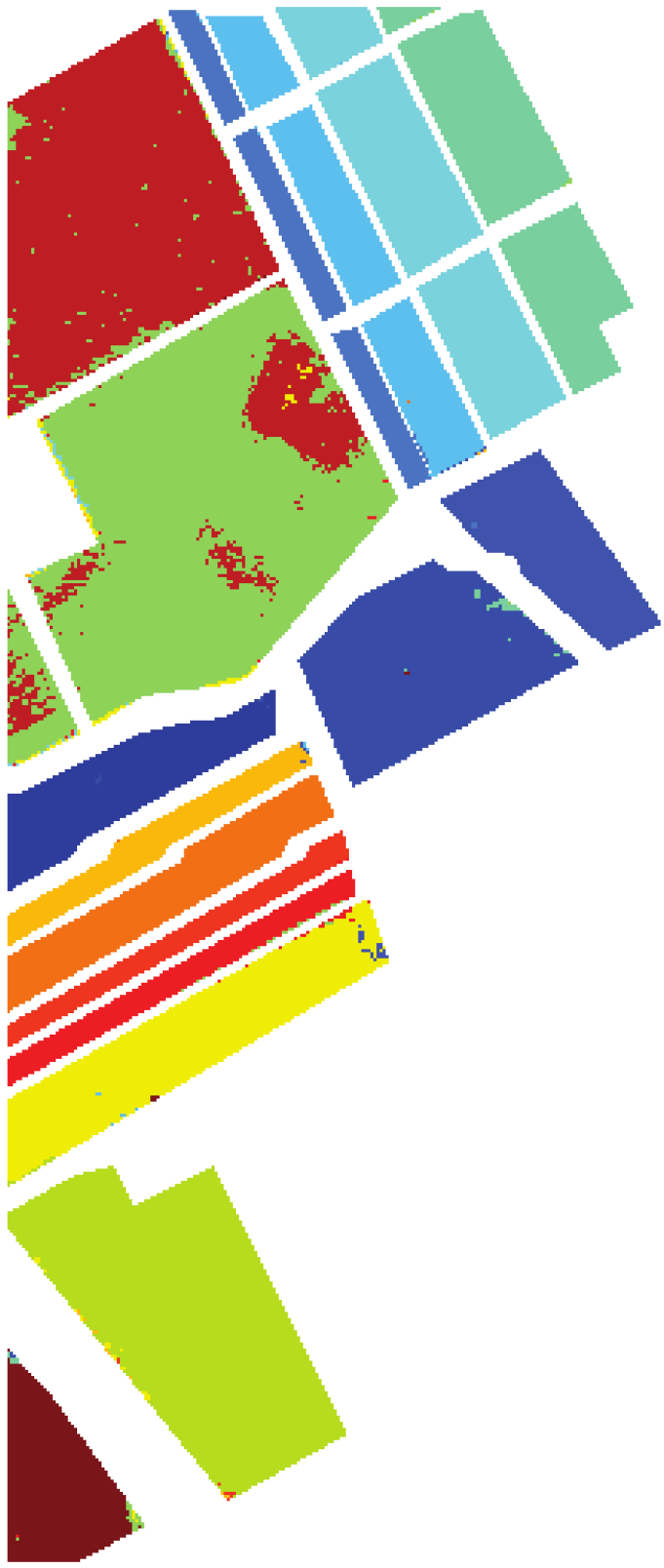}     
      }
      \subfigure[]
      { 
      \label{fig11:i}     
      \includegraphics[width=1.547cm]{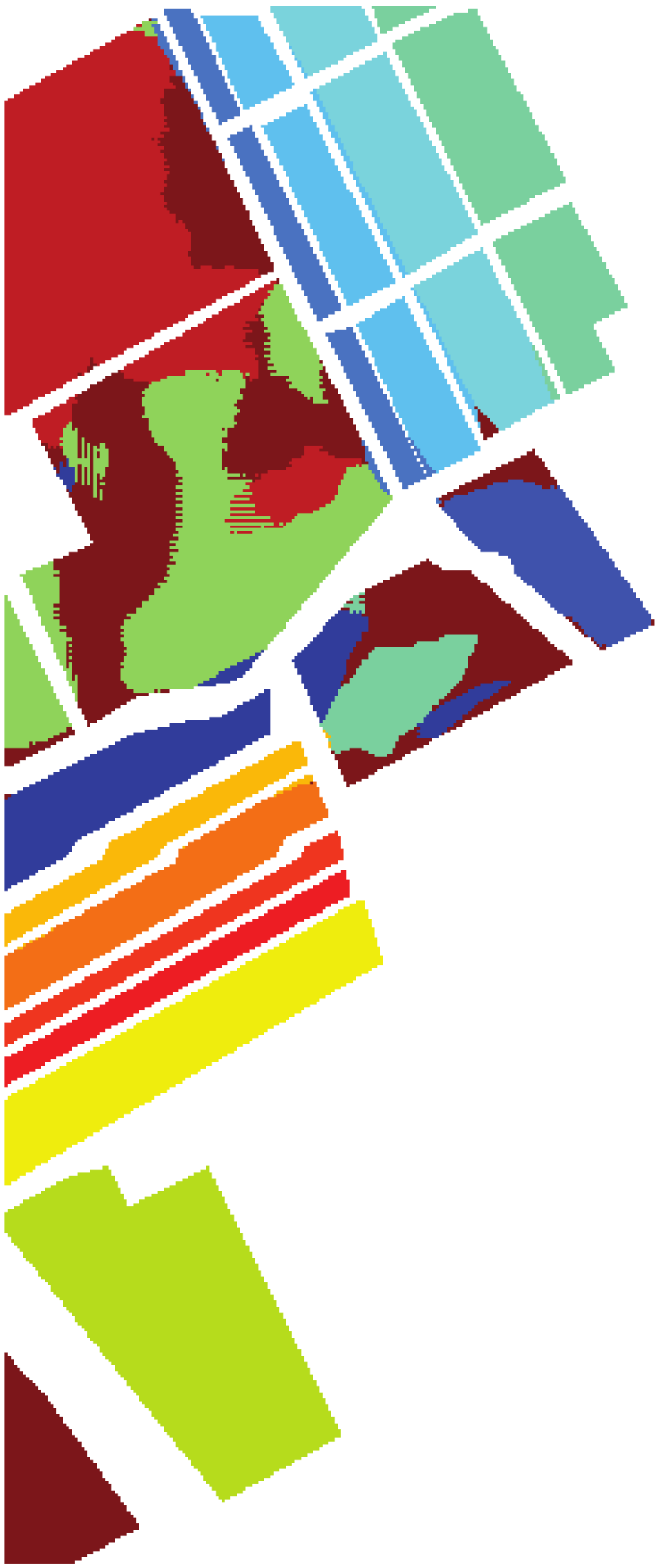}     
      }
      \subfigure[]
      { 
      \label{fig11:j}     
      \includegraphics[width=1.547cm]{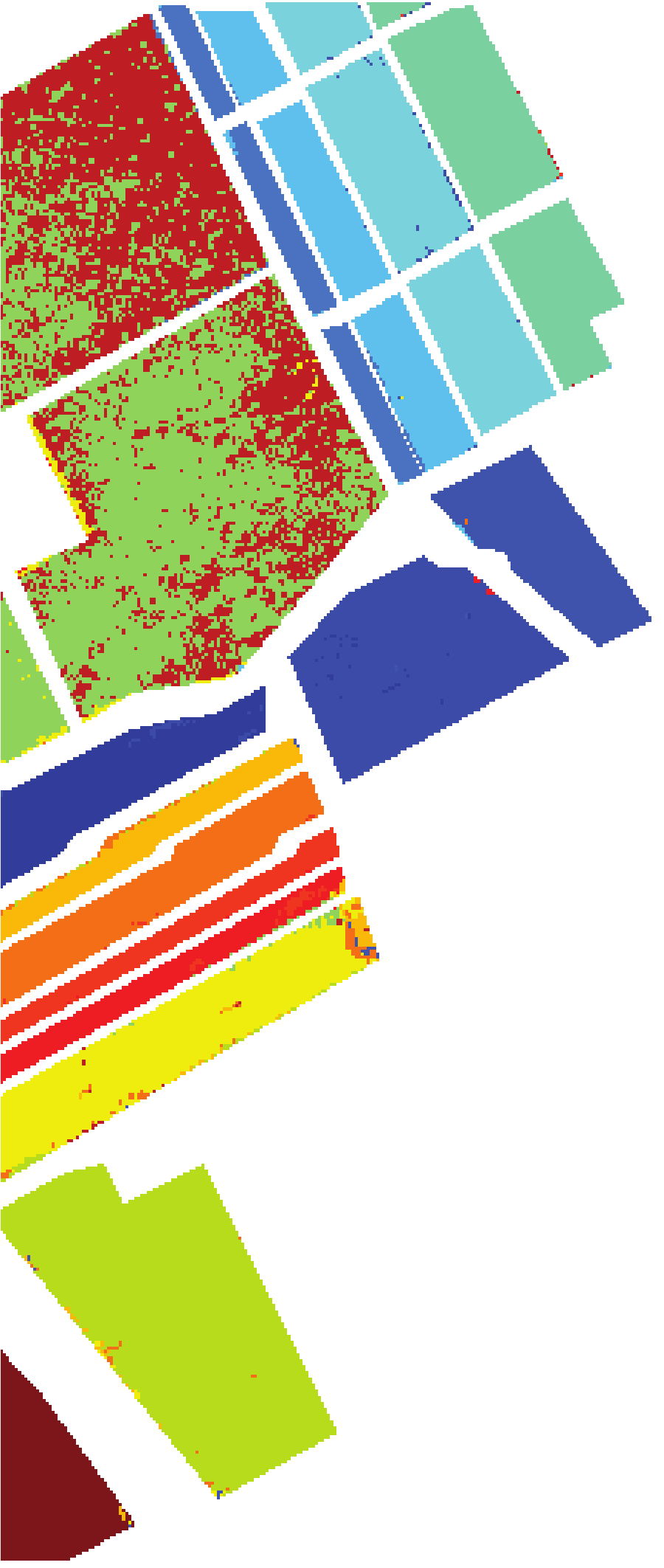}     
      }
      \subfigure[]
      { 
      \label{fig11:k}     
      \includegraphics[width=1.547cm]{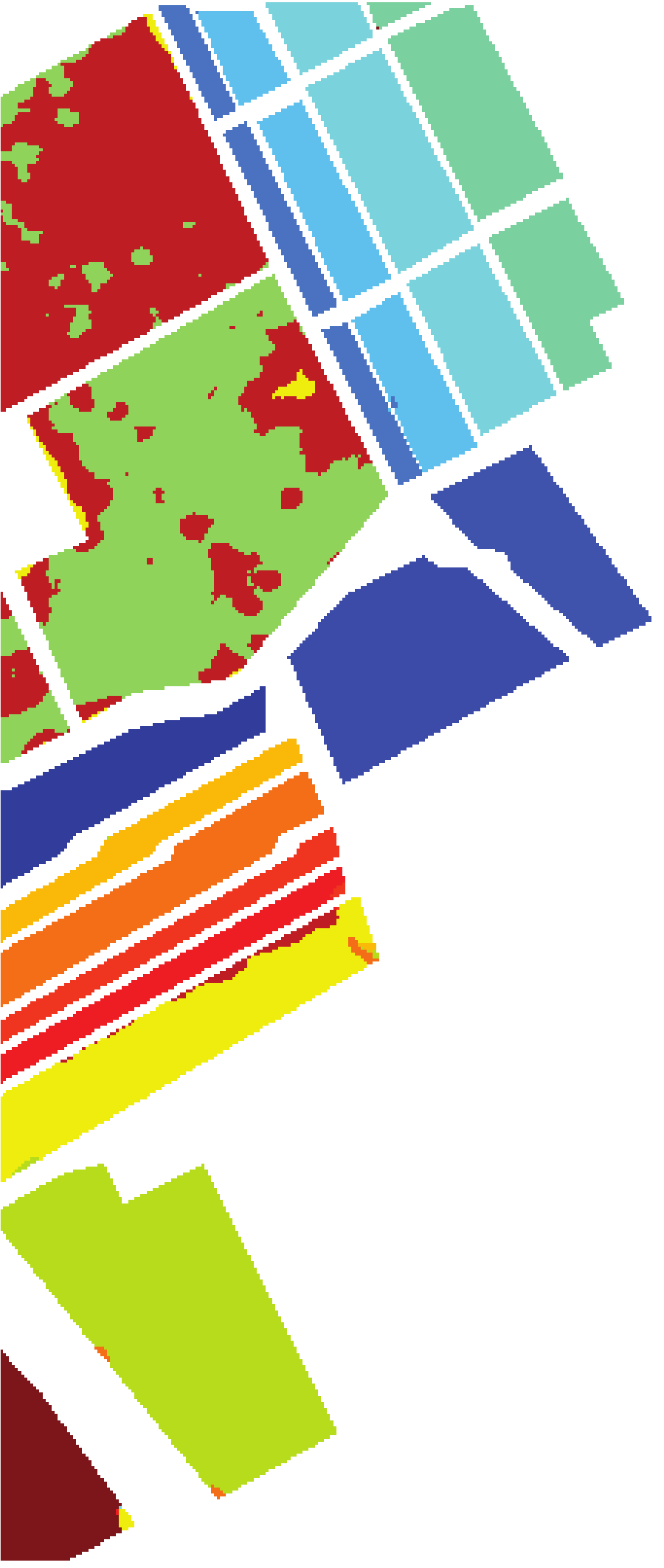}     
      }
      \subfigure[]
      { 
      \label{fig11:l}     
      \includegraphics[width=1.547cm]{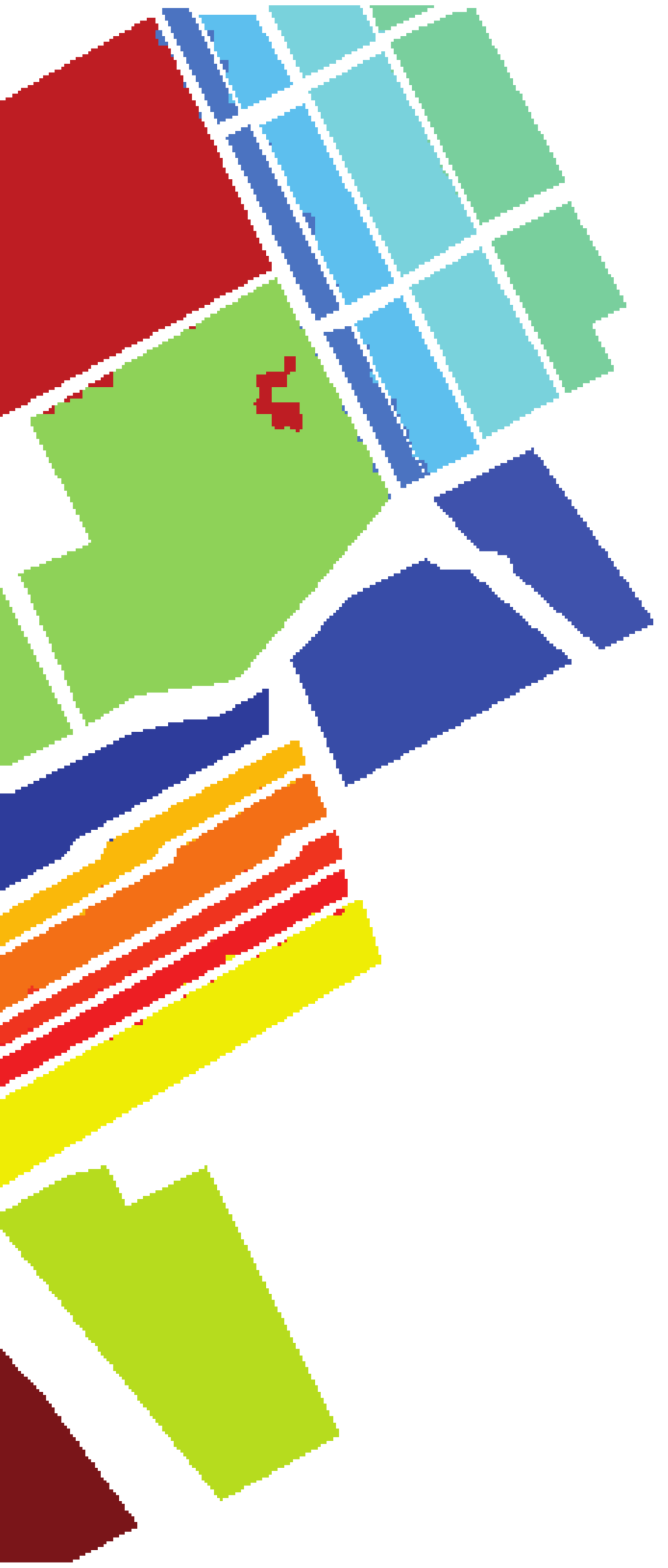}     
      }
      \caption{Classification maps obtained by different methods on \textit{Salinas} dataset. (a) False-color image. (b) Ground-truth map. (c) SMR. (d) MSSGU. (e) SGL. (f) SSCL. (g) A$^{2}$S$^{2}$K. (h) ASSMN. (i) ADGAN. (j) MFL. (k) JSDF. (l) ConGCN (Proposed).}    
      \label{fig11}
      \end{figure*}

Fig. \ref{fig11} visualizes the classification results generated by different methods. It can be observed that some regions in the classification map of our proposed ConGCN (Fig.~\ref{fig11:l}) are less noisy than those of other methods, which is consistent with the results listed in \autoref{table6}.

\begin{table*}[]
    \scriptsize
    \centering
    \caption{Per-class accuracy, OA, AA (\%), and Kappa coefficient achieved by different methods on \textit{Salinas} dataset. The best and second best records in each row are \textbf{bolded} and \underline{underlined}, respectively.}
    \label{table7}
    \begin{tabular}{c|c|c|c|c|c|c|c|c|c|c}
    \hline
    \hline
    ID    & SMR \cite{wang2021toward} & MSSGU \cite{liu2021multilevel} & SGL \cite{sellars2020superpixel}  & SSCL \cite{9664575} & A$^{2}$S$^{2}$K \cite{roy2020attention}      & ASSMN \cite{wang2020adaptive1}  & ADGAN \cite{wang2020adaptive}  & MFL \cite{li2014multiple} & JSDF \cite{bo2015hyperspectral}  & ConGCN                   \\ \hline
    1     & \underline{99.82$\pm$0.31}            & \textbf{100.00$\pm$0.00}       & \textbf{100.00$\pm$0.00}          & 81.57$\pm$13.21     & \textbf{100.00$\pm$0.00}                     & 97.68$\pm$6.17                  & 95.36$\pm$6.40                 & 99.63$\pm$0.07            & \textbf{100.00$\pm$0.00}         & \textbf{100.00$\pm$0.00} \\
    2     & 99.53$\pm$0.73            & \underline{99.99$\pm$0.03}                 & \textbf{100.00$\pm$0.00}          & 81.56$\pm$8.56      & 99.98$\pm$0.05                               & 99.02$\pm$0.46                  & 54.19$\pm$42.34                & 99.34$\pm$0.06            & \textbf{100.00$\pm$0.00}         & \textbf{100.00$\pm$0.00} \\
    3     & 99.74$\pm$0.21            & \textbf{100.00$\pm$0.00}       & \textbf{100.00$\pm$0.00}          & 43.48$\pm$10.02     & \underline{99.98$\pm$0.02}                               & 99.57$\pm$0.39                  & 85.70$\pm$20.43                & 99.77$\pm$0.03            & \textbf{100.00$\pm$0.00}         & \textbf{100.00$\pm$0.00} \\
    4     & 99.86$\pm$0.28            & \underline{99.88$\pm$0.07}                 & 98.15$\pm$0.85                    & 91.79$\pm$14.35     & 99.25$\pm$0.42                               & 99.87$\pm$0.14                  & 95.09$\pm$2.96                 & 98.88$\pm$0.07            & \textbf{99.93$\pm$0.09}          & 98.50$\pm$0.82           \\
    5     & 97.23$\pm$1.18            & 99.50$\pm$0.17                 & 98.36$\pm$0.05                    & 95.72$\pm$1.36      & \underline{99.52$\pm$0.28}                               & 98.62$\pm$0.86                  & 92.34$\pm$14.98                & 98.72$\pm$0.04            & \textbf{99.77$\pm$0.31}          & 97.58$\pm$0.60           \\
    6     & 99.53$\pm$1.04            & \textbf{100.00$\pm$0.00}       & \textbf{100.00$\pm$0.00}          & 98.00$\pm$2.05      & \underline{99.99$\pm$0.02}                               & 99.93$\pm$0.10                  & 83.01$\pm$28.61                & 99.18$\pm$0.11            & \textbf{100.00$\pm$0.00}         & 99.81$\pm$0.07           \\
    7     & 99.81$\pm$0.16            & \textbf{100.00$\pm$0.01}       & 99.89$\pm$0.00                    & 99.26$\pm$0.59      & 99.95$\pm$0.08                               & 99.81$\pm$0.10                  & 96.40$\pm$3.93                 & 98.61$\pm$0.12            & \underline{99.99$\pm$0.01}                   & 99.94$\pm$0.01           \\
    8     & 86.35$\pm$7.29            & 97.65$\pm$0.70                 & \textbf{98.52$\pm$0.39}           & 79.63$\pm$5.01      & 88.89$\pm$2.12                               & 77.54$\pm$7.53                  & 35.07$\pm$36.66                & 76.57$\pm$0.71            & 87.79$\pm$4.89                   & \underline{98.33$\pm$1.16}           \\
    9     & 99.37$\pm$0.63            & \textbf{100.00$\pm$0.00}       & \textbf{100.00$\pm$0.00}          & 97.57$\pm$2.33      & 99.49$\pm$0.05                               & 99.06$\pm$0.38                  & 92.68$\pm$16.75                & 99.01$\pm$0.05            & \underline{99.67$\pm$0.33}                   & \textbf{100.00$\pm$0.00} \\
    10    & 97.19$\pm$1.20            & \textbf{99.47$\pm$0.10}        & 97.98$\pm$1.70                    & 66.90$\pm$15.77     & 98.74$\pm$0.68                               & 97.04$\pm$1.87                  & 96.82$\pm$2.36                 & 93.10$\pm$0.30            & 96.53$\pm$2.55                   & \underline{99.28$\pm$0.68}           \\
    11    & 97.05$\pm$1.93            & \textbf{100.00$\pm$0.00}       & 97.86$\pm$2.27                    & 40.83$\pm$36.55     & \textbf{100.00$\pm$0.00}                     & 98.85$\pm$1.40                  & 99.03$\pm$0.03                 & 96.81$\pm$0.30            & \underline{99.76$\pm$0.21}                   & 99.74$\pm$0.09           \\
    12    & \textbf{100.00$\pm$0.00}  & 99.87$\pm$0.17                 & 99.74$\pm$0.00                    & 86.39$\pm$15.94     & \underline{99.93$\pm$0.13}                               & 99.76$\pm$0.38                  & 93.63$\pm$3.82                 & 98.84$\pm$0.20            & \textbf{100.00$\pm$0.00}         & 98.26$\pm$0.71           \\
    13    & 99.93$\pm$0.17            & \textbf{100.00$\pm$0.00}       & 98.70$\pm$0.06                    & 28.77$\pm$35.22     & \underline{99.95$\pm$0.08}                               & 99.40$\pm$0.26                  & 95.49$\pm$3.60                 & 99.38$\pm$0.08            & \textbf{100.00$\pm$0.00}         & 97.58$\pm$0.47           \\
    14    & 98.91$\pm$0.69            & \underline{99.81$\pm$0.23}                 & 94.86$\pm$1.09                    & 68.89$\pm$26.04     & \textbf{99.95$\pm$0.07}                      & 98.31$\pm$1.32                  & 98.41$\pm$1.21                 & 96.20$\pm$0.32            & 98.71$\pm$0.72                   & 98.86$\pm$0.46           \\
    15    & 77.41$\pm$9.95            & \textbf{99.95$\pm$0.07}        & 98.96$\pm$0.36                    & 61.92$\pm$10.33     & 85.44$\pm$4.44                               & 89.34$\pm$7.49                  & 94.48$\pm$4.49                 & 78.85$\pm$0.56            & 81.86$\pm$5.26                   & \underline{99.63$\pm$0.29}           \\
    16    & 99.30$\pm$0.51            & \textbf{100.00$\pm$0.00}       & 98.62$\pm$0.90                    & 42.50$\pm$23.01     & 99.57$\pm$0.67                               & 97.87$\pm$0.59                  & 96.11$\pm$0.52                 & \underline{99.69$\pm$0.06}            & 98.99$\pm$0.63                   & \textbf{100.00$\pm$0.00}  \\ \hline
    OA    & 93.60$\pm$0.93            & \textbf{99.43$\pm$0.14}        & 99.08$\pm$0.11                    & 78.34$\pm$3.52      & 95.44$\pm$0.54                               & 93.17$\pm$1.56                  & 78.31$\pm$9.20                 & 91.20$\pm$0.13            & 94.67$\pm$0.77                   & \underline{99.25$\pm$0.29}           \\
    AA    & 96.94$\pm$0.37            & \textbf{99.76$\pm$0.05}        & 98.85$\pm$0.18                    & 72.80$\pm$5.02      & 98.17$\pm$0.22                               & 96.98$\pm$0.94                  & 87.74$\pm$4.92                 & 95.79$\pm$0.04            & 97.69$\pm$0.34                   & \underline{99.22$\pm$0.19}           \\
    Kappa & 92.87$\pm$1.03            & \textbf{99.37$\pm$0.16}        & 98.97$\pm$0.12                    & 75.78$\pm$3.97      & 94.91$\pm$0.60                               & 92.41$\pm$1.73                  & 76.52$\pm$9.84                 & 90.21$\pm$0.14            & 94.06$\pm$0.85                   & \underline{99.17$\pm$0.33}           \\ \hline
    \hline
    \end{tabular}
    \end{table*}
  
\subsubsection{Results on \textit{Houston University} Dataset} 
\autoref{table8} shows the classification results obtained by different methods on the \textit{Houston University} dataset. We can observe that the proposed ConGCN outperforms all the other compared methods by a substantial margin in terms of OA, AA, and Kappa coefficient. In addition, the standard deviations of ConGCN are relatively small as well. Another notable fact is that our proposed ConGCN outperforms other methods in seven land cover categories, which reveals the effectiveness of our method. Generally, our proposed ConGCN achieves stable and encouraging performance.

\begin{figure*} \centering    
    \subfigure[]
    {
      \label{fig12:a}     
    \includegraphics[width=8.6cm]{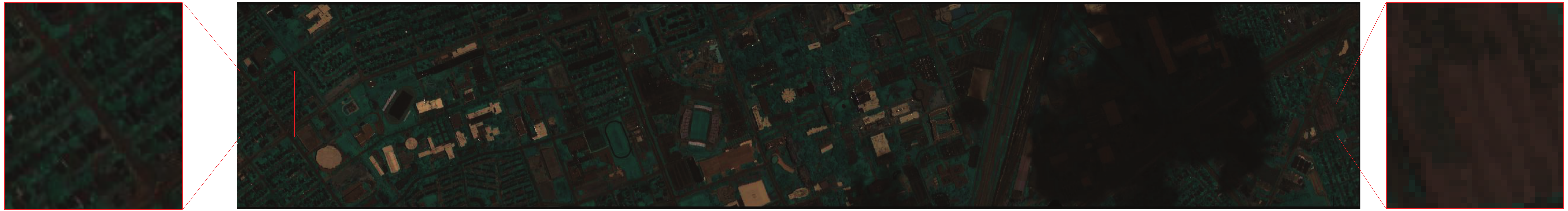}  
    }
    \subfigure[]
    { 
    \label{fig12:b}
    \includegraphics[width=8.6cm]{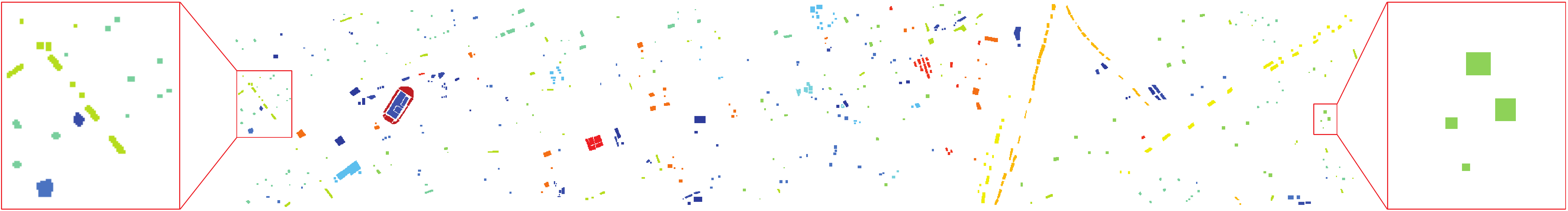}     
    }    
    \subfigure[]
    { 
    \label{fig12:c}     
    \includegraphics[width=8.6cm]{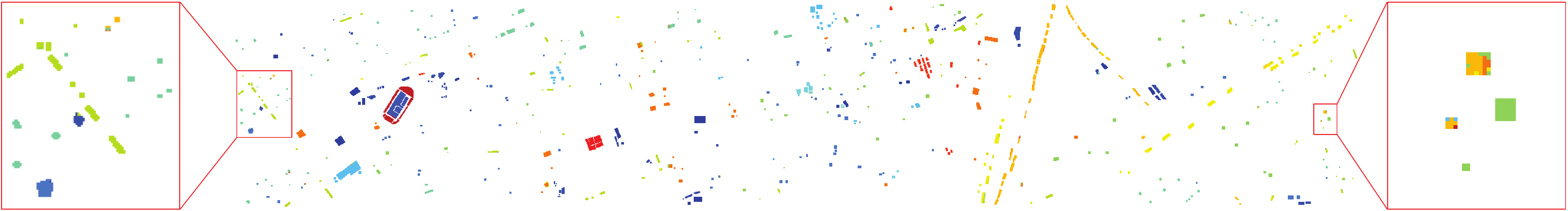}     
    }
    \subfigure[]
    { 
    \label{fig12:d}     
    \includegraphics[width=8.6cm]{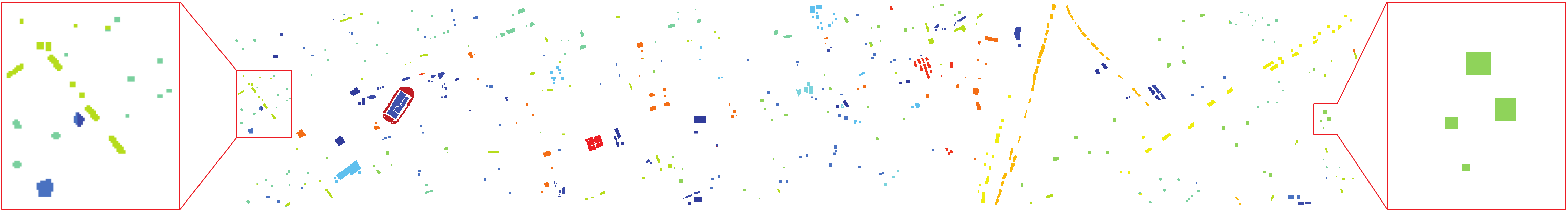}     
    }
    \subfigure[]
    { 
    \label{fig12:e}     
    \includegraphics[width=8.6cm]{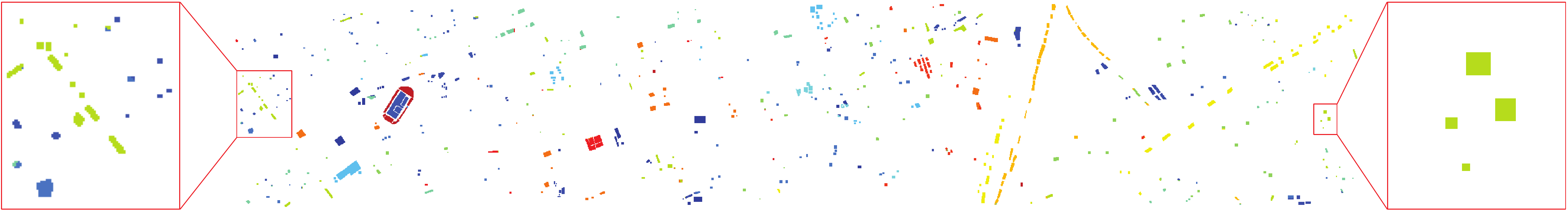}     
    }
    \subfigure[]
    { 
    \label{fig12:f}     
    \includegraphics[width=8.6cm]{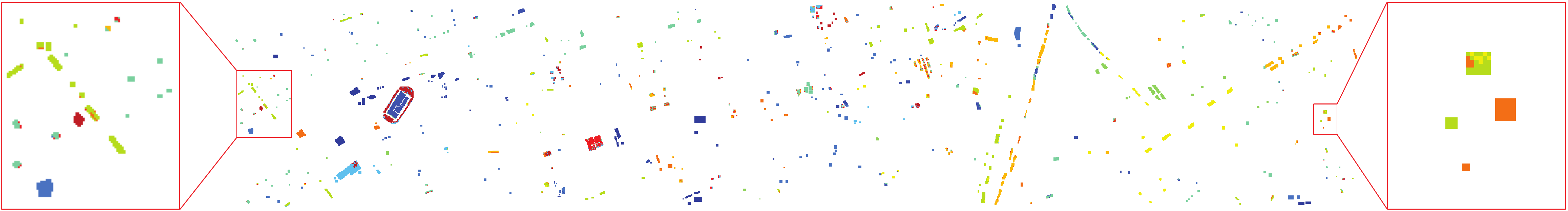}     
    }
    \subfigure[]
    { 
    \label{fig12:g}     
    \includegraphics[width=8.6cm]{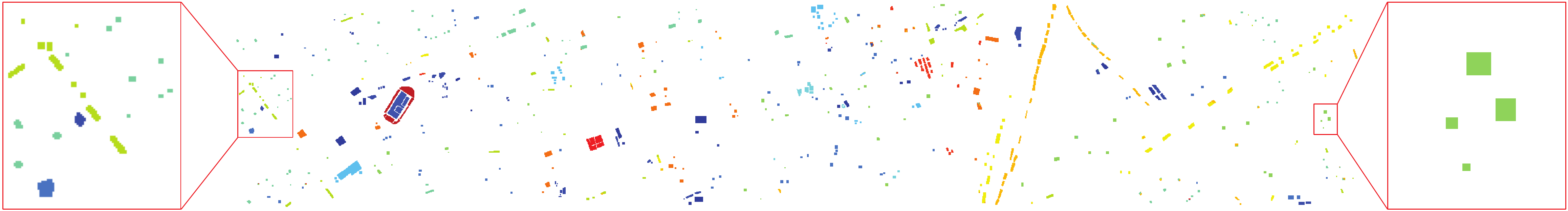}     
    }
    \subfigure[]
    { 
    \label{fig12:h}     
    \includegraphics[width=8.6cm]{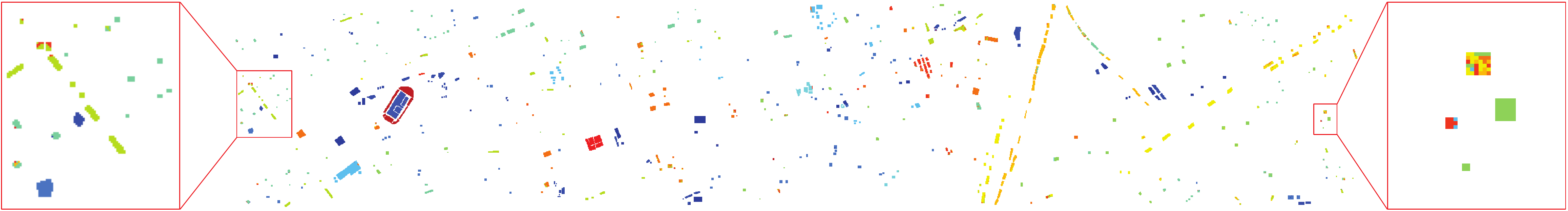}     
    }
    \subfigure[]
    { 
    \label{fig12:i}     
    \includegraphics[width=8.6cm]{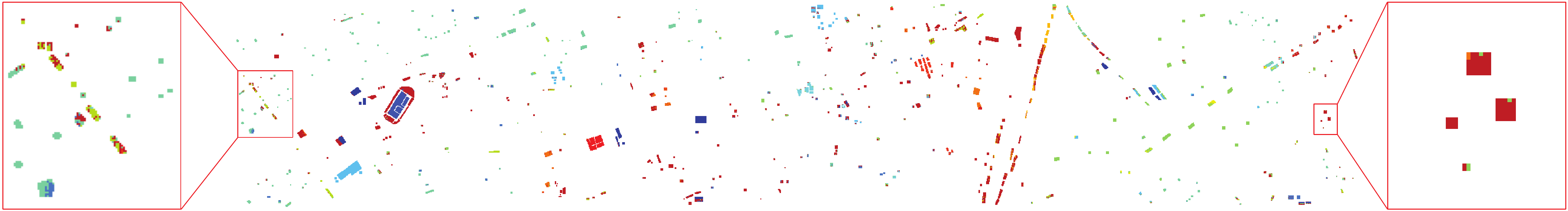}     
    }
    \subfigure[]
    { 
    \label{fig12:j}     
    \includegraphics[width=8.6cm]{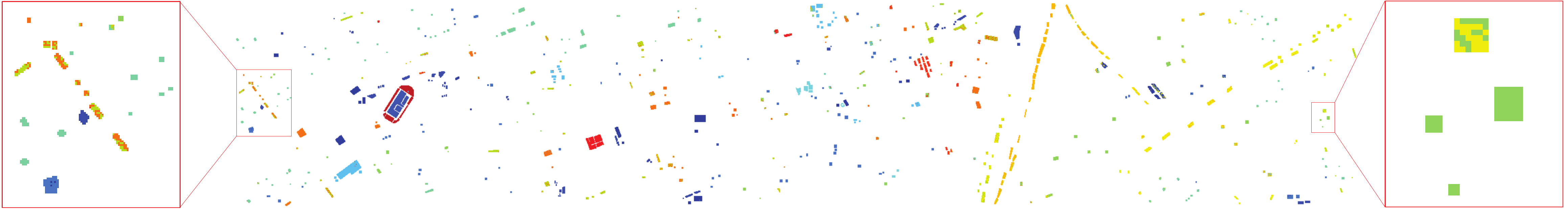}     
    }
    \subfigure[]
    { 
    \label{fig12:k}     
    \includegraphics[width=8.6cm]{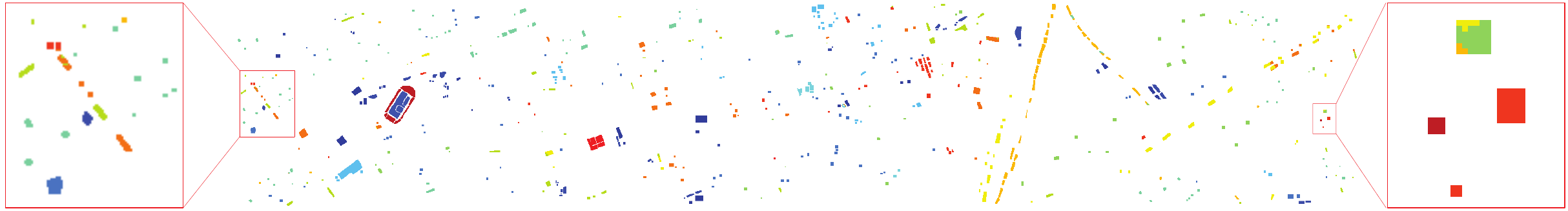}     
    }
    \subfigure[]
    { 
    \label{fig12:l}     
    \includegraphics[width=8.6cm]{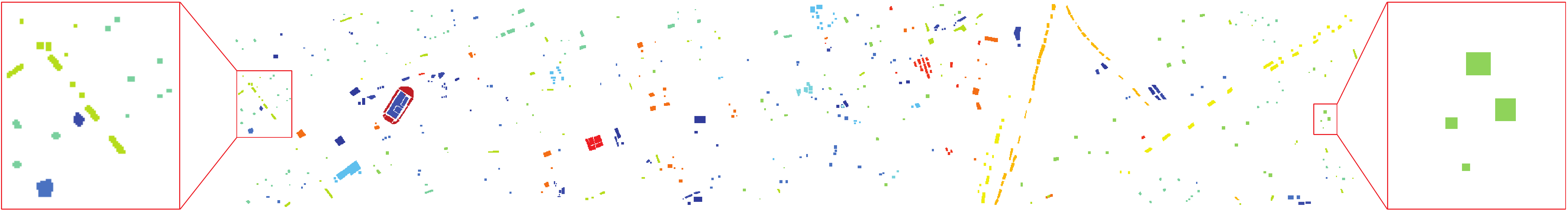}     
    }
    \caption{Classification maps obtained by different methods on \textit{Houston University} dataset. (a) False-color image. (b) Ground-truth map. (c) SMR. (d) MSSGU. (e) SGL. (f) SSCL. (g) A$^{2}$S$^{2}$K. (h) ASSMN. (i) ADGAN. (j) MFL. (k) JSDF. (l) ConGCN (Proposed). In (a)–(l), zoomed-in views of the regions are denoted by red boxes.}     
    \label{fig12}
    \end{figure*}

The classification results produced by different methods are visualized in Fig.~\ref{fig12}. As can be seen, there are noticeable errors in the classification maps of other methods (see the zoomed-in regions of Fig.~\ref{fig12}). By contrast, our proposed ConGCN (Fig.~\ref{fig12:l}) achieves good classification result, which confirms the advantage of our method.\vspace{-1em}

\begin{table*}[]
  \scriptsize
  \centering
  \caption{Per-class accuracy, OA, AA (\%), and Kappa coefficient achieved by different methods on \textit{Houston University} dataset. The best and second best records in each row are \textbf{bolded} and \underline{underlined}, respectively.}
  \label{table8}
  \begin{tabular}{c|c|c|c|c|c|c|c|c|c|c}
  \hline
  \hline
  ID    & SMR \cite{wang2021toward} & MSSGU \cite{liu2021multilevel} & SGL \cite{sellars2020superpixel} & SSCL \cite{9664575}  & A$^{2}$S$^{2}$K \cite{roy2020attention}      & ASSMN \cite{wang2020adaptive1}  & ADGAN \cite{wang2020adaptive} & MFL \cite{li2014multiple} & JSDF \cite{bo2015hyperspectral}  & ConGCN                   \\ \hline
  1     & 92.64$\pm$3.13            & \textbf{98.35$\pm$2.46}        & 89.13$\pm$6.46                   & 78.18$\pm$8.63       & \underline{97.80$\pm$1.49}                               & 94.52$\pm$2.79                  & 60.27$\pm$24.21               & 91.00$\pm$0.90            & 97.41$\pm$1.21                   & 97.79$\pm$0.92          \\
  2     & 98.21$\pm$1.01            & 91.81$\pm$2.24                 & 75.79$\pm$6.67                   & 42.96$\pm$11.79      & \underline{98.53$\pm$1.02}                      & 97.46$\pm$2.53                  & 23.69$\pm$16.97               & 94.97$\pm$0.62            & \textbf{99.48$\pm$0.25}                   & 97.35$\pm$0.59           \\
  3     & 97.14$\pm$1.97            & 99.60$\pm$0.34                 & 99.60$\pm$0.04                   & 95.91$\pm$1.20       & \underline{99.76$\pm$0.48}                      & 98.22$\pm$0.29                  & 94.49$\pm$6.77                & 99.74$\pm$0.01            & \textbf{99.88$\pm$0.22}                   & 98.25$\pm$0.32           \\
  4     & 98.28$\pm$1.78            & \underline{99.68$\pm$0.48}                 & 75.16$\pm$2.67                   & 66.31$\pm$7.73       & 97.00$\pm$1.08                               & 92.75$\pm$7.85                  & 42.58$\pm$16.45               & 93.14$\pm$0.45            & 98.22$\pm$2.80                   & \textbf{99.74$\pm$0.52}  \\
  5     & 97.75$\pm$1.01            & \textbf{100.00$\pm$0.00}       & 99.05$\pm$0.47                   & 70.72$\pm$13.86      & 98.55$\pm$0.94                               & 97.08$\pm$1.26                  & 53.16$\pm$16.55               & 98.36$\pm$0.15            & \textbf{100.00$\pm$0.00}         & \underline{99.79$\pm$0.33}           \\
  6     & 98.58$\pm$1.23            & \textbf{99.63$\pm$0.65}        & 96.47$\pm$2.54                   & 10.62$\pm$15.89      & 98.70$\pm$1.81                               & 98.07$\pm$1.06                  & 91.83$\pm$7.87                & 97.24$\pm$0.40            & \underline{99.32$\pm$1.09}                   & 97.29$\pm$0.00           \\
  7     & 92.93$\pm$2.78            & \underline{95.33$\pm$1.36}                 & 69.51$\pm$6.61                   & 64.06$\pm$10.15      & 94.74$\pm$3.38                               & 86.52$\pm$5.93                  & 57.26$\pm$33.71               & 88.02$\pm$0.52            & 91.93$\pm$4.91                   & \textbf{98.18$\pm$0.43}  \\
  8     & 81.22$\pm$3.26            & 90.30$\pm$2.88                 & 67.83$\pm$5.24                   & 19.08$\pm$10.07      & \underline{95.53$\pm$5.09}                               & 73.38$\pm$3.93                  & 31.79$\pm$11.21               & 64.28$\pm$0.71            & 68.82$\pm$6.16                   & \textbf{97.75$\pm$1.38}  \\
  9     & 85.01$\pm$4.59            & 90.66$\pm$3.00                 & 75.73$\pm$4.31                   & 67.63$\pm$10.58      & \underline{92.19$\pm$5.37}                               & 72.18$\pm$7.57                  & 26.59$\pm$15.06               & 67.91$\pm$0.57            & 69.47$\pm$8.56                   & \textbf{97.77$\pm$1.28}  \\
  10    & 90.37$\pm$5.50            & \textbf{100.00$\pm$0.00}       & \underline{97.14$\pm$7.22}                   & 29.60$\pm$15.16      & 83.72$\pm$2.76                               & 82.47$\pm$5.35                  & 27.29$\pm$23.47               & 87.64$\pm$0.95            & 85.63$\pm$9.32                   & \textbf{100.00$\pm$0.00}   \\
  11    & 91.74$\pm$7.04            & \textbf{99.99$\pm$0.02}        & 94.91$\pm$2.77                   & 50.08$\pm$7.59       & 90.78$\pm$4.37                               & 82.91$\pm$6.05                  & 48.64$\pm$24.83               & 89.20$\pm$0.47            & 94.51$\pm$3.82                   & \underline{98.91$\pm$0.96}           \\
  12    & 85.85$\pm$5.03            & \underline{92.78$\pm$2.08}                 & 79.31$\pm$5.19                   & 21.69$\pm$9.27       & 91.31$\pm$6.98                               & 76.80$\pm$7.85                  & 73.49$\pm$27.32               & 77.65$\pm$0.46            & 84.33$\pm$5.33                   & \textbf{99.45$\pm$1.28}  \\
  13    & 83.37$\pm$7.43            & 86.76$\pm$2.48                 & 92.31$\pm$2.84                   & 19.68$\pm$14.86      & 96.59$\pm$3.49                               & 70.46$\pm$8.76                  & 68.81$\pm$17.97               & 80.76$\pm$0.40            & \underline{98.10$\pm$1.28}                   & \textbf{98.69$\pm$2.00}  \\
  14    & 96.21$\pm$8.18            & \textbf{100.00$\pm$0.00}       & 98.45$\pm$1.02                   & 69.96$\pm$12.94      & \textbf{100.00$\pm$0.00}                     & 98.94$\pm$1.07                  & 90.29$\pm$8.00                & 98.01$\pm$0.29            & \textbf{100.00$\pm$0.00}         & \underline{99.38$\pm$1.86}           \\
  15    & 99.29$\pm$0.85            & \textbf{99.99$\pm$0.04}        & 96.81$\pm$0.45                   & 74.38$\pm$11.62      & 99.13$\pm$0.72                               & 98.70$\pm$0.96                  & 93.60$\pm$6.11                & 98.47$\pm$0.11            & \underline{99.86$\pm$0.36}                   & 98.19$\pm$0.32           \\ \hline
  OA    & 91.89$\pm$1.93            & \underline{96.03$\pm$0.52}                 & 84.71$\pm$1.28                   & 52.69$\pm$2.23       & 94.37$\pm$0.74                               & 86.60$\pm$3.33                  & 51.57$\pm$9.80                & 86.66$\pm$0.13            & 90.51$\pm$0.95                   & \textbf{98.61$\pm$0.37}  \\
  AA    & 92.57$\pm$1.96            & \underline{96.33$\pm$0.45}                 & 87.15$\pm$1.04                   & 52.06$\pm$2.82       & 95.62$\pm$0.46                               & 88.03$\pm$3.02                  & 58.92$\pm$8.97                & 88.43$\pm$0.11            & 92.46$\pm$0.75                   & \textbf{98.57$\pm$0.41}  \\
  Kappa & 91.22$\pm$2.09            & \underline{95.71$\pm$0.56}                 & 83.49$\pm$1.38                   & 48.86$\pm$2.40       & 93.90$\pm$0.80                               & 85.50$\pm$3.60                  & 48.34$\pm$10.36               & 85.56$\pm$0.14            & 89.74$\pm$1.03                   & \textbf{98.49$\pm$0.40}  \\ \hline
  \hline
  \end{tabular}
  \end{table*}

\subsection{Impact of the Number of Labeled Examples}
\begin{figure*} \centering
  \subfigure[]
  {
  \label{fig13:a}     
  \includegraphics[width=4.2cm]{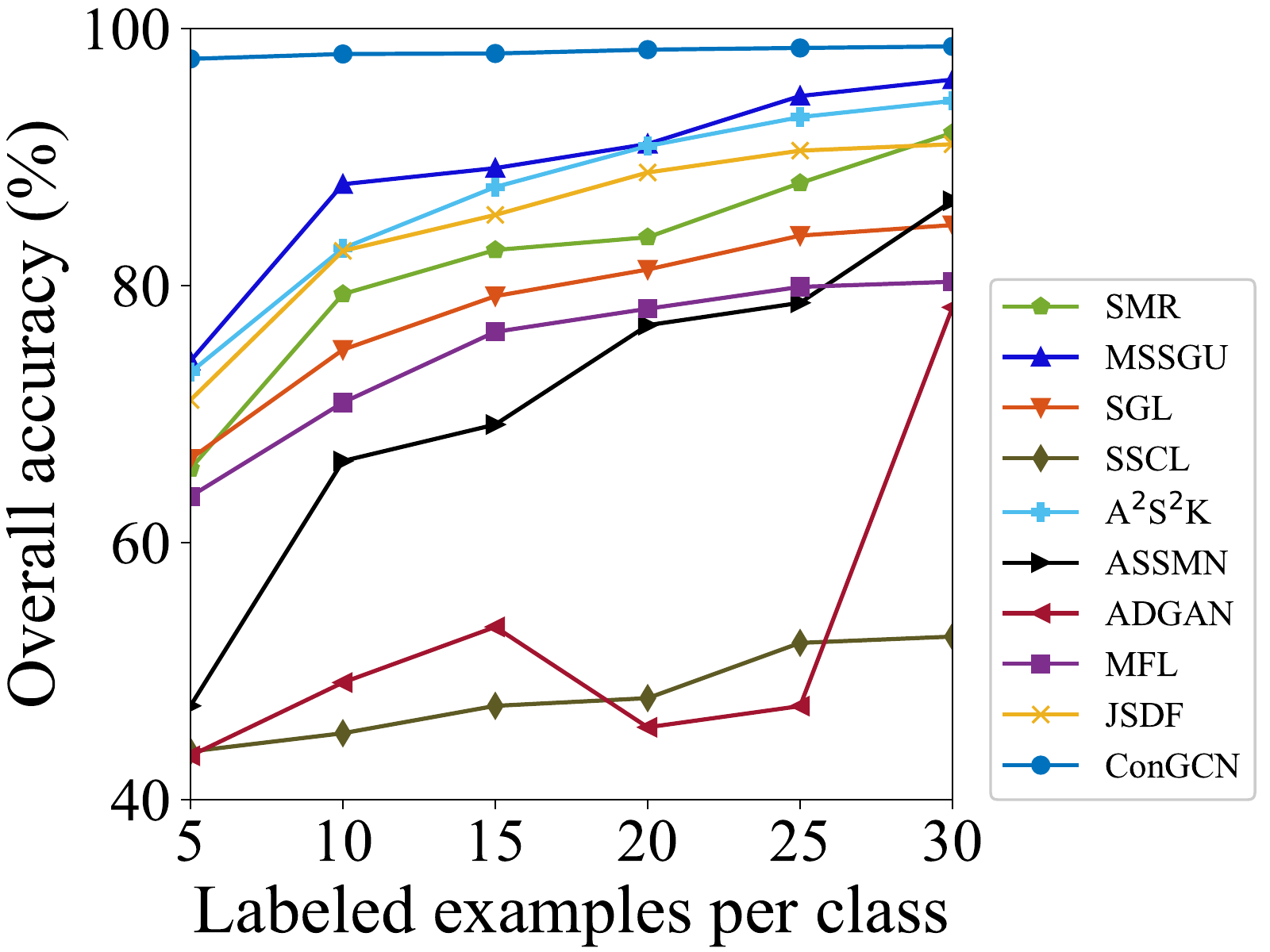}  
  }
  \subfigure[]
  {
  \label{fig13:b}
  \includegraphics[width=4.2cm]{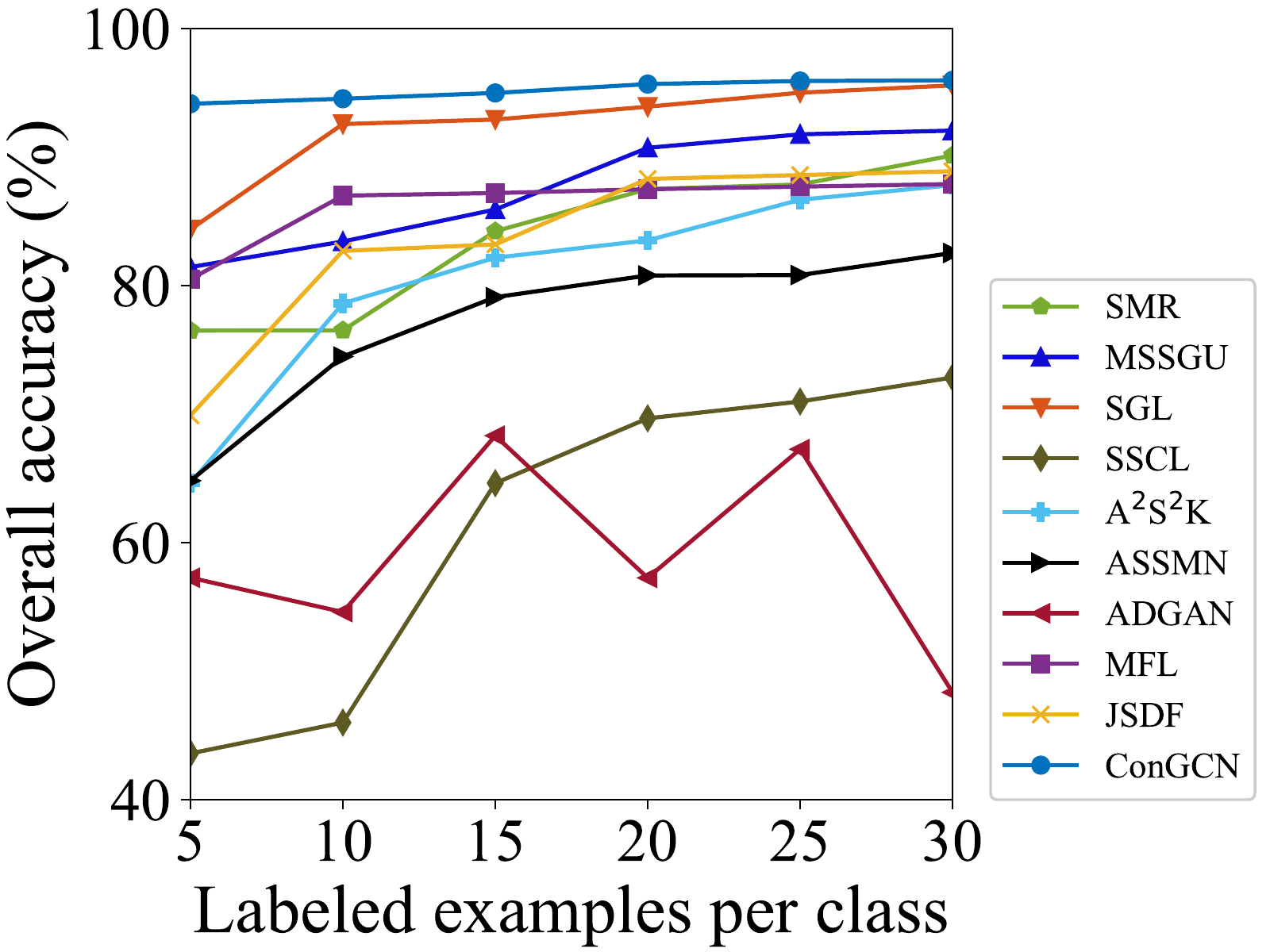}     
  }
  \subfigure[]
  { 
  \label{fig13:c}     
  \includegraphics[width=4.2cm]{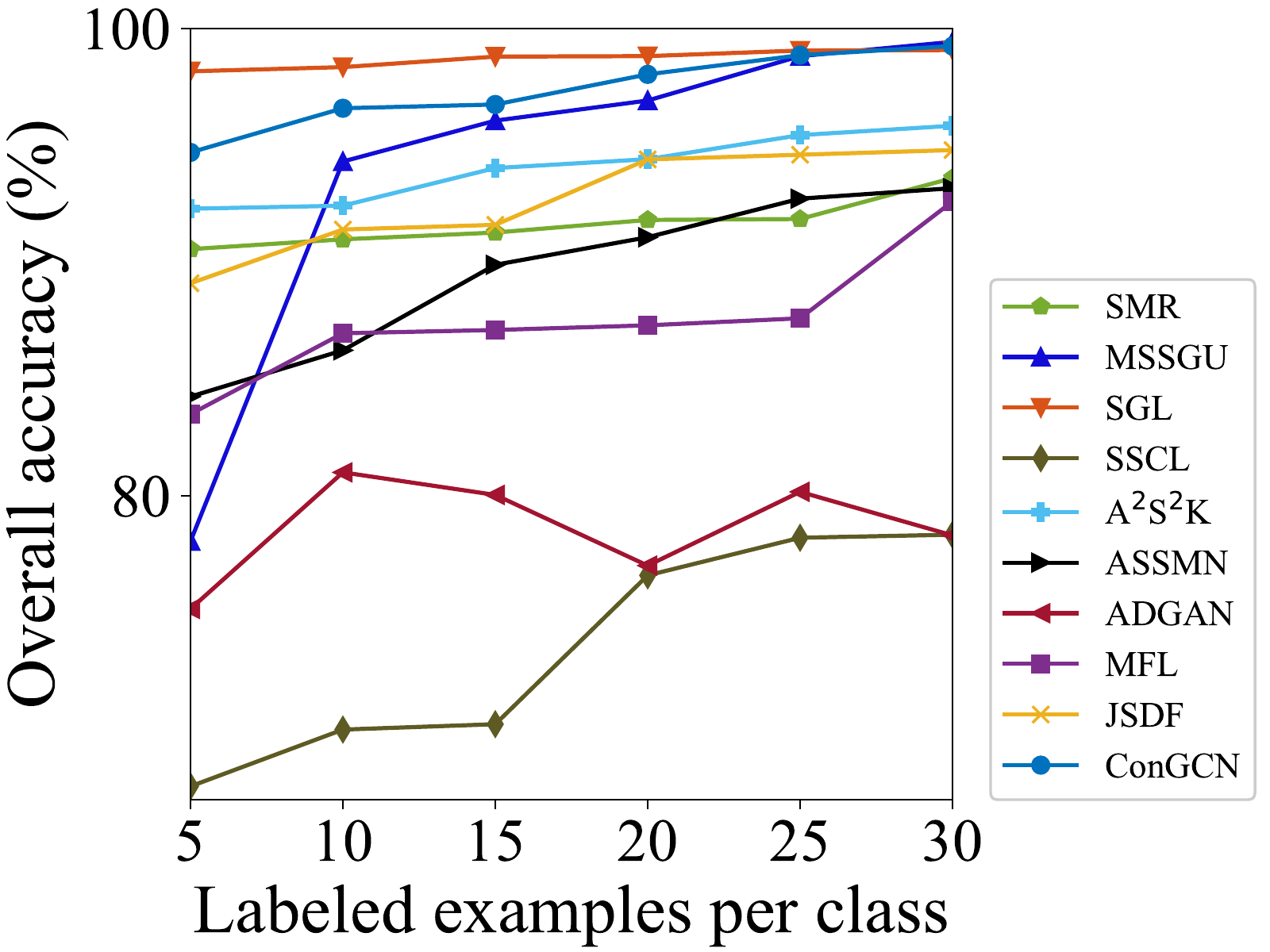}     
  }
  \subfigure[]
  {
  \label{fig13:d}     
  \includegraphics[width=4.2cm]{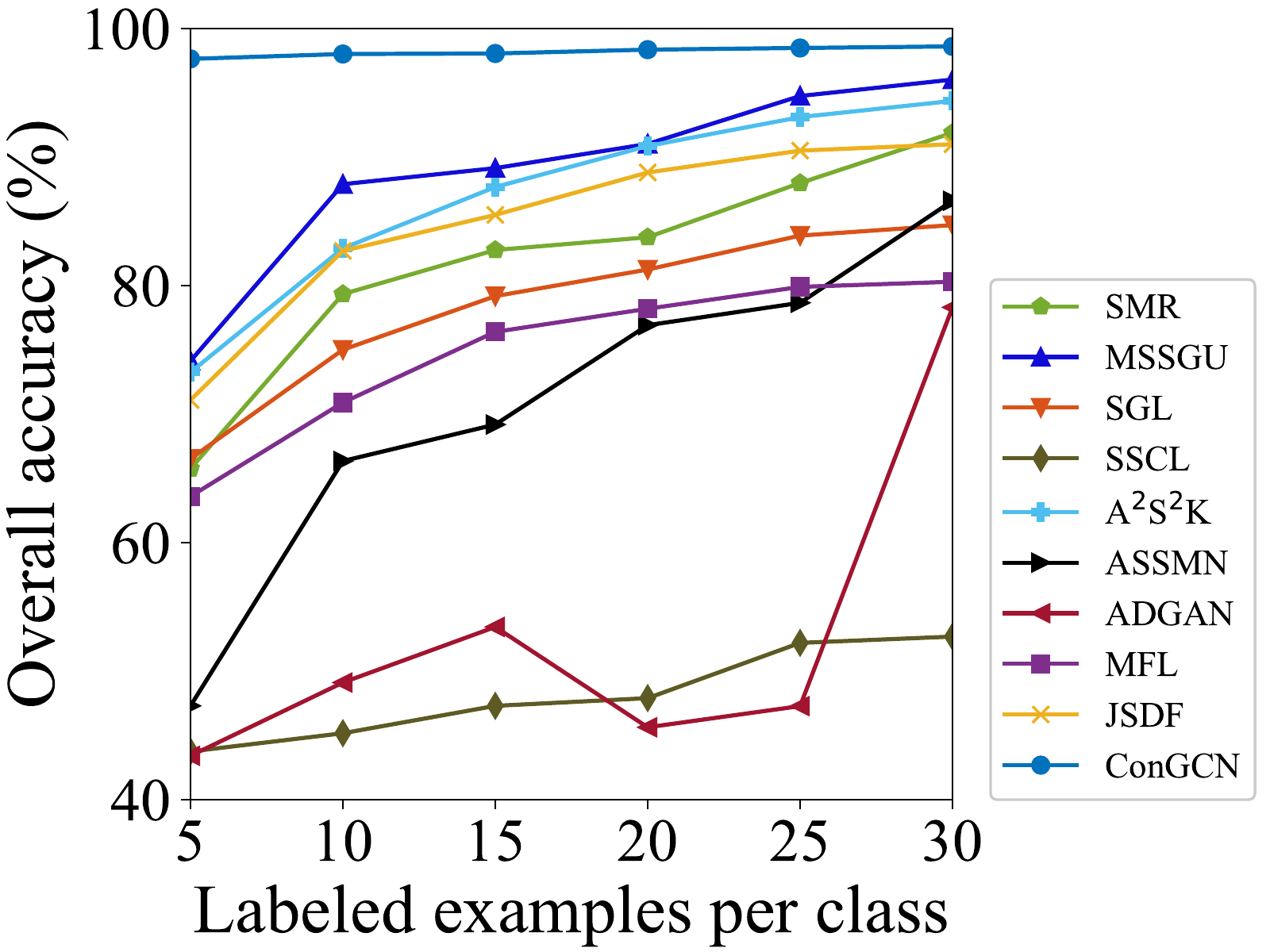}     
  }
  \caption{Overall accuracies of various methods under different numbers of labeled examples per class. (a) \textit{Indian Pines} dataset. (b) \textit{University of Pavia} dataset. (c) \textit{Salinas} dataset. (d) \textit{Houston University} dataset.}     
  \label{fig13}
  \end{figure*}

In this subsection, classification accuracies of the proposed ConGCN and other methods under different numbers of labeled examples are shown in Fig.~\ref{fig13}. We vary the number of labeled examples per class from 5 to 30 with an interval of 5 and report the OA gained by all the methods on four datasets. As observed in Fig.~\ref{fig13}, except ADGAN, the classification performances of all other methods can be generally improved by increasing the number of labeled examples. Due to the instability of GAN during training~\cite{qin2020training}, the OA of ADGAN could drop significantly on the \textit{University of Pavia} (Fig.~\ref{fig13:b}) and the \textit{Salinas} datasets (Fig.~\ref{fig13:c}) even when the number of labeled examples increases. Thanks to the exploration of the supervision signals from both spectral and spatial aspects of HSI, our ConGCN still achieves relatively high OA even if the labeled examples are quite limited (\textit{i.e.}, five or ten labeled examples per class), which suggests the good stability of ConGCN in HSI classification.\vspace{-1em}

\subsection{Ablation Study}
As mentioned in the introduction (Section~\ref{Introduction}), the proposed ConGCN contains two parts that are critical for enriching the supervision signals from the spectral-spatial information of HSI, \textit{i.e.}, the semi-supervised contrastive loss function and the graph generative loss function. We use the four datasets to shed light on the contributions of these two components, where the number of labeled pixels per class is kept identical to the above experiments in Section~\ref{ExperimentalResults}. Every time we report the OA, AA, and kappa coefficient obtained by ConGCN without one of the aforementioned loss functions. For simplicity, ``w/o Closs" and ``w/o Gloss" indicate the reduced models by removing the contrastive loss function $\mathcal{L}_\text{ssc}$ and the graph generative loss $\mathcal{L}_{\text{g}^{2}}$, respectively. In addition, we also investigate the effectiveness of spatial-level and spectral-level graph augmentation. To be concrete, we utilize ``w/o SpaAug", ``w/o SpeAug'', and ``w/o SpaAug and SpeAug" to indicate the reduced models that remove the spatial-level graph augmentation, spectral-level graph augmentation, and spatial-level and spectral-level graph augmentation, respectively. Concretely, \autoref{table_IP_ablation}, \autoref{table_paviaU_ablation}, \autoref{table_Salinas_ablation}, and \autoref{table_UH_ablation} exhibit the comparative results on four datasets. It is apparent that the overall classification accuracy will decline when any one of the aforementioned components is removed. It reveals that each component makes an essential contribution to boosting the classification performance. We can also observe that the reduced model ``w/o Gloss" consistently achieves higher OA, AA, and Kappa coefficient than ``w/o Closs" on the four datasets, which validates that the semi-supervised contrastive loss makes a greater contribution than the generative loss to performance improvement. Another interesting observation is that the performance degradation of ``w/o SpaAug and SpeAug" is more conspicuous than that of ``w/o SpaAug" or ``w/o SpeAug'' on four datasets. This indicates that spatial-level and spectral-level graph augmentation work collaboratively and complementarily to boost the performance of contrastive learning.

\begin{table}[t]
  \centering
  \caption{Per-class accuracy, OA, AA (\%), and Kappa coefficient achieved by different settings on \textit{Indian Pines} dataset. The best records in each row are \textbf{bolded}.}
  \label{table_IP_ablation}
  \scalebox{0.6}{
  \begin{tabular}{c|c|c|c|c|c|c}
  \hline
  \hline
  metrics    & w/o CLoss            & w/o GLoss                & w/o SpaAug              & w/o SpeAug              & \begin{tabular}[c]{@{}c@{}}w/o SpaAug\\and SpeAug\end{tabular} & ConGCN                   \\ \hline
  OA    & 96.44$\pm$0.55           & 96.46$\pm$0.68           & 96.58$\pm$0.82           & 96.56$\pm$0.68            & 96.50$\pm$0.75                        & \textbf{96.74$\pm$0.50}  \\
  AA    & 97.18$\pm$0.33           & \textbf{97.28$\pm$0.37}           & 97.25$\pm$0.43           & 97.25$\pm$0.36            & 97.27$\pm$0.34                        & \textbf{97.28$\pm$0.29}  \\
  Kappa & 95.92$\pm$0.63           & 95.95$\pm$0.78           & 96.09$\pm$0.93           & 96.06$\pm$0.77            & 96.00$\pm$0.85                        & \textbf{96.27$\pm$0.57}  \\ \hline
  \hline
  \end{tabular}
  }
  \end{table}

\begin{table}[t]
  \centering
  \caption{Per-class accuracy, OA, AA (\%), and Kappa coefficient achieved by different settings on \textit{University of Pavia} dataset. The best records in each row are \textbf{bolded}.}
  \label{table_paviaU_ablation}
  \scalebox{0.6}{
  \begin{tabular}{c|c|c|c|c|c|c}
  \hline
  \hline
  metrics    & w/o CLoss            & w/o GLoss                & w/o SpaAug              & w/o SpeAug              & \begin{tabular}[c]{@{}c@{}}w/o SpaAug\\and SpeAug\end{tabular} & ConGCN                   \\ \hline
  OA    & 95.05$\pm$1.13           & 95.32$\pm$1.54           & 95.37$\pm$1.62           & 95.79$\pm$1.64            & 95.36$\pm$1.52                        & \textbf{95.97$\pm$0.90}  \\
  AA    & 94.95$\pm$0.88           & 95.01$\pm$0.71           & \textbf{95.70$\pm$0.39}           & 95.56$\pm$0.65            & 95.22$\pm$0.71               & 95.14$\pm$0.57           \\
  Kappa & 93.51$\pm$1.47           & 93.86$\pm$1.99           & 93.94$\pm$2.06           & 94.48$\pm$2.09            & 93.93$\pm$1.95                        & \textbf{94.69$\pm$1.17}  \\ \hline
  \hline
  \end{tabular}
  }
  \end{table}

\begin{table}[t]
\centering
\caption{Per-class accuracy, OA, AA (\%), and Kappa coefficient achieved by different settings on \textit{Salinas} dataset. The best records in each row are \textbf{bolded}.}
\label{table_Salinas_ablation}
\scalebox{0.6}{
\begin{tabular}{c|c|c|c|c|c|c}
\hline
\hline 
metrics    & w/o CLoss            & w/o GLoss                & w/o SpaAug              & w/o SpeAug              & \begin{tabular}[c]{@{}c@{}}w/o SpaAug\\and SpeAug\end{tabular} & ConGCN                   \\ \hline
OA    & 97.80$\pm$4.06       & 98.55$\pm$1.35           & 99.09$\pm$0.31           & 99.02$\pm$0.78            & 99.00$\pm$0.43                        & \textbf{99.25$\pm$0.29}  \\
AA    & 98.50$\pm$2.06       & 98.90$\pm$0.74           & 99.20$\pm$0.13           & 99.07$\pm$0.42            & 99.08$\pm$0.41                        & \textbf{99.22$\pm$0.19}  \\
Kappa & 97.56$\pm$4.48       & 98.39$\pm$1.50           & 98.99$\pm$0.35           & 98.91$\pm$0.86            & 98.88$\pm$0.47                        & \textbf{99.17$\pm$0.33}  \\ \hline
\hline
\end{tabular}
}
\end{table}

\begin{table}[t]
  \centering
  \caption{Per-class accuracy, OA, AA (\%), and Kappa coefficient achieved by different settings on \textit{Houston University} dataset. The best records in each row are \textbf{bolded}.}
  \label{table_UH_ablation}
  \scalebox{0.6}{
  \begin{tabular}{c|c|c|c|c|c|c}
  \hline
  \hline
  metrics    & w/o CLoss            & w/o GLoss                & w/o SpaAug              & w/o SpeAug              & \begin{tabular}[c]{@{}c@{}}w/o SpaAug\\and SpeAug\end{tabular} & ConGCN                   \\ \hline
  OA    & 97.99$\pm$2.06           & 98.32$\pm$0.71           & 98.53$\pm$0.34           & 98.48$\pm$0.41            & 98.41$\pm$0.34                        & \textbf{98.61$\pm$0.37}  \\
  AA    & 97.93$\pm$2.25           & 98.33$\pm$0.69           & 98.53$\pm$0.29           & 98.49$\pm$0.38            & 98.41$\pm$0.37                        & \textbf{98.57$\pm$0.41}  \\
  Kappa & 97.82$\pm$2.22           & 98.18$\pm$0.77           & 98.41$\pm$0.37           & 98.36$\pm$0.44            & 98.28$\pm$0.37                        & \textbf{98.49$\pm$0.40}  \\ \hline
  \hline
  \end{tabular}
  }
  \end{table}

\section{Conclusion}
In this paper, we propose a ConGCN algorithm  for HSI classification. To improve feature representation ability, we explore the supervision signals based on the spectral and spatial information of HSI. Specifically, we devise a semi-supervised contrastive loss to exploit the supervision contained in the spectral signatures of image regions. Meanwhile, we develop a graph generative loss to explore supplementary supervision signals from the spatial relations among image regions. Last but not least, we devise an adaptive graph augmentation technique via incorporating the spectral-spatial priors to boost the performance of contrastive learning. As a consequence, the expressive power of the generated representation can be enhanced, which leads to the improved classification results. Experimental results on four real-world HSI datasets validate the effectiveness of our proposed ConGCN.

\bibliographystyle{IEEEtran}
\bibliography{cite/cite}



\end{document}


\title{Hyperspectral Image Classification With Contrastive Graph Convolutional Network (Appendix)}

\author{~Wentao~Yu,~\IEEEmembership{Student Member,~IEEE},~Sheng Wan,~Jian~Yang,~\IEEEmembership{Member,~IEEE}~and Chen~Gong,~\IEEEmembership{Member,~IEEE}

\thanks{The authors are with the PCA Laboratory, the Key Laboratory of Intelligent Perception and Systems for High-Dimensional Information of Ministry of Education, the Jiangsu Key Laboratory of Image and Video Understanding for Social Security, and the School of Computer Science and Engineering, Nanjing University of Science and Technology, Nanjing 210094, China (e-mail: wentao.yu@njust.edu.cn; wansheng315@hotmail.com; csjyang@njust.edu.cn; chen.gong@njust.edu.cn).}
}

\markboth{IEEE TRANSACTIONS ON GEOSCIENCE AND REMOTE SENSING, ~Vol.~XX, No.~X, XX~202X}{YU \MakeLowercase{\textit{et al.}}: HSI CLASSIFICATION WITH ConGCN}

\maketitle

\IEEEpeerreviewmaketitle

In this appendix, we provide some details on the description of the adopted hyperspectral datasets.\vspace{-1em}

\section{Datasets}
The performances of all methods are evaluated on four datasets including \textit{Salinas}, \textit{Indian Pines}, \textit{University of Pavia}, and \textit{Houston University}, which will be described as follows:\vspace{-1em}
\subsection{\textit{Salinas}}
\begin{figure}[]
	\centering
	\includegraphics[width=8cm]{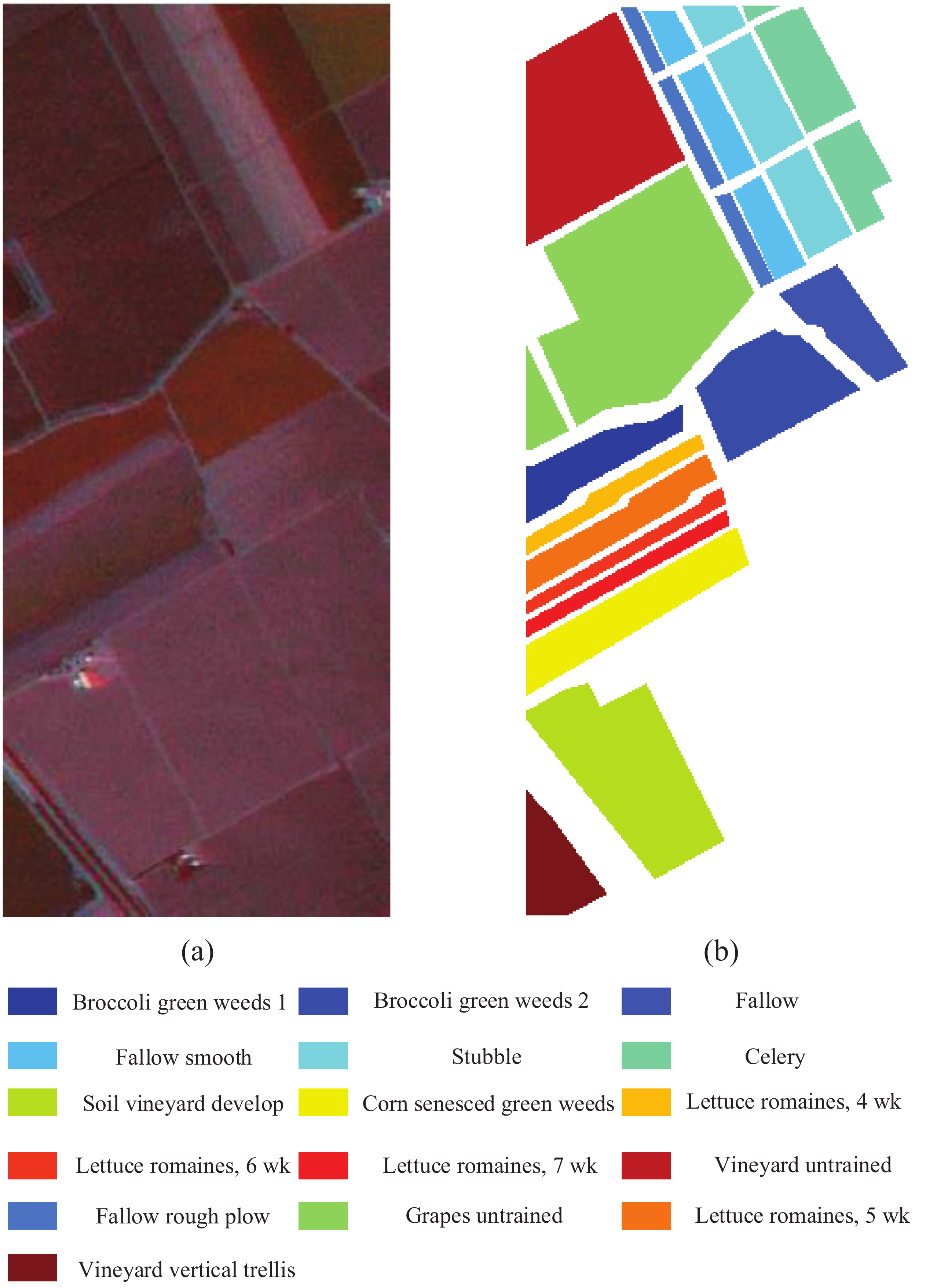}
	\caption{The \textit{Salinas} dataset. (a) False-color image. (b) Ground-truth map.}
	\label{fig7}
\end{figure}

The \textit{Salinas} dataset was collected by the AVIRIS sensor in Salinas Valley, CA, USA. This dataset comprises $512 \times 217$ pixels with a spatial resolution of 3.7 m $\times$ 3.7~m. After removing 20 water absorption spectral bands as usual, there are 204 remaining spectral bands. There are 16 land cover categories in this dataset including ``Grapes untrained", ``Soil vineyard develop", ``Vineyard vertical trellis", and so on. The false-color image and ground-truth map of the \textit{Salinas} dataset are exhibited in Fig. \ref{fig7}. The quantities of labeled and unlabeled pixels of various land cover categories used in our experiments are listed in \autoref{table3}.\vspace{-1em}

\begin{table}[t]
  \vspace{-2em}
  \scriptsize
  \centering
  \caption{Quantities of labeled and unlabeled pixels of all land cover categories in \textit{Salinas} dataset.}
  \label{table3}
  \begin{tabular}{c|c|c|c}
  \hline
  \hline
  ID & Class                     & \#Labeled & \#Unlabeled \\ \hline
  1  & Broccoli green weeds 1    & 30      & 1979      \\
  2  & Broccoli green weeds 2    & 30      & 3696      \\
  3  & Fallow                    & 30      & 1946      \\
  4  & Fallow rough plow         & 30      & 1364      \\
  5  & Fallow smooth             & 30      & 2648      \\
  6  & Stubble                   & 30      & 3929      \\
  7  & Celery                    & 30      & 3549      \\
  8  & Grapes untrained          & 30      & 11241     \\
  9  & Soil vineyard develop     & 30      & 6173      \\
  10 & Corn senesced green weeds & 30      & 3248      \\
  11 & Lettuce romaines, 4 wk    & 30      & 1038      \\
  12 & Lettuce romaines, 5 wk    & 30      & 1897      \\
  13 & Lettuce romaines, 6 wk    & 30      & 886       \\
  14 & Lettuce romaines, 7 wk    & 30      & 1040      \\
  15 & Vineyard untrained        & 30      & 7238      \\
  16 & Vineyard vertical trellis & 30      & 1777      \\ \hline
  \hline
  \end{tabular}
  \end{table}

\subsection{\textit{Indian Pines}}
\begin{figure}[]
	\centering
	\includegraphics[width=8cm]{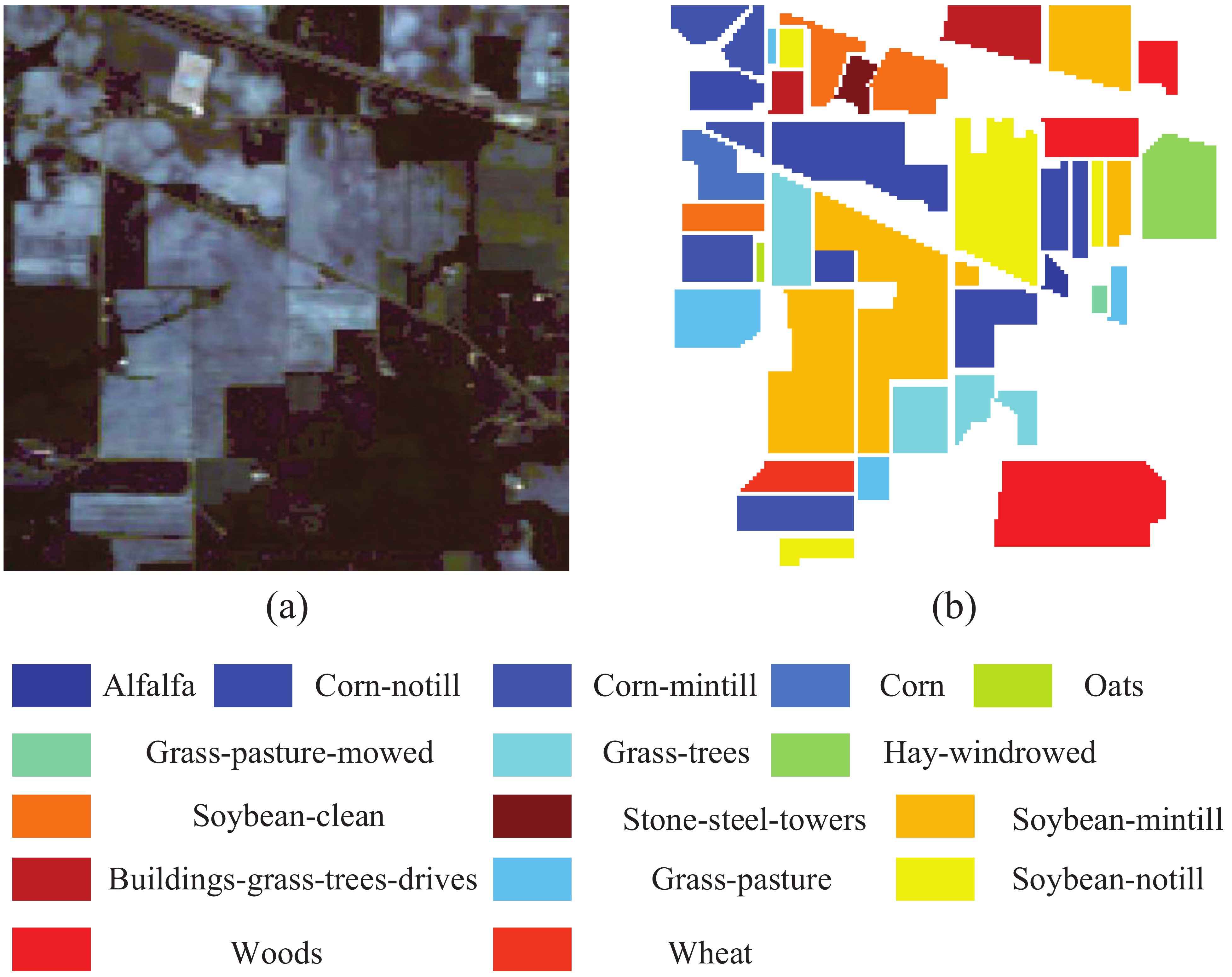}
	\caption{The \textit{Indian Pines} dataset. (a) False-color image. (b) Ground-truth map.}
	\label{fig5}
\end{figure}

The \textit{Indian Pines} dataset was collected in northwestern India, in 1992. This dataset consists of $145 \times 145$ pixels with a spatial resolution of 20 m $\times$ 20 m. In this dataset, there are 220 spectral bands ranging from 0.4 to 2.5 $\upmu$m. As a usual step, 20 water absorption and noisy spectral bands are removed. As a result, 200 spectral bands are remained. There are 16 land cover categories in this dataset, such as ``Grass-pasture-mowed", ``Soybean-mintill", ``Buildings-grass-trees-drives", and so on. The false-color image and ground-truth map of the \textit{Indian Pines} dataset are illustrated in Fig.~\ref{fig5}. The quantities of labeled and unlabeled pixels of various land cover categories used in our experiments are listed in \autoref{table1}.\vspace{-1em}

\begin{table}[t]
  \vspace{-2em}
  \scriptsize
  \centering
  \caption{Quantities of labeled and unlabeled pixels of all land cover categories in \textit{Indian Pines} dataset.}
  \label{table1}
  \begin{tabular}{c|c|c|c}
  \hline
  \hline
  ID & Class                        & \#Labeled & \#Unlabeled \\ \hline
  1  & Alfalfa                      & 30      & 16        \\
  2  & Corn-notill                  & 30      & 1398      \\
  3  & Corn-mintill                 & 30      & 800       \\
  4  & Corn                         & 30      & 207       \\
  5  & Grass-pasture                & 30      & 453       \\
  6  & Grass-trees                  & 30      & 700       \\
  7  & Grass-pasture-mowed          & 15      & 13        \\
  8  & Hay-windrowed                & 30      & 448       \\
  9  & Oats                         & 15      & 5         \\
  10 & Soybean-notill               & 30      & 942       \\
  11 & Soybean-mintill              & 30      & 2425      \\
  12 & Soybean-clean                & 30      & 563       \\
  13 & Wheat                        & 30      & 175       \\
  14 & Woods                        & 30      & 1235      \\
  15 & Buildings-grass-trees-drives & 30      & 356       \\
  16 & Stone-steel-towers           & 30      & 63        \\ \hline
  \hline
  \end{tabular}
  \end{table}

\subsection{\textit{University of Pavia}}
\begin{figure}[]
	\centering
	\includegraphics[width=8cm]{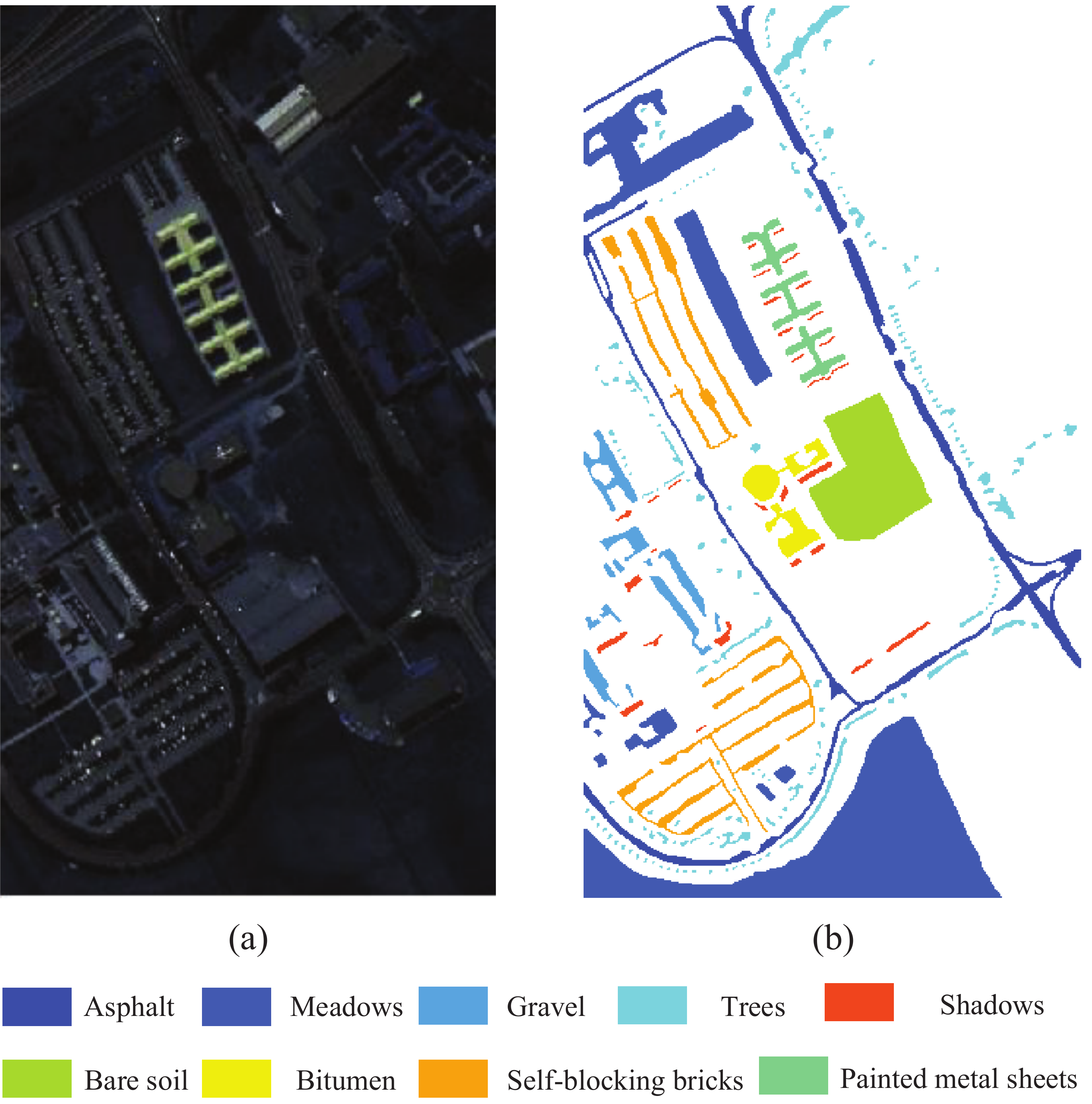}
	\caption{The \textit{University of Pavia} dataset. (a) False-color image. (b) Ground-truth map.}
	\label{fig6}
\end{figure}

In 2001, the \textit{University of Pavia} dataset was captured by the ROSIS sensor in the Pavia University, Italy. This dataset consists of $610 \times 340$ pixels with a spatial resolution of 1.3 m $\times$ 1.3 m. After removing noisy spectral bands, there are 103 spectral bands in the wavelength ranging from 0.43 to 0.86 $\upmu$m. There are 9 land cover categories in this dataset including ``Bare soil", ``Painted metal sheets", ``Self-blocking bricks", and so on. The false-color image and ground-truth map of the \textit{University of Pavia} dataset are illustrated in Fig.~\ref{fig6}. The quantities of labeled and unlabeled pixels of various land cover categories used in our experiments are listed in \autoref{table2}.\vspace{-1em}

\begin{table}[t]
  \scriptsize
  \centering
  \caption{Quantities of labeled and unlabeled pixels of all land cover categories in \textit{University of Pavia} dataset.}
  \label{table2}
  \begin{tabular}{c|c|c|c}
  \hline
  \hline
  ID & Class                & \#Labeled & \#Unlabeled \\ \hline
  1  & Asphalt              & 30      & 6601      \\
  2  & Meadows              & 30      & 18619     \\
  3  & Gravel               & 30      & 2069      \\
  4  & Trees                & 30      & 3034      \\
  5  & Painted metal sheets & 30      & 1315      \\
  6  & Bare soil            & 30      & 4999      \\
  7  & Bitumen              & 30      & 1300      \\
  8  & Self-blocking bricks & 30      & 3652      \\
  9  & Shadows              & 30      & 917      \\ \hline
  \hline
  \end{tabular}
  \end{table}

\subsection{\textit{Houston University}}
\begin{figure}[]
	\centering
	\includegraphics[width=8cm]{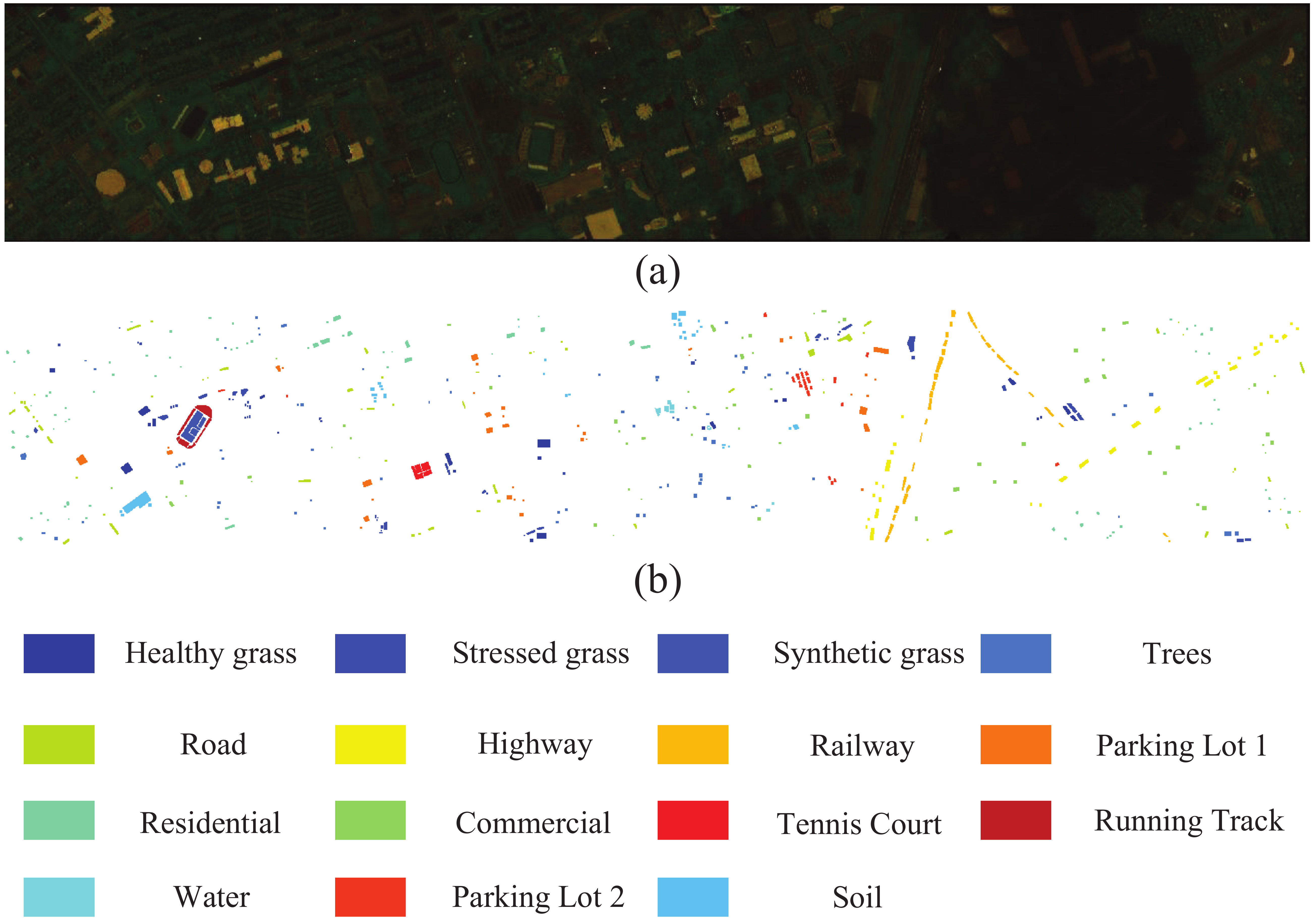}
	\caption{The \textit{Houston University} dataset. (a) False-color image. (b) Ground-truth map.}
	\label{fig8}
\end{figure}

The \textit{Houston University} dataset covered the Houston University campus and its neighboring areas. In 2013, the \textit{Houston University} dataset was utilized in the Geoscience and Remote Sensing Society Data Fusion Contest. This dataset contains $349 \times 1905$ pixels with a spatial resolution of 2.5 m $\times$ 2.5 m. In this dataset, there are 144 spectral bands ranging from 380 to 1050 nm. There are 15 land cover categories in this dataset, such as ``Tennis Court", ``Road", ``Highway", and so on. The false-color image and ground-truth map of the \textit{Houston University} dataset are illustrated in Fig. \ref{fig8}. The quantities of labeled and unlabeled pixels of various land cover categories used in our experiments are listed in \autoref{table4}.\vspace{-1em}

\begin{table}[t]
      \scriptsize
      \centering
      \caption{Quantities of labeled and unlabeled pixels of all land cover categories in \textit{Houston University} dataset.}
      \label{table4}
      \begin{tabular}{c|c|c|c}
      \hline
      \hline
      ID & Class           & \#Labeled & \#Unlabeled \\ \hline
      1  & Healthy grass   & 30      & 1344      \\
      2  & Stressed grass  & 30      & 1424      \\
      3  & Synthetic grass & 30      & 730       \\
      4  & Trees           & 30      & 1234      \\
      5  & Soil            & 30      & 1268      \\
      6  & Water           & 30      & 295       \\
      7  & Residential     & 30      & 1446      \\
      8  & Commercial      & 30      & 1324      \\
      9  & Road            & 30      & 1524      \\
      10 & Highway         & 30      & 1394      \\
      11 & Railway         & 30      & 1483      \\
      12 & Parking Lot 1   & 30      & 1399      \\
      13 & Parking Lot 2   & 30      & 540       \\
      14 & Tennis Court    & 30      & 451       \\
      15 & Running Track   & 30      & 728      \\ \hline
      \hline
      \end{tabular}
\end{table}

